\documentclass[runningheads]{llncs}
\usepackage[T1]{fontenc}
\usepackage{newtxtext}       %
\usepackage[varvw]{newtxmath}       

\usepackage{graphicx}
\usepackage[numbers]{natbib}
\usepackage{hyperref}
\hypersetup{
	colorlinks=true,
	linkcolor=blue,
	citecolor=blue,
	urlcolor=blue
}
\usepackage{url}
\usepackage{caption}
\usepackage{subcaption}
\usepackage[ruled,linesnumbered]{algorithm2e}
\usepackage{booktabs}
\usepackage{multirow}
\usepackage{array}
\usepackage{makecell}
\newcolumntype{C}[1]{>{\centering\arraybackslash}m{#1}}
\usepackage{comment}
\begin{document}
\title{Adaptive Conformal Redistribution for Inter-class Transitional Uncertainty in Medical Image Classification}
%
%
\author{Saibal Ghosh\inst{1} \and
Samarup Bhattacharya\inst{2} \and
Sanjoy Kumar Saha\inst{2} \and
Umapada Pal\inst{1} \and
Tapabrata
Chakraborti\inst{3}}
\authorrunning{S. Ghosh et al.}
%
\institute{Indian Statistical Institute, Kolkata 700108, India  
	\and
Jadavpur University, Kolkata 700032, India
\and
University College London, London WC1E 7JE, United Kingdom\\
}
\maketitle              
\begin{abstract}
    \textit{Background and Objective:} 	
	Medical image classification is frequently complicated by transitional categories whose feature distributions overlap those of adjacent classes, producing ambiguous decision boundaries. Conformal prediction returns uncertainty-aware prediction sets, but these are not directly actionable in clinical screening, where a single decision is required. This work proposes adaptive conformal redistribution (AdaConRed), a label-free post-conformal decision rule that converts ambiguous prediction sets into refined class assignments.
	
	\textit{Methods:} A five-stage pipeline is developed. Vision-language generative augmentation addresses minority-class scarcity; a frozen DermFoundation encoder provides embeddings; a lightweight multi-layer perceptron performs classification; an entropy-modulated, margin-aware nonconformity score constructs adaptive prediction sets; samples predicted as transitional with multi-label sets are reassigned to the most probable alternative class within the set, using only model outputs at inference. Evaluation uses the OSCC oral lesion and ISIC skin lesion benchmarks at a miscoverage level of $0.2$.
	
	\textit{Results:} On the 3-class OSCC benchmark, overall accuracy improves from 73.54\% to 77.38\%, with oral cancer accuracy rising from 64.29\% to 82.14\% and benign accuracy from 56.57\% to 70.20\%. Reassignment of transitional samples reduces OPMD accuracy from 84.78\% to 80.16\%, consistent with the asymmetric cost of missed malignancy. On ISIC, overall accuracy improves from 85.83\% to 87.19\%, melanoma accuracy rising from 66.04\% to 68.34\%. AdaConRed outperforms LAC, APS and RAPS under an identical backbone and redistribution rule. Code available at \url{https://github.com/saibal436ghosh/AdaConRed}.
	
	\textit{Conclusions:} Conformal prediction can be extended beyond uncertainty quantification toward actionable decision support where transitional disease categories are present, with gains concentrated in the clinically critical malignant categories.

\end{abstract}
\section{Introduction}
\label{sec:intro}
Medical image classification frequently involves disease categories that lie between well-defined states rather than forming discrete classes. For example, in oral cancer screening, oral potentially malignant disorders (OPMD) occupy an intermediate position between benign lesions and oral cancer (OCA), sharing visual characteristics with both~\cite{welikala2020automated}. Comparable ambiguity arises in dermoscopic assessment too, where heterogeneous lesion categories aggregate diagnoses that overlap several defined groups~\cite{ISIC1}. Such transitional categories produce indistinct decision boundaries and unreliable predictions, and the resulting errors are clinically asymmetric. A malignant lesion assigned to a transitional category delays diagnosis and treatment, whereas the converse error incurs only an additional review. This asymmetry is most consequential in resource-limited screening scenarios, where photographic or mobile phone-based imaging is often the only available modality and specialist confirmation is delayed~\cite{fu2020deep}.

These difficulties are compounded by the data conditions typical of medical imaging, where malignant categories are frequently the scarcest. Data scarcity and class imbalance remain persistent obstacles in practical medical AI deployments~\cite{ma2025cmpb}. Insufficient or non-uniform representation of such categories biases learning toward majority classes under conventional supervised objectives~\cite{cui2019class}, degrading sensitivity precisely where it matters most. To mitigate these issues, generative models have emerged as an effective approach for augmenting training datasets through synthetic sample generation~\cite{moris2024adapted}. Recent research has proposed the use of generative adversarial networks (GANs) and diffusion models in generating diverse synthetic samples, enriching underrepresented data distributions, and improving subsequent classification performance~\cite{frid2018gan,kazerouni2023diffusion}.
Additionally, recent advances in large-scale vision foundation models (VFMs) have further improved representation learning under data-constrained conditions through knowledge transfer from extensive pretraining and self-supervised learning~\cite{rani2024self}. These models have demonstrated strong generalization capabilities across diverse imaging tasks by providing rich semantic embeddings and transferable feature representations~\cite{zhang2023biomedclip}. In particular, domain-specific foundation models trained using multimodal supervision or large-scale datasets have shown improved robustness and data efficiency in challenging scenarios~\cite{kiraly2024dermfound,sellergreen2025medsiglip}.


Despite these advances, neither generative augmentation nor foundation model representation addresses the ambiguity introduced by transitional categories themselves. Classifiers built on them still produce deterministic point predictions that convey no measure of confidence in the assigned label. Generative augmentation adds a further consideration, since generative models introduce an additional source of stochastic variability~\cite{tzeng2025uncertaintydiffusion} and synthetic samples may contain hallucinated structures that deviate from the true data distribution~\cite{fu2025counting,aithal2024understanding}. In this work, synthetic samples are used only to enrich the training set, while calibration and evaluation are performed exclusively on real samples. The objective is therefore to improve classifier reliability under data scarcity rather than to quantify the uncertainty of the generated images themselves. Reliable uncertainty quantification is accordingly central to the trustworthiness of AI-assisted clinical decision systems~\cite{zou2023uncertainty}. Existing uncertainty modeling approaches include Bayesian deep learning, Monte Carlo dropout, aleatoric uncertainty estimation, and uncertainty-aware learning frameworks designed to identify unreliable or ambiguous predictions, which have shown considerable utility in high-stakes applications such as disease classification~\cite{nair2020exploring}.
Nevertheless, obtaining uncertainty estimates with rigorous statistical guarantees still remains an open challenge.
Conformal prediction addresses this limitation by offering a distribution-free uncertainty quantification framework that generates prediction sets with formal statistical guarantees, allowing uncertainty to be quantified in an interpretable manner~\cite{vovk2005algorithmic}. 
However, most existing conformal prediction methods primarily emphasize uncertainty quantification, while comparatively limited efforts toward explicitly leveraging such information for prediction refinement or improved decision-making~\cite{shi2024conformal}. 
Notably, under transitional or overlapping class boundaries, conformal predictors frequently return multi-label sets such as {OPMD, OCA}. Such sets are statistically valid but not directly actionable. Whereas, screening and triage require a single decision per patient, and an unresolved set defers that decision rather than supporting it. Recent empirical work has confirmed the reliability of conformal prediction for lesion classification~\cite{fayyad2024cmpb} but has likewise stopped at quantification. Converting these sets into refined single-label decisions, without discarding the uncertainty information they encode, is the gap this work addresses.


To address this limitation, a novel \textit{adaptive conformal redistribution} (AdaConRed) strategy is proposed. The framework constructs prediction sets using depth-wise cumulative probability mass, allowing richer uncertainty representation. Building on this, the proposed redistribution mechanism exploits the internal structure of prediction sets to resolve ambiguity. Uncertain samples are reassigned toward plausible alternative classes according to their relative confidence scores.
Main contributions of this work are summarized as:
(\textit{i}) a vision-language generative augmentation strategy addressing minority-class scarcity in malignant lesion categories, (\textit{ii}) an entropy-modulated, margin-aware conformal prediction framework constructing adaptive prediction sets that reflect transitional ambiguity, and (\textit{iii}) a label-free post-conformal redistribution rule converting ambiguous sets into single-label decisions, improving malignant-class sensitivity at a controlled cost to the transitional category.
The OSCC dataset~\cite{OSCC} is used as the primary benchmark due to its data scarcity, class imbalance, and transitional ambiguity, while additional experiments on the ISIC dataset~\cite{ISIC1} are conducted to evaluate generalizability under a diverse multi-class setting.

\section{Literature Survey}
\label{sec:lit_survey}
The early diagnosis of oral cancer is a clinical challenge, especially in situations where there is access to limited resources, and where specialist care cannot be rendered easily. Mobile phone based imaging is an alternative for such scenarios to address the problem of scanty resources. Myriad studies have been conducted to gauge the application of deep learning methodologies for the classification of oral lesions using photographs. 
Due to limited access to specialist care in rural and remote scenarios, prior studies have explored mobile phone-based imaging for oral cancer detection. In \cite{fu2020deep}, Fu et al. demonstrated that photographic images of the oral cavity, when combined with cascaded convolutional neural networks, can detect oral cavity squamous cell carcinoma with accuracy comparable to that of oral cancer specialists. Welikala et al. highlighted deep learning-based multi-class classification of oral lesions and reported potential misclassification among benign lesions, OPMD, and malignant lesions \cite{welikala2020automated}.
One of the primary challenges encountered in earlier works is the nature of OPMD lesions, which enhances the ambiguity between the other two classes of lesions - namely, benign and malignant.  
Such misclassification of malignant lesions can be lethal to the patient, as it leads to delayed diagnosis and treatment. 
Advances in pretrained VFMs have significantly enhanced deep learning performance in medical imaging, especially in scenarios with limited data availability. In \cite{dosovitskiy2020image}, Dosovitskiy et al. demonstrated that large-scale pretrained models can learn highly transferable representations. 
These models lessen reliance on large labeled datasets, thus making them suitable in domains where data is scarce such as in oral cancer classification tasks in low resource areas. 
However, despite improved feature learning, these approaches produce deterministic outputs and do not explicitly account for uncertainty arising from intermediate disease states such as OPMD, which share characteristics with both benign and malignant lesions. To address class imbalance, prior studies have explored data augmentation using generative models in various applications. For instance, Frid-Adar et al. demonstrated that data augmentation using GAN-generated samples can improve classification performance in minority as well as borderline classes \cite{frid2018gan}. Although such approaches can address data scarcity, class ambiguity in transitional or borderline classes remains an active challenge.

To handle uncertainty in classification tasks, \textit{conformal prediction} has been a statistically grounded approach for producing prediction sets with guaranteed coverage. Recent empirical evaluation of conformal prediction for skin-lesion classification demonstrated robust uncertainty quantification across multiple medical imaging datasets and under out-of-distribution conditions~\cite{fayyad2024cmpb}. Conformal prediction was first introduced by Vovk et al. as a framework for reliable uncertainty quantification \cite{vovk2005algorithmic}. In \cite{romano2020classification}, Romano et al. demonstrated the robustness of conformal prediction in handling classification tasks with ambiguous decision boundaries. In the context of medical imaging, conformal prediction has been increasingly explored to improve reliability and trustworthiness of deep learning models. 
Bhattacharyya et al. employed conformal prediction to study uncertainty and fairness characteristics of AI models for skin lesion classification across different patient subgroups \cite{bhattacharyya2026}. Banerji et al. emphasized that clinical AI systems must move beyond population-level accuracy and provide individualized predictive uncertainty, motivating the use of frameworks such as conformal prediction for more reliable and clinically meaningful decision-making \cite{banerji2023}. Chakraborti et al. emphasized the need for personalized uncertainty quantification in high-stakes AI applications, highlighting conformal prediction as a promising framework for generating distribution-free, set-valued predictions with statistical coverage guarantees \cite{chakraborti2025}. Dey et al. proposed a crossmodal generative framework that leverages histopathology images to enhance cancer grading and survival prediction, demonstrating improved predictive performance along with reliable uncertainty estimation using conformal prediction \cite{dey2026}. However, in most of these studies, conformal prediction has been used primarily for uncertainty quantification by mitigating overconfident errors and identifying uncertain predictions. In contrast, the proposed approach not only quantifies uncertainty but also leverages it to redistribute borderline predictions into clinically meaningful alternative categories, thereby improving classification performance.

Despite all these advancements, current methods fail to comprehensively address data imbalance, ambiguity in borderline classes, and uncertainty-aware decision-making in classification of oral lesions. Specifically, OPMD as a transitional disease class remains underexplored. To address this gap, this study combines generative AI-based data augmentation with conformal prediction-based redistribution to augment classification accuracy of oral cancer detection while minimizing clinically significant false negatives that may alter the clinical course of a patient. The proposed framework leverages robust feature representations from pretrained models, strengthens minority class learning, and incorporates uncertainty-aware reassignment of ambiguous samples, thereby augmenting  accuracy and boosting clinically relevant coverage.

\section{Methodology}
\label{sec:method}
\begin{figure*}[!t]
	\centering
	\includegraphics[width=\linewidth]{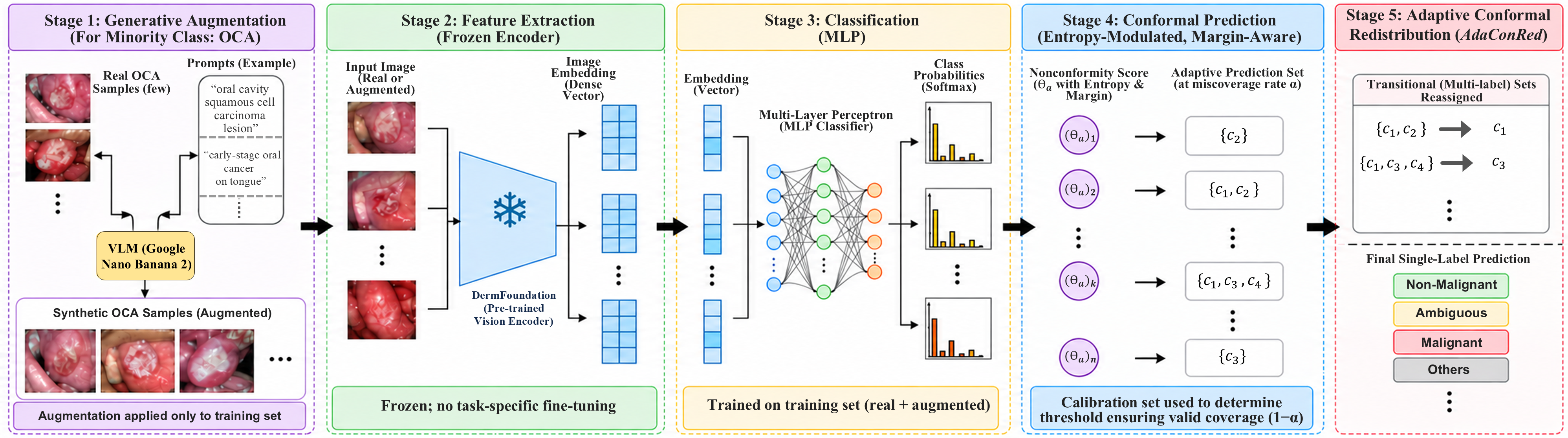}
	\vspace{1mm}
	\caption{Overview of the proposed AdaConRed pipeline with stages: (1) vision-language generative augmentation, (2) frozen DermFoundation encoding, (3) MLP-based classification, (4) entropy-modulated, margin-aware conformal prediction, and (5) adaptive conformal redistribution (AdaConRed).}
	\label{fig:block}
\end{figure*}

The pipeline proposed in this study comprises five sequential stages: image generation, image encoding, classification, conformal prediction, and adaptive conformal redistribution.
The subsequent sections detail each of these stages and the proposed pipeline is illustrated by Fig. \ref{fig:block}. 

\subsection{Dataset Description}
In this study, two publicly available medical image datasets are considered for evaluation. (\textit{i}) The OSCC oral lesion dataset~\cite{OSCC} serves as the primary experimental dataset. It contains three oral lesion categories, namely benign, oral potentially malignant disorder (OPMD), and oral cancer (OCA), comprising $748$, $1393$, and $129$ samples, respectively, together with $729$ healthy samples. The dataset exhibits severe data scarcity and class imbalance, particularly due to the limited number of OCA samples. Therefore, generative augmentation is further performed for the OCA category to alleviate class imbalance, as detailed in Section~\ref{subsec:augment}. (\textit{ii}) The ISIC skin lesion dataset~\cite{ISIC1} under a 7-class dermoscopic classification setting serves as a secondary benchmark to assess the generalizability of the proposed framework. A customized taxonomy is constructed by combining six lesion categories from ISIC-2019, namely actinic keratosis (AK: $1241$), basal cell carcinoma (BCC: $4298$), benign keratosis (BKL: $3284$), melanoma (MEL: $5849$), melanocytic nevus (NV: $15370$), and squamous cell carcinoma (SCC: $793$), together with the heterogeneous \textit{other/unknown} (UNK: $29171$) category from ISIC-2020~\cite{ISIC2}. Since sufficient samples are available across the selected categories, no additional augmentation strategy is employed for this dataset. Notably, the OPMD and UNK categories in the OSCC and ISIC datasets, respectively, exhibit transitional characteristics and are therefore utilized to evaluate ambiguity-aware behavior of the proposed framework.

\subsection{Generative Augmentation}
\label{subsec:augment}
\begin{figure*}[!t]
	\centering
	\includegraphics[width=\linewidth]{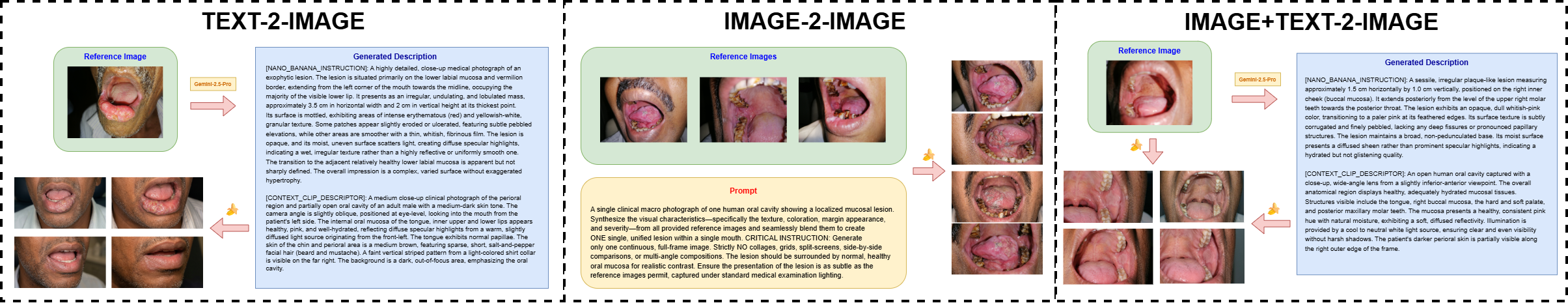}
	\caption{Illustration of different prompting modalities used to generate OCA samples.\hfill\null 
	}
	\label{fig:prompts}
	
\end{figure*}

Recent systematic evidence has highlighted the growing application of vision-language foundation models across medical imaging tasks, including classification, segmentation, and surface imaging, while also emphasizing challenges related to robustness and generalizability~\cite{sun2025cmpb}. In this work, the pre-trained vision language model (VLM) Google Nano Banana 2 was utilized to synthetically augment samples from the OCA class of the OSCC dataset. To preserve realistic visual and textural characteristics of the original OCA samples, a multi-modal prompting strategy was employed. Image synthesis was performed through the Flow interface~\cite{google2025flow}, generating up to four images per prompt. A total of $750$ synthetic OCA images were generated using three distinct prompting modalities, as detailed below and illustrated in Fig.~\ref{fig:prompts}.

\begin{enumerate}
	\item \textbf{Text-to-Image (T2I)}: 
	Individual OCA images from the training set were processed through the Gemini-2.5-Pro API~\cite{comanici2025gemini25} to extract clinically relevant textual descriptions. These descriptions captured lesion-specific characteristics such as texture, anatomical location, size, and color, together with global visual context from the image. The generated textual descriptions were subsequently provided as prompts to the Nano Banana 2 model to guide synthetic image generation.
	\item \textbf{Image-to-Image (I2I)}: 
	The prompt incorporated two to three reference images randomly selected from OCA samples within the training set. In addition, textual constraints were provided to preserve texture, illumination conditions, and overall visual characteristics of the reference images, while restricting collaged or unrealistic image generation.
	\item \textbf{Image+Text-to-Image (IT2I)}: 
	This hybrid prompting strategy combines both visual and textual information. Specifically, each training image was processed using the Gemini-2.5-Pro API to extract lesion-specific descriptions along with broader image characteristics. These generated textual cues were then combined with the corresponding reference image and provided to the Nano Banana 2 model to guide the synthesis.
\end{enumerate}

\subsection{Image Encoding}
Subsequently, high-dimensional feature representations were extracted from input images using Google’s DermFoundation~\cite{kiraly2024dermfound}, a state-of-the-art vision foundation model (VFM). The model is built upon the Big Transfer (BiT-M) framework and utilizes a widened ResNet101x3 backbone with group normalization and weight standardization to improve training stability. Its strong representation capability comes from a two-stage training strategy involving large-scale contrastive pretraining on image-text pairs, followed by supervised fine-tuning on geographically diverse dermatology datasets~\cite{kiraly2024dermfound}. 
In all experiments, DermFoundation~\cite{google2024dermfoundation,kiraly2024dermfound} was used as a fixed feature extractor without any task-specific fine-tuning or parameter updates. Images were processed directly through the frozen network in a zero-shot manner to obtain dense $6144$-dimensional embeddings while avoiding computationally expensive end-to-end model training. Since DermFoundation is pretrained predominantly on dermatological imagery, its zero-shot application to intraoral photographs involves a domain shift~\cite{pal2026isbi}. The empirical results indicate that the learned representations transfer effectively to this setting.

\subsection{Classification}
A lightweight multi-layer perceptron (MLP) is employed as the classification head for the high-dimensional embeddings extracted by the DermFoundation encoder. Instead of fine-tuning the entire foundation model, only the shallow MLP component is trained, allowing efficient adaptation while retaining the generalized feature representations learned by the backbone. To reduce the risk of overfitting under limited and imbalanced data conditions, a compact architecture is adopted. For the OSCC dataset, the MLP contains a single hidden layer with 128 neurons and a rectified linear unit (ReLU) activation, followed by a fully connected output layer for class prediction. For the 7-class ISIC setting, a second hidden layer with 32 neurons is inserted before the output layer to accommodate the larger label space; all other settings are identical across the two datasets. Training is performed using the Adam optimizer with a learning rate of 0.001 and a weight decay of $1\times10^{-4}$ for L2 regularization. To address class imbalance, class-weighted cross-entropy loss is incorporated so that greater emphasis is placed on underrepresented categories. In addition, label smoothing with a factor of 0.1 is incorporated to reduce prediction overconfidence and improve generalization, which also benefits the subsequent conformal prediction stage.

\subsection{Conformal Prediction Framework}
\label{sec:conformal}
To quantify prediction uncertainty in a statistically rigorous manner, the \textit{split conformal prediction} (SCP) framework is employed here~\cite{Angelopoulos2022}. Let the dataset be denoted as $\mathcal{D} = \{(x_i, y_i)\}_{i=1}^{N}$, where $x_i \in \mathbb{R}^d$ represents the feature embeddings extracted from the VFM and $y_i \in \mathcal{Y}=\{1,\dots,C\}$ denotes the corresponding class labels, where $C$ represents the total number of classes. For demonstration, lesion categories are conceptually grouped into non-malignant (NM), ambiguous (AMB), and malignant (M) types. The entire dataset is partitioned into four disjoint subsets: a training set $\mathcal{D}_{train}$ and validation set $\mathcal{D}_{val}$ for the classifier model (MLP in this study), a calibration set $\mathcal{D}_{cal}$ for the conformal quantile estimation, and a testing set $\mathcal{D}_{test}$ for evaluation. Here, the MLP network (defined as $f_\phi$) is trained on $\mathcal{D}_{train}$ to output softmax probabilities
$\hat{p}(y|x) = f_\phi(x)$. The calibration set $\mathcal{D}_{cal}$ is used to estimate how unusual or ``nonconforming'' a prediction is with respect to the model’s learned behavior. For each sample in this set, a \textit{nonconformity score} is computed, which measures the degree to which the model is uncertain or makes an incorrect prediction. Intuitively, a low nonconformity score indicates that the sample is well-aligned with the model’s predictions, whereas a high score suggests that the sample is atypical, uncertain, or potentially misclassified. 

\noindent \textbullet \quad \textbf{\textit{Nonconformity Score Formulation}}: A suitable nonconformity measure is defined based on the softmax outputs of the trained classifier. For a calibration sample $(x_i, y_i)$ defined above, the nonconformity score can be computed as:
\begin{equation}
	\theta_i = 1 - \hat{p}(y_i \mid x_i)
\end{equation}

\noindent The set of nonconformity scores i.e., calibration scores $\{\theta_i\}_{i=1}^{n_{cal}}$ is used to estimate a quantile threshold. For a desired miscoverage level $\alpha \in (0,1)$, corresponding quantile level is defined as
\begin{equation}
	\tau = \frac{\left\lceil (n_{cal}+1)(1-\alpha) \right\rceil}{n_{cal}}
	\label{eq:quantile}
\end{equation}

\noindent where $n_{cal} = |\mathcal{D}_{cal}|$ denotes the number of samples in the calibration set. Subsequently, the conformal threshold $\hat{q}_{\alpha}$ is computed as:
\begin{equation}
	\hat{q}_{\alpha} = \text{Quantile}\left( \{\theta_i\}_{i=1}^{n_{cal}} \; ; \; \tau \right)
\end{equation}
This threshold serves as the decision boundary for constructing prediction sets on unseen test samples.

\begin{figure}[!t]
	\centering
	\includegraphics[width=0.7\linewidth]{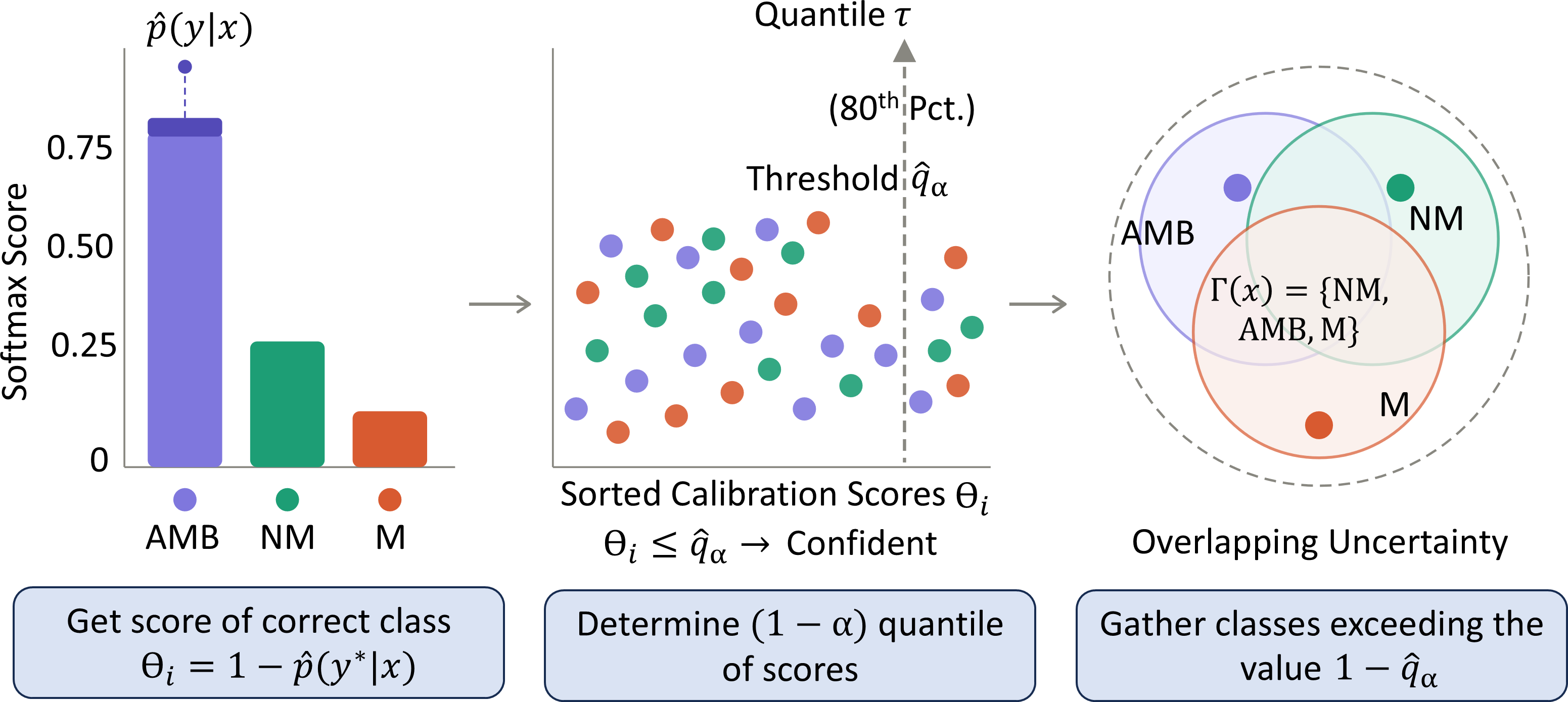}
	\vspace{2.5mm}
	\caption{An illustration of conformal prediction set construction using calibration-based quantile thresholding, resulting in uncertainty-aware overlapping class regions.}
	\label{fig:Conformal}
\end{figure}

\noindent \textbullet \quad \textbf{\textit{Prediction Set Construction and Coverage Guarantee}}: Given a test sample $x$, the conformal prediction set is defined as:
\begin{equation}
	\Gamma(x) = \left\{ y \in \mathcal{Y} \;:\; 1 - \hat{p}(y \mid x) \leq \hat{q}_{\alpha} \right\}
	\label{eq:cp_set}
\end{equation}
This can be equivalently written as:
\begin{equation}
	\Gamma(x) = \left\{ y \in \mathcal{Y} \;:\; \hat{p}(y \mid x) \geq 1 - \hat{q}_{\alpha} \right\}
\end{equation}
Thus, all labels whose predicted probability exceeds a data-driven threshold $1 - \hat{q}_{\alpha}$ are included in the prediction set. Under the exchangeability assumption between calibration and test samples~\cite{Angelopoulos2022}, SCP provides the following marginal coverage guarantee:
\begin{equation}
	\mathbb{P}(y \in \Gamma(x)) \geq 1 - \alpha
	\label{eq:coverage}
\end{equation}
This property holds regardless of the underlying data distribution or model complexity, making SCP well suited for uncertainty-aware image analysis tasks. An illustration of the conformal prediction set construction process, highlighting the role of calibration scores and uncertainty-guided prediction set formation, is presented in Fig.~\ref{fig:Conformal}.

\subsection{Proposed Adaptive Conformal Redistribution Strategy}
\label{subsec:adaconred}
Although the split conformal prediction framework provides statistically valid prediction sets, ambiguity arising from transitional classes remains unresolved. 
As a result, the conformal predictor frequently produces multi-label sets such as: $\Gamma(x) = \{ \text{AMB}, \text{M} \}, \quad \Gamma(x) = \{ \text{NM}, \text{AMB} \}$. In particular, the AMB category (e.g., OPMD or UNK) often introduces transitional ambiguity due to its overlapping morphological and visual characteristics with both non-malignant and malignant lesion patterns. Consequently, the generated conformal prediction sets appropriately capture uncertainty near clinically ambiguous decision boundaries. However, although such multi-label sets effectively reflect uncertainty-aware decision behavior, they are not always directly actionable in practical clinical screening and triaging scenarios, where a single refined decision is often required. To address this limitation, a conformal-based redistribution strategy is proposed to transform uncertainty-aware prediction sets into refined class predictions by leveraging the ranked probability structure and geometric characteristics of the conformal prediction set.
In the standard SCP framework, prediction sets are constructed using a calibrated threshold on individual class probabilities. While this guarantees marginal coverage, it does not account for the \textit{relative ranking} or \textit{distribution of probabilities} across classes. Consequently, in transitional scenarios such as AMB, probability mass is often distributed across neighboring classes. As a result, the resulting prediction sets may either exclude relevant alternative classes or include them in a manner that does not faithfully reflect their relative importance.
This limits the ability of SCP to fully utilize the underlying uncertainty information, particularly when distinguishing between closely related classes. 

Building upon this observation, an \textit{adaptive conformal redistribution} (AdaConRed) strategy is introduced in which prediction sets are formed using \textit{cumulative probability mass} rather than static thresholds on individual class probabilities. This allows the prediction set size to adapt according to the uncertainty associated with each sample. Consequently, multiple plausible class labels can be retained in ambiguous cases, providing a richer characterization of model uncertainty. In contrast to the standard formulation, which depends only on the probability assigned to an individual class, the adaptive strategy also incorporates information from the \textit{ranked probability distribution}, capturing how deeply a true label is positioned within the ordered confidence structure.

A suitable adaptive nonconformity measure is defined using the softmax outputs of the trained classifier $f_\phi$. Let $\hat{\pi}(x)= (\hat p(y|x))_{y\in\mathcal{Y}}$ denote the predicted class probability vector, where $\mathcal{Y}$ represents the set of class labels. The predicted probabilities satisfy $0 \leq \hat p(y|x)\leq 1$ and $\sum_{y\in\mathcal{Y}} \hat p(y|x)=1$. Unlike conventional adaptive conformal prediction methods~\cite{romano2020classification}, which rely solely on cumulative probability mass, the proposed approach further incorporates geometric characteristics of the probability distribution through a margin-aware weighting mechanism. This design is motivated by the observation that ambiguous samples typically produce flatter probability distributions with smaller confidence gaps between neighboring classes. In contrast, confident predictions exhibit a steeper decay in probability values. 
For a calibration sample $(x_i,y_i)$, the predicted probabilities are first arranged in descending order as $\hat p_{(1)} \geq \hat p_{(2)} \geq \cdots \geq \hat p_{(C)}$, where $C$ denotes the total number of classes. Let $r_i$ represent the rank of the true label $y_i$ within the ordered probability sequence. The proposed margin-aware adaptive nonconformity score is then defined as:
\begin{equation}
	(\theta_a)_i
	=
	\sum_{k=1}^{r_i}
	\left[
	1+\gamma_i
	\left|
	\hat p_{(k)}-\hat p(y_i|x_i)
	\right|
	\right]
	\hat p_{(k)}
	\label{eq:adaptive_score} 
\end{equation}

\noindent where $\hat p(y_i|x_i)$ denotes the predicted probability assigned to the true class, and the term $\hat p_{(k)}-\hat p(y_i|x_i)$ captures the confidence gap between the $k$-th ranked class and the true class probability. To adapt the influence of this gap-aware weighting according to the uncertainty associated with each sample, the scaling coefficient $\gamma_i$ is estimated using the entropy of the predicted probability distribution:
\begin{equation}
	\gamma_i
	=
	\lambda \cdot H(\hat{\pi}(x_i))
	\label{eq:gamma}
\end{equation}
where $\lambda \geq 0$ is a tunable scaling hyperparameter and $H(\hat{\pi}(x_i))$ denotes the Shannon entropy of the predicted probability distribution, given by:
\begin{equation}
	H(\hat{\pi}(x_i))
	=
	-\sum_{y\in\mathcal{Y}}
	\hat p(y|x_i)\log \hat p(y|x_i)
\end{equation}
This formulation allows the weighting mechanism to adapt according to the uncertainty characteristics of individual samples. For highly confident samples with low entropy, the contribution of the margin term becomes small, resulting in compact prediction sets. Conversely, ambiguous samples exhibiting flatter probability distributions produce larger entropy values, thereby increasing the influence of the margin-aware weighting and yielding more cautious prediction sets. The set of margin-aware adaptive nonconformity scores $\{(\theta_a)_i\}_{i=1}^{n_{cal}}$ is subsequently used to estimate the \textit{adaptive nonconformity threshold} as:
\begin{equation}
(\hat q_a)_\alpha
=
\mathrm{Quantile}
\left(
\{(\theta_a)_i\}_{i=1}^{n_{cal}};
\tau
\right)
\end{equation}
where $\tau$ is defined in~\eqref{eq:quantile}. Given a test sample $x$, the \textit{adaptive conformal prediction set} is then constructed by sequentially accumulating weighted probability mass according to the ranked probability distribution until the cumulative score exceeds the calibrated threshold $(\hat q_a)_\alpha$. Formally:
\begin{equation}
	\begin{split}
		\Gamma_a(x) = \left\{ y_1, y_2, \dots, y_k \right\}, \text{such that} 
		 \quad \sum_{i=1}^{k}
		\left[ 1+\gamma_x \Delta_i \right] \hat p(y_i|x)
		\geq (\hat q_a)_\alpha
	\end{split}
\label{eq:adaptive_set}
\end{equation}
where the class labels are ordered according to descending predicted probability, $\gamma_x$ denotes the entropy-adaptive scaling factor computed for the test sample, and $\Delta_i$ represents the probability margin between neighboring classes in the ordered probability sequence. Specifically:
\begin{equation}
	\Delta_i=
	\hat p(y_{i-1}|x)-\hat p(y_i|x)
\end{equation}
for $i>1$, while $\Delta_1=0$.

The prediction set is therefore constructed adaptively according to both cumulative probability mass and the local geometric structure of the probability distribution, enabling improved characterization of transitional ambiguity and uncertainty concentration. It should be noted that the adaptive score in~\eqref{eq:adaptive_score} is evaluated at the true label during calibration, whereas the test-time construction in~\eqref{eq:adaptive_set} accumulates margins between neighbouring classes. The two functionals therefore differ, and 
AdaConRed is accordingly presented as a post-conformal decision-refinement mechanism, while formal coverage remains the role of the standard split conformal set $\Gamma(x)$.

\begin{figure}[!t]
	\centering
	\includegraphics[width=0.7\linewidth]{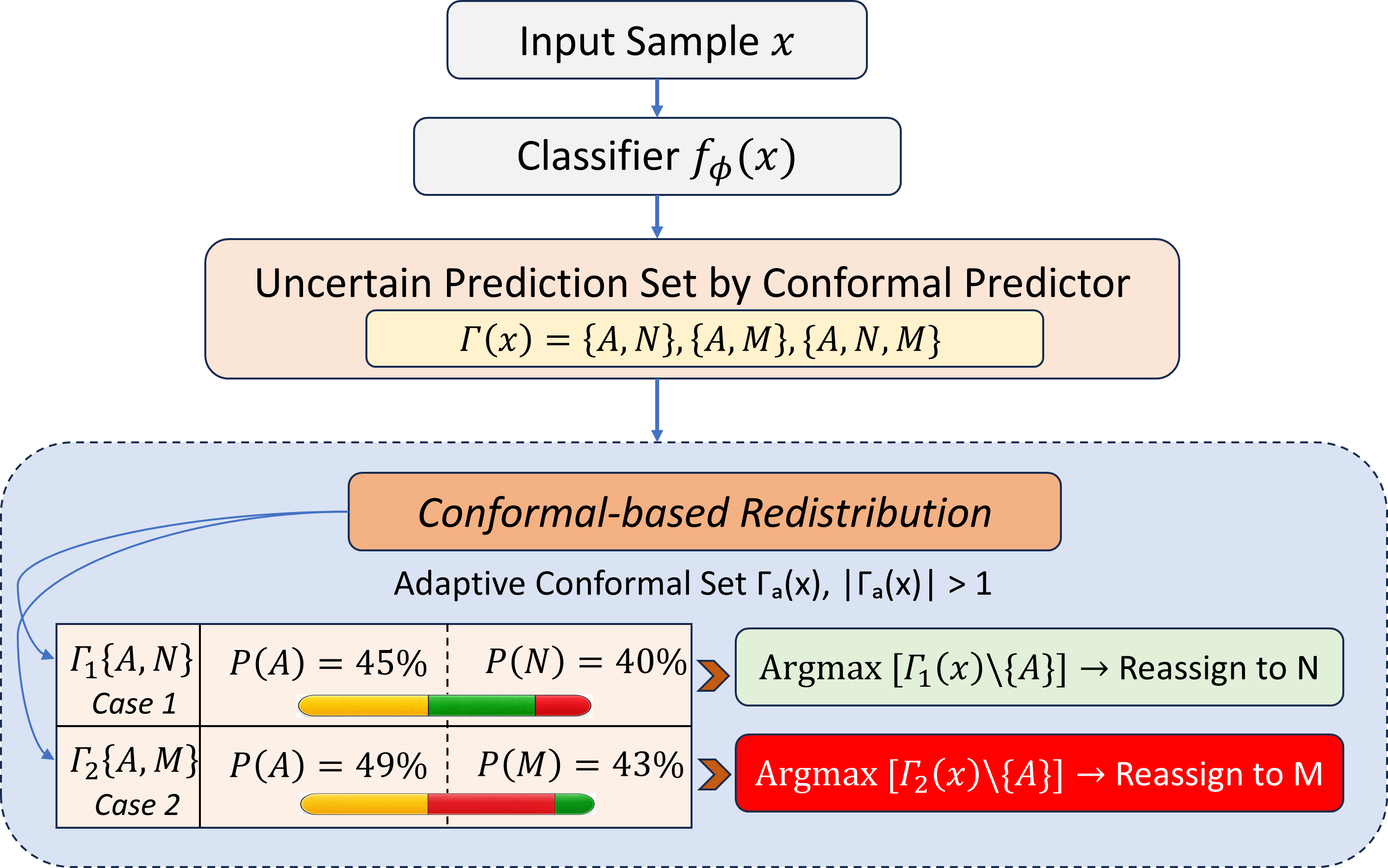}
	\vspace{2.5mm}
	\caption{Overview of the proposed redistribution strategy, where ambiguous samples from conformal prediction sets are reassigned to non-malignant or malignant categories, based on relative model confidence (N: non-malignant, A: ambiguous, M: malignant).}
	\label{fig:AdaConRed}
\end{figure}

\noindent \textbullet \quad \textbf{\textit{Redistribution Framework}}: Following the construction of adaptive conformal sets, we discuss the proposed redistribution framework here. Let $\hat{y}(x) = \arg\max_{y \in \mathcal{Y}} \hat{p}(y|x)$ denote the standard point prediction. The proposed method operates specifically on samples where:
\begin{equation}
	\hat{y}(x) = \text{AMB} \quad \text{and} \quad |\Gamma_{a}(x)| > 1
\end{equation}
\noindent The rule is label-free at inference, since it depends only on the point prediction and the cardinality of the adaptive conformal set, both computable from the model output alone. When a sample predicted as AMB is associated with a multi-label set, the presence of additional labels in $\Gamma_{a}(x)$ indicates ambiguity between AMB and its neighboring classes. Instead of assigning the transitional label, the method redistributes the decision to an alternative class based on relative model confidence. Formally, the redistributed label $\tilde{y}(x)$ is defined as:
\begin{equation}
	\tilde{y}(x) = \arg\max_{y \in \Gamma_{a}'(x)} \hat{p}(y|x)
\end{equation}

\noindent where $\Gamma_{a}'(x)$ is defined as the reduced conformal prediction set obtained by excluding the transitional class:
$\Gamma_{a}'(x) = \Gamma_{a}(x) \setminus \{\text{AMB}\}$. This corresponds to selecting the most probable non-AMB class within the conformal prediction set. 
The proposed redistribution mechanism can be interpreted as a post-conformal decision rule that preserves the uncertainty information encoded in $\Gamma_{a}(x)$ while improving classification performance. 
From a clinical perspective, the proposed strategy is designed to reduce \textit{false negatives} in malignant lesion detection by leveraging confidence-guided resolution of ambiguous (AMB) predictions toward more definitive classes. 
A visual illustration of the proposed redistribution strategy is presented in Fig.~\ref{fig:AdaConRed}.
The complete procedure is summarized in Algorithm~\ref{alg:adaconred}.

\begin{algorithm}[!t]
	\caption{Adaptive Conformal Redistribution (AdaConRed)}
	\label{alg:adaconred}
	\small
	
	\KwIn{Calibration scores $P_{cal}$, calibration labels $Y_{cal}$, test scores $P_{test}$, miscoverage level $\alpha$, entropy parameter $\lambda$}
	
	\KwOut{Refined predictions $\hat Y_{adj}$}
	
	\For{$i=1,\dots,N_{cal}$}{
		
		Sort probabilities:
		$p_{(1)}\ge p_{(2)}\ge\dots$
		
		Find rank $r_i$ of the true label
		
		Compute true-label probability:
		$p_{true}=p(y_i|x_i)$
		
		Compute entropy: $H_i = -\sum_{y \in \mathcal{Y}} p(y \mid x_i)\,\log p(y \mid x_i)$
		
		Compute adaptive parameter:
		$\gamma_{i}=\lambda H_i$
		
		Compute adaptive nonconformity score:
		$\theta_i = \sum_{k=1}^{r_i}\bigl(1 + \gamma_i\,\lvert p_{(k)} - p_{true}\rvert\bigr)p_{(k)}$
		
	}
	
	Estimate quantile level:
	$\tau = \frac{\left\lceil (N_{cal}+1)(1-\alpha) \right\rceil}{N_{cal}}$
	
	Estimate conformal threshold: $(\hat{q}_a)_\alpha = \mathrm{Quantile}\bigl(\{\theta_i\}_{i=1}^{N_{cal}};\,\tau\bigr)$
	
	\vspace{1mm}
	
	\For{$j=1,\dots,N_{test}$}{
		
		Sort probabilities:
		$p_{(1)}\ge p_{(2)}\ge\dots$
		
		Compute entropy: $H_j = -\sum_{y \in \mathcal{Y}} p(y \mid x_j)\,\log p(y \mid x_j)$
		
		Compute adaptive parameter:
		$\gamma_{gap}=\lambda H_j$
		
		Initialize:
		$S_0\leftarrow0$
		
		\For{$k=1,\dots,C$}{
			
			\eIf{$k=1$}{
				
				$\Delta_k\leftarrow0$
				
			}{
				
				$\Delta_k\leftarrow p_{(k-1)}-p_{(k)}$
				
			}
			
			Update cumulative score:
			$S_k
			\leftarrow
			S_{k-1}
			+
			(1+\gamma_{gap}\Delta_k)p_{(k)}$
			
		}
		
		Find:
		$k^*
		=
		\min\{k:S_k\ge (\hat{q}_a)_\alpha\}$
				
		Construct prediction set: $\Gamma_a(x_j) = \{y_{(1)}, \ldots, y_{(k^*)}\}$
		
		Compute initial prediction: $\hat{y}_j = \displaystyle\arg\max_{y \in \mathcal{Y}}\; \hat{p}(y \mid x_j)$
		
		\If{$\hat{y}_j = \textit{AMB}$ \textbf{and} $|\Gamma_a(x_j)| \geq 2$}{
			Construct reduced prediction set:
			$\Gamma'_a(x_j) \leftarrow \Gamma_a(x_j) \setminus \{\textit{AMB}\}$ 
			
			$\hat{y}_j \leftarrow \displaystyle\arg\max_{y \in \Gamma'_a(x_j)}\; \hat{p}(y \mid x_j)$
			
		}

	}
	
	Return refined predictions $\hat Y_{adj}$
	
\end{algorithm}


\section{Results and Discussion}
\label{sec:result}
This section presents a comprehensive evaluation of the proposed framework across diverse experimental settings, focusing on the effects of (\textit{i}) data augmentation and (\textit{ii}) conformal-based redistribution. The experiments are designed to analyze both classification performance and statistical coverage guarantees under severe class imbalance and transitional ambiguity. The OSCC~\cite{OSCC} and ISIC~\cite{ISIC1} datasets are considered for evaluation, where both exhibit inter-class ambiguity, while the OSCC dataset additionally presents severe minority-class underrepresentation and is therefore utilized for augmentation-based studies.

\subsection{Experimental Setup}
Two experimental configurations are considered for the OSCC dataset. The primary analysis is performed under a 3-class setting comprising benign, OPMD, and OCA, excluding the healthy category. Healthy samples do not represent lesion-bearing cases and are therefore less relevant from a clinical risk assessment perspective. Their exclusion also concentrates the evaluation on the transition between benign, potentially malignant, and malignant lesions, which is the regime the proposed redistribution strategy is designed to address. This configuration is used for the class-wise accuracy (Table~\ref{tab:classification_results}) and coverage (Table~\ref{tab:coverage_results}) results and the confusion matrices (Fig.~\ref{fig:confusion_matrices}), where malignant-class sensitivity is the principal outcome of interest. A 4-class configuration including the healthy category is additionally reported in Section~\ref{subsec:conformal_baseline} and~\ref{subsec:sota} to enable comparison with previously published studies, which adopt that taxonomy.
In the non-augmented scenario, the 3-class OSCC dataset is partitioned into training, validation, calibration, and test sets comprising $1320$, $50$, $300$, and $600$ samples, respectively. Due to the limited representation of OCA samples, only $34$ OCA instances are present in the test set, making it insufficient for robust performance estimation. In the augmented scenario (Aug. in tables), $750$ OCA samples are generated through the VLM and incorporated into the training data. Generated samples are \emph{only} added to the training set since their labels do not correspond to clinically validated ground truths. Subsequent to augmentation, $50$ real OCA instances were released from the training set and reassigned to the test set, thereby increasing the number of test OCA samples to $84$. No augmentation is performed for the ISIC dataset, and evaluations are conducted under a 7-class setting. The training, validation, calibration, and test sets contain $48004$, $631$, $3790$, and $7581$ samples, respectively. All experiments are performed with a conformal miscoverage rate $\alpha=0.2$ i.e., a target marginal coverage of $80\%$.

\subsection{Classification Performance Analysis}
\begin{table*}[!t]
	\renewcommand{\thetable}{1a}
	\caption{Class-wise accuracy (\%) comparison on the OSCC dataset (3-class)}
	\vspace{2mm}
	\label{tab:classification_results}
	\centering
	

\begin{tabular}{|l|c|c|c|c|}
	\hline
	{Method} & {All} & {Benign} & {OPMD} & {OCA} \\
	\hline
	ResNet-50 & 65.33 & 25.25 & \textbf{91.85} & 11.76 \\
	\hline
	ResNet-50 (Aug.) & 58.00 & 39.39 & 76.36 & 21.43 \\
	\hline
	InceptionV3 & 66.83 & 51.52 & 78.26 & 32.35 \\
	\hline
	InceptionV3 (Aug.) & 61.08 & 38.38 & 78.80 & 36.90 \\
	\hline
	MobileNet & 68.67 & 54.04 & 79.62 & 35.29 \\
	\hline
	MobileNet (Aug.) & 61.69 & 36.36 & 80.43 & 39.29 \\
	\hline
	MSL+MLP & 73.50 & 67.17 & 79.08 & 50.00 \\
	\hline
	MSL+MLP (Aug.) & 72.62 & 60.10 & 83.70 & 53.57 \\
	\hline
	MSL+MLP+AdaConRed (Aug.) & 75.23 & 69.19 & 79.35 & 71.43 \\
	\hline
	DF+MLP & 74.67 & \textbf{74.24} & 78.53 & 35.29 \\
	\hline
	DF+MLP (Aug.) & 73.54 & 56.57 & 84.78 & 64.29 \\
	\hline
	DF+MLP+AdaConRed (Aug.) & \textbf{77.38} & 70.20 & 80.16 & \textbf{82.14} \\
	\hline
\end{tabular}

\vspace{1mm}
\raggedright
\scriptsize Non-augmented and augmented rows are evaluated on test sets containing $34$ and $84$ OCA samples respectively ($600$ and $650$ samples in total), and are therefore not directly comparable. Comparisons should be drawn within, not across, the two blocks.
\end{table*}

\begin{table*}[!t]
	\renewcommand{\thetable}{1b}
	\caption{Class-wise accuracy (\%) comparison on the ISIC dataset (7-class)}
	\vspace{2mm}
	\label{tab:classification_results_isic}
	\centering
	

\begin{tabular}{|l|c|c|c|c|c|c|c|c|c|}
	\hline
	{Method} & {All} & {AK} & {BCC} & {BKL} & {MEL} & {NV} & {SCC} & {UNK} \\
	\hline
	ResNet-50 & 73.37 & 54.78 & 73.30 & 38.55 & 53.59 & 65.14 & 54.00 & 86.92 \\
	\hline
	InceptionV3 & 73.46 & 51.59 & 63.72 & 53.25 & 47.77 & 68.49 & 67.00 & 86.05 \\
	\hline
	MobileNet & 75.52 & \textbf{61.78} & 55.25 & 41.69 & 48.31 & 78.32 & 52.00 & 87.52 \\
	\hline
	MSL+MLP & 85.78 & 50.32 & 84.90 & 63.86 & 65.90 & 86.92 & 62.00 & 93.92 \\
	\hline
	MSL+MLP+AdaConRed & 85.82 & 50.96 & 86.56 & \textbf{73.73} & 67.39 & 88.21 & 64.00 & 91.59 \\
	\hline
	DF+MLP & 85.83 & 59.87 & 84.71 & 58.07 & 66.04 & 86.51 & 73.00 & \textbf{94.19} \\
	\hline
	DF+MLP+AdaConRed & \textbf{87.19} & 61.15 & \textbf{86.92} & 61.20 & \textbf{68.34} & \textbf{93.61} & \textbf{74.00} & 92.02 \\
	\hline
\end{tabular}
\end{table*}

\setcounter{table}{1}
\renewcommand{\thetable}{\arabic{table}}

\begin{figure}[!t]
	\centering
	\begin{subfigure}[t]{0.3\linewidth}
		\centering
		\includegraphics[width=\linewidth]{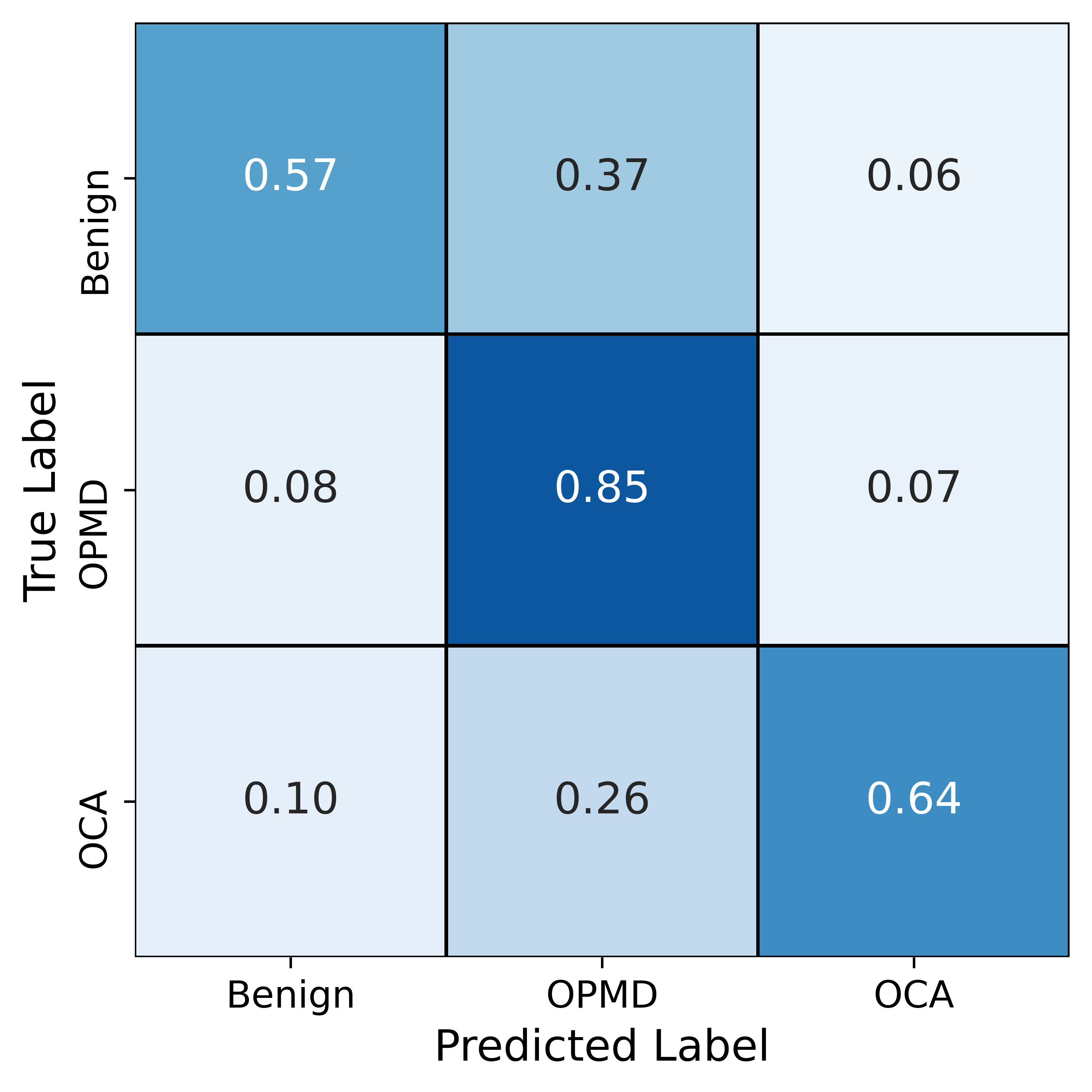}
		\caption{}
		\label{fig:cm_before}
	\end{subfigure}
	\hspace{2mm}
	\begin{subfigure}[t]{0.3\linewidth}
		\centering
		\includegraphics[width=\linewidth]{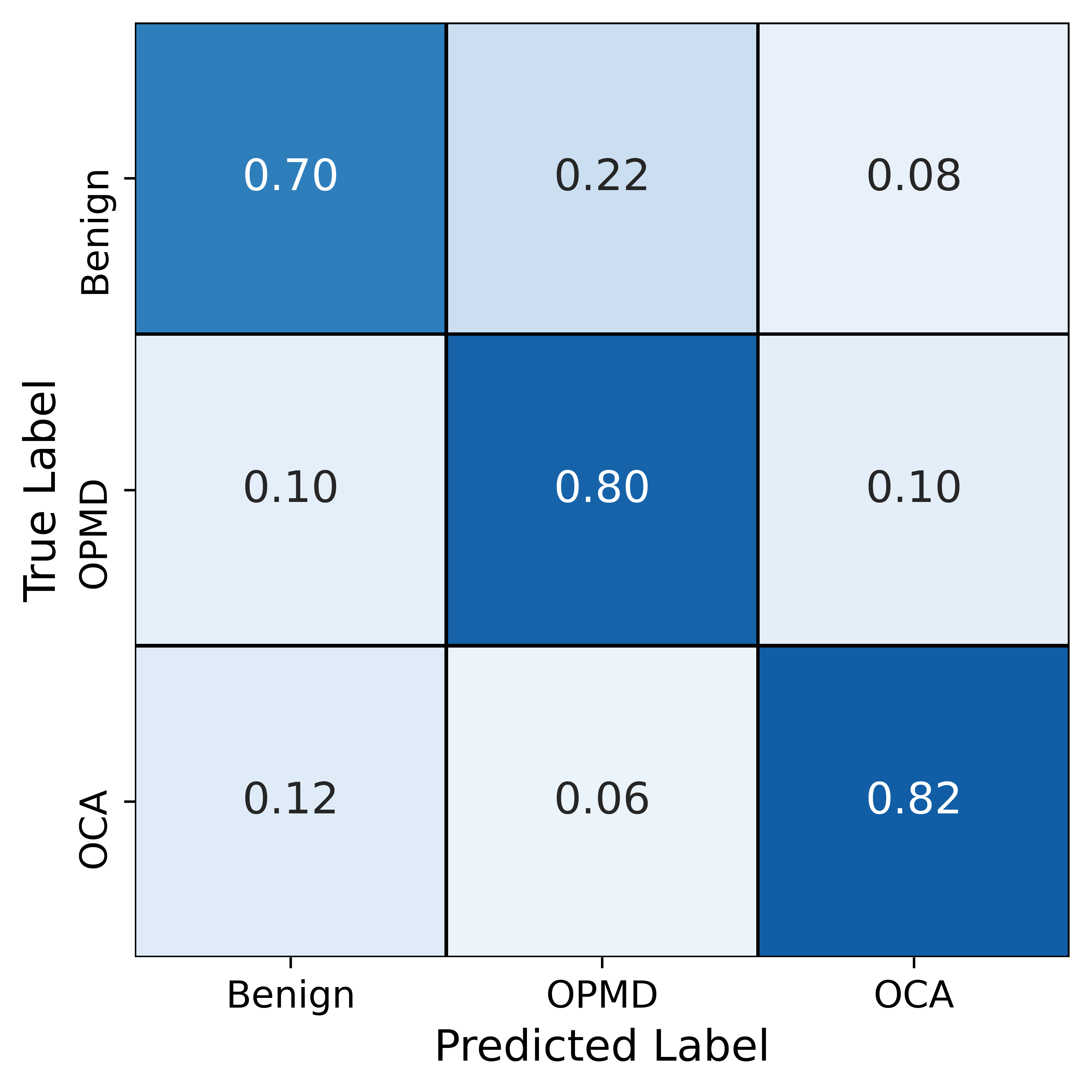}
		\caption{}
		\label{fig:cm_after}
	\end{subfigure}
	\caption{Confusion matrices (row-normalised) for DF+MLP (Aug.) on the 3-class OSCC dataset
		(a) before and (b) after adaptive conformal redistribution.}
	\label{fig:confusion_matrices}
\end{figure}

Table~\ref{tab:classification_results} summarizes the classification performance across different methods for the 3-class OSCC dataset (benign, OPMD, and OCA). MSL and DF refer to the MedSigLIP~\cite{sellergreen2025medsiglip} and DermFoundation~\cite{kiraly2024dermfound} embedding models, respectively. MedSigLIP is chosen as an alternative embedding model due to its pretraining on medical imaging datasets, particularly including dermatology images, enabling a fair and clinically relevant comparison. Across all three convolutional baselines, augmentation raises OCA accuracy while reducing overall accuracy, mirroring the trade-off observed for the foundation-model backbones and indicating that the effect is a property of class rebalancing rather than of any particular encoder. Without augmentation, the DF+MLP model achieves an overall accuracy of $74.67\%$, with comparable performance on benign ($74.24\%$) and OPMD ($78.53\%$) categories. However, the OCA accuracy remains critically low at $35.29\%$, indicating poor sensitivity toward malignant lesions. After augmentation and increasing test representation of the OCA class, its accuracy improves significantly to $64.29\%$. 
However, this improvement comes at the cost of reduced benign accuracy ($56.57\%$) and a slight drop in overall accuracy ($73.54\%$). This trade-off likely reflects a shift in decision boundaries induced by class rebalancing. 

The proposed AdaConRed strategy further refines predictions, leading to an overall accuracy of 
$77.38\%$ as shown in Table~\ref{tab:classification_results}, surpassing both the baseline and augmented scenarios. Notably, OCA accuracy increases to $82.14\%$ and benign accuracy recovers to $70.20\%$, demonstrating that redistribution effectively recovers cancerous lesions previously misassigned to OPMD. Since redistribution is applied without access to GT labels, a portion of genuinely transitional samples is also reassigned: 17 of the 368 true-OPMD test samples are moved (6 to benign, 11 to OCA), reducing OPMD accuracy from $84.78\%$ to $80.16\%$. The rule therefore trades a modest loss on the transitional category for improved malignancy sensitivity, consistent with the asymmetric clinical cost of missed cancers in lesion screening. Confusion matrices presented in Fig.~\ref{fig:cm_before} and ~\ref{fig:cm_after} show the effect of redistribution at the level of individual predictions. Samples previously assigned to OPMD are reassigned to benign and OCA, increasing the diagonal entries for both classes while reducing the OPMD diagonal entries. Qualitative examples illustrating the redistribution mechanism are presented in Table~\ref{tab:conformal_examples}. AdaConRed utilizes clinically plausible alternative labels within the conformal prediction set to reassign ambiguous non-OPMD samples to their correct classes.

 \begin{table*}[!t]
	 \renewcommand{\thetable}{2a}
	 	\caption{Illustrative examples of conformal redistribution on the OSCC dataset}
	 	\vspace{2mm}
	 	\label{tab:conformal_examples}
	 	\centering
	 	\setlength{\tabcolsep}{4pt}
	 	\renewcommand{\arraystretch}{1.1}
	
	 	\begin{tabular}{|c|c|c|p{4.8cm}|c|}
		 		\hline
		 		Sample & GT & Predicted & \centering Conformal Set & Final Label \\
		 		\hline
		
		 		\begin{tabular}{c}
			 			\includegraphics[width=2cm]{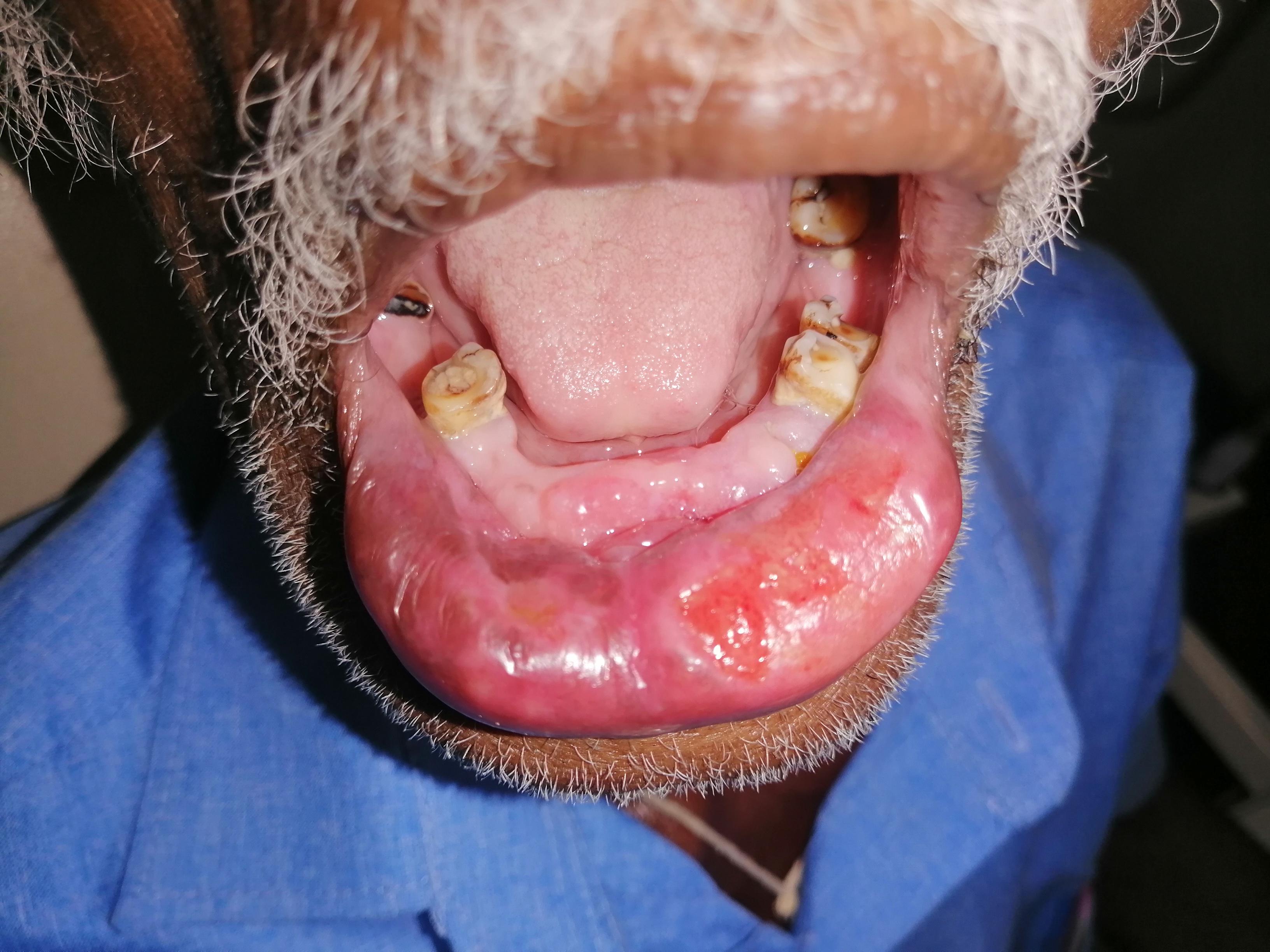} \\
			 			ID: R-107-04
			 		\end{tabular}
		 		& benign & OPMD & 
		 		\raggedright \{OPMD: 0.4780, benign: 0.3320\} & benign \\
		 		\hline
		
		 		\begin{tabular}{c}
			 			\includegraphics[width=2cm]{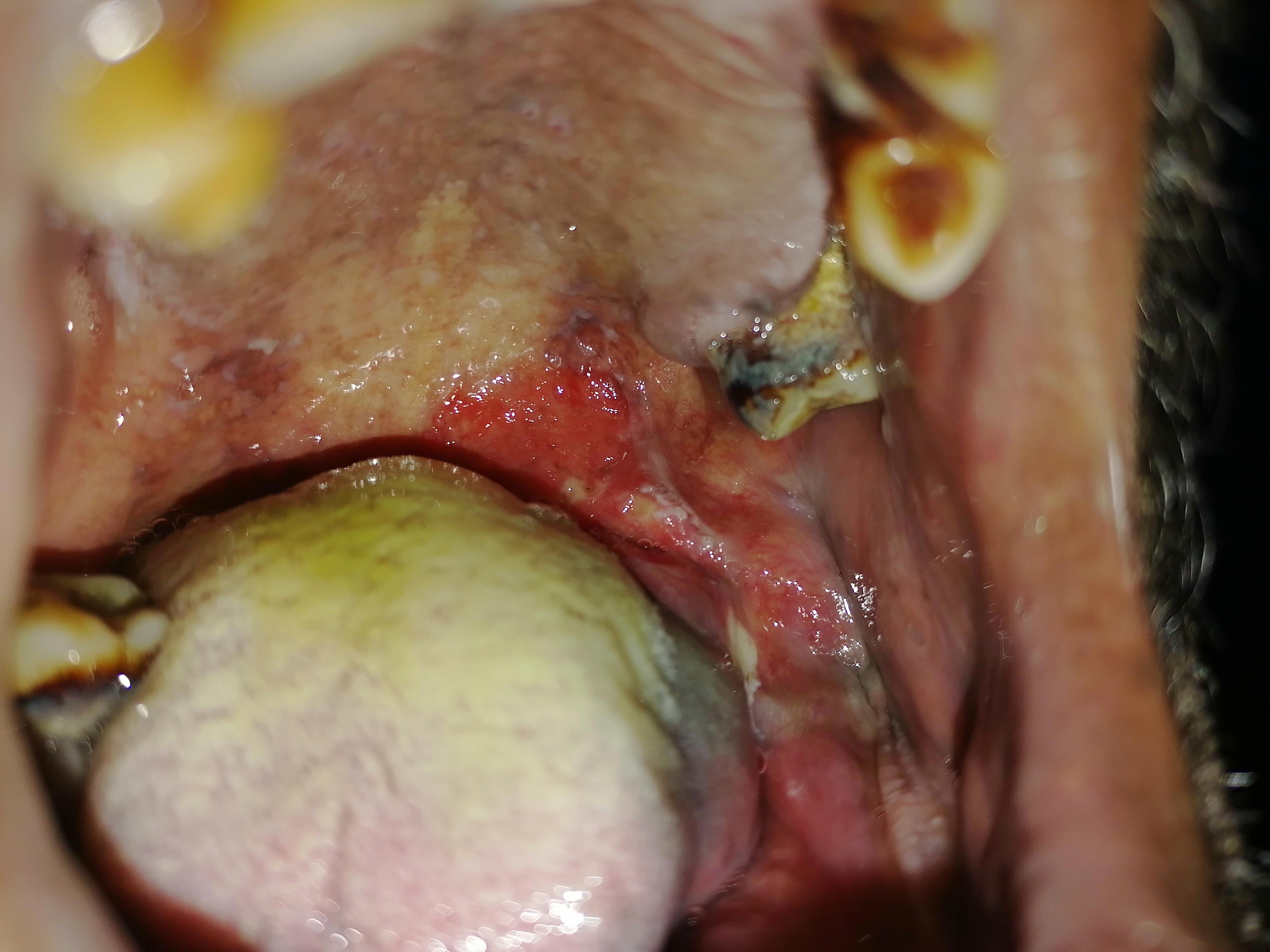} \\
			 			ID: R-209-01
			 		\end{tabular}
		 		& OCA & OPMD & 
		 		\raggedright \{OPMD: 0.6117, OCA: 0.3132\} & OCA \\
		 		\hline
		
		 	\end{tabular}
	 \end{table*}


 \begin{table*}[!t]
	 \renewcommand{\thetable}{2b}
	 	\caption{Illustrative examples of conformal redistribution on the ISIC dataset}
	 	\vspace{2mm}
	 	\label{tab:conformal_examples_ISIC}
	 	\centering
	 	\setlength{\tabcolsep}{4pt}
	 	\renewcommand{\arraystretch}{1.1}
	
	 	\begin{tabular}{|c|c|c|p{4.8cm}|c|}
		 		\hline
		 		Sample & GT & Predicted & \centering Conformal Set & Final Label \\
		 		\hline
		
		 		\begin{tabular}{c}
			 			\includegraphics[width=2cm,height=2cm]{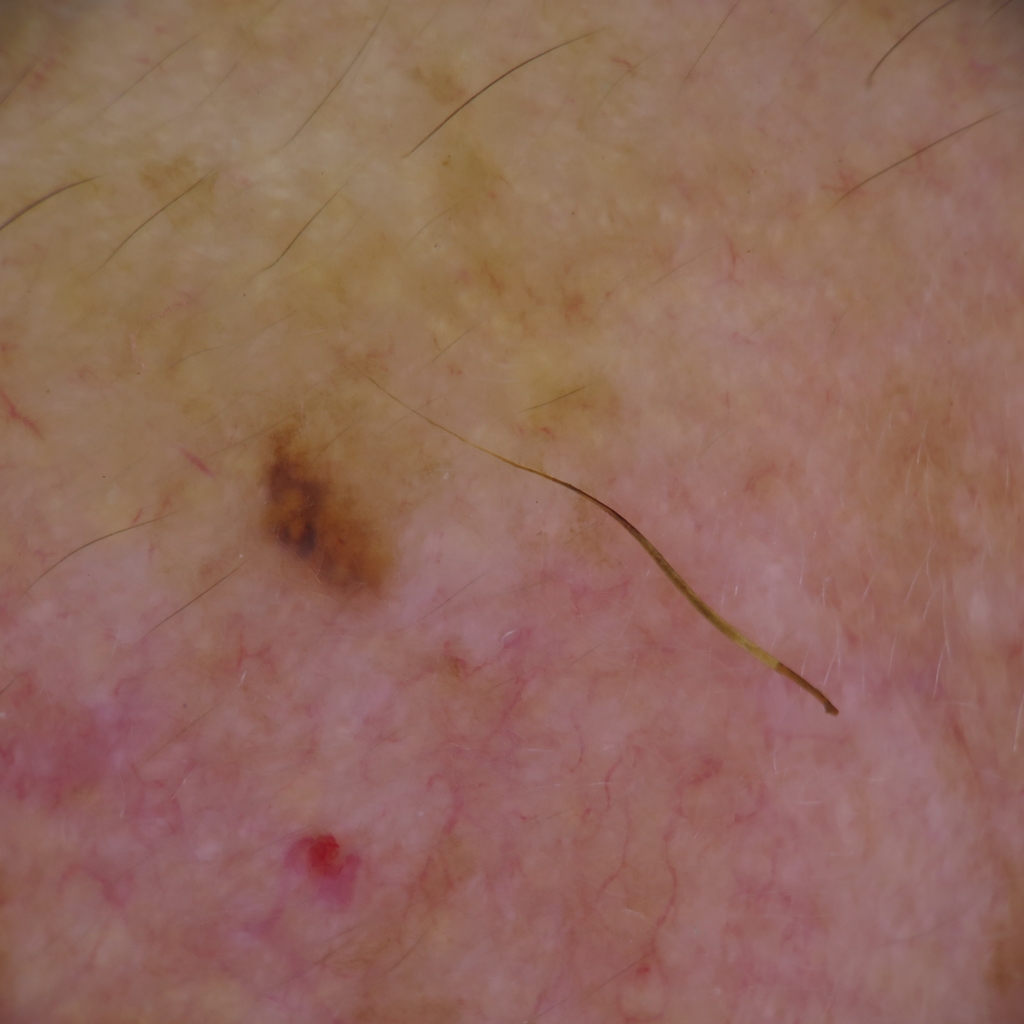} \\
			 			ID: ISIC\_0061156
			 		\end{tabular}
		 		& BCC & UNK & 
		 		\raggedright \{UNK: 0.5934, BCC: 0.4036\} & BCC \\
		 		\hline
		
		 		\begin{tabular}{c}
			 			\includegraphics[width=2cm,height=2cm]{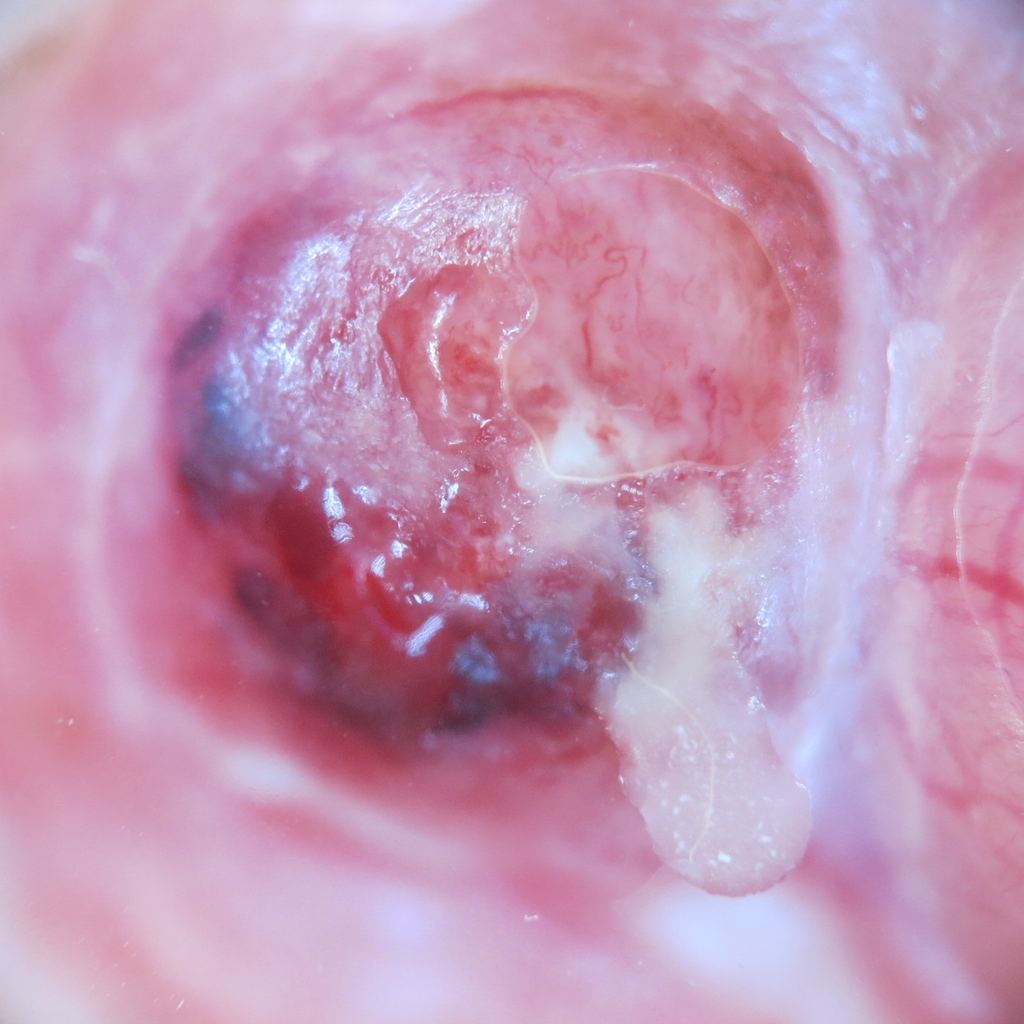} \\
			 			ID: ISIC\_0063360
			 		\end{tabular}
		 		& MEL & UNK & 
		 		\raggedright \{UNK: 0.4158, MEL: 0.3279\} & MEL \\
		 		\hline
		
		         \begin{tabular}{c}
			 			\includegraphics[width=2cm,height=2cm]{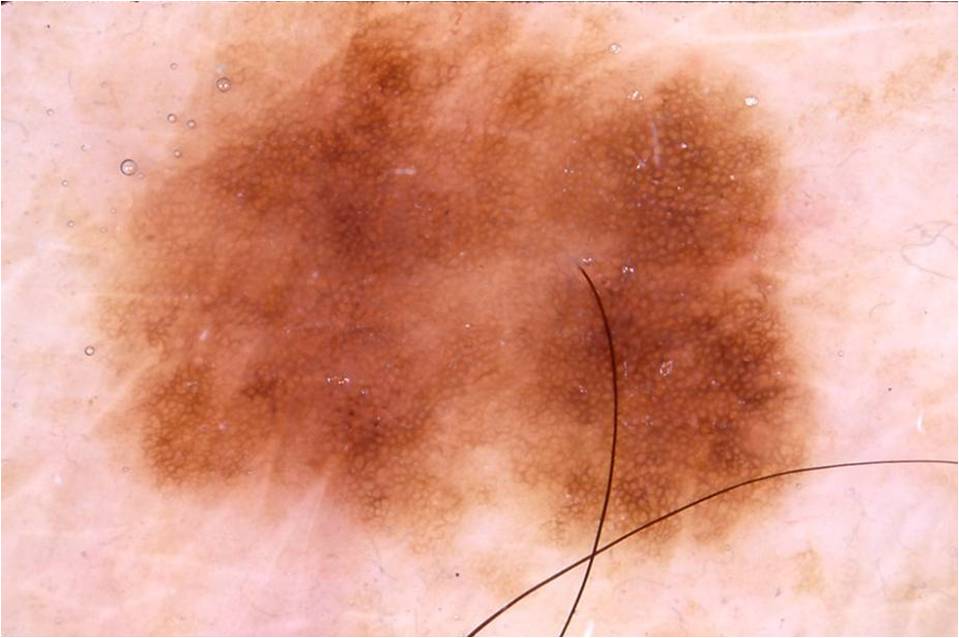} \\
			 			ID: ISIC\_0000217
			 		\end{tabular}
		 		& NV & UNK & 
		 		\raggedright \{UNK: 0.5546, NV: 0.3878\} & NV \\
		 		\hline
		
		 	\end{tabular}
	 \end{table*}


\setcounter{table}{2}
\renewcommand{\thetable}{\arabic{table}}

To evaluate generalizability, additional experiments are conducted on the ISIC dataset. Table~\ref{tab:classification_results_isic} summarizes the class-wise classification performance under the 7-class setting. The DF+MLP model achieves an overall accuracy of 85.83\%, which increases to 87.19\% after incorporating AdaConRed. Gains are observed across every lesion category, most substantially for NV (+7.10 points), followed by BKL (+3.13), MEL (+2.30) and BCC (+2.21), with the corresponding reduction confined to the transitional UNK category (-2.17 points). Qualitative examples illustrating the redistribution process on the ISIC dataset are provided in Table~\ref{tab:conformal_examples_ISIC}. 
AdaConRed is intended as a decision-support mechanism rather than an autonomous diagnostic system. Redistribution is applied only when the conformal prediction set contains multiple clinically plausible labels, and the original set remains available alongside the refined decision, so that the underlying ambiguity is preserved for clinical review.

\subsection{Statistical Coverage Analysis}
The detailed conformal coverage results on the OSCC dataset across different experimental settings (miscoverage rate $\alpha=0.2$) are presented in Table~\ref{tab:coverage_results}. From an uncertainty standpoint, the OCA class exhibits a substantial improvement in coverage after augmentation, increasing from $44.12\%$ to $76.19\%$ for DF+MLP, which indicates enhanced prediction reliability for this minority class. A similar improvement for OCA coverage in the same direction is observed for MSL+MLP, from 58.82\% to 69.05\%, though its overall coverage decreases slightly from 80.33\% to 78.31\%. Since split conformal prediction provides marginal rather than class-conditional coverage, per-class coverage is not guaranteed to meet the nominal level. The OCA coverage of $76.19\%$ accordingly remains below the $80\%$ target, which is expected for a minority class under marginal calibration. Overall coverage exceeds the nominal level for both DermFoundation configurations (84.00\% and 82.31\%) and for the non-augmented MedSigLIP model (80.33\%), while the augmented MedSigLIP configuration falls marginally below at 78.31\%.

Table~\ref{tab:coverage_results_isic} summarizes the conformal coverage performance on the ISIC dataset under the 7-class setting. The DF+MLP model achieves an overall coverage of $89.90\%$, reflecting reliable uncertainty quantification across a diverse range of lesion categories. High coverage values are observed for classes such as NV ($91.81\%$) and UNK ($95.79\%$), whereas relatively lower values for AK and BKL suggest greater predictive uncertainty for these lesion types. Overall, these findings highlight the ability of the proposed framework to maintain reliable uncertainty-aware behavior across diverse medical imaging scenarios.

\begin{table}[!t]
	\renewcommand{\thetable}{3a}
	\caption{Conformal prediction coverage (\%) for the standard conformal set $\Gamma(x)$ on the OSCC dataset (3-class)}
	\vspace{2mm}
	\label{tab:coverage_results}
	\centering
	\begin{tabular}{|l|c|c|c|c|}
		\hline
		{Method} & {All} & {Benign} & {OPMD} & {OCA} \\
		\hline
		MSL+MLP & 80.33 & 73.23 & 86.14 & 58.82 \\
		\hline
		MSL+MLP (Aug.) & 78.31 & 63.64 & 88.32 & 69.05 \\
		\hline
		DF+MLP & 84.00 & 83.84 & 87.77 & 44.12 \\
		\hline
		DF+MLP (Aug.) & 82.31 & 67.17 & 91.85 & 76.19 \\
		\hline
	\end{tabular}
\end{table}

\begin{table*}[!t]
	\renewcommand{\thetable}{3b}
	\caption{Conformal prediction coverage (\%) for the standard conformal set $\Gamma(x)$ on the ISIC dataset (7-class)}
	\vspace{2mm}
	\label{tab:coverage_results_isic}
	\centering
	\begin{tabular}{|l|c|c|c|c|c|c|c|c|}
		\hline
		{Method} & {All} & {AK} & {BCC} & {BKL} & {MEL} & {NV} & {SCC} & {UNK} \\
		\hline
		MSL+MLP & 80.73 & 45.86 & 77.53 & 65.06 & 53.59 & 77.39 & 50.00 & 92.48 \\
		\hline
		DF+MLP & 89.90 & 68.15 & 85.45 & 67.95 & 77.40 & 91.81 & 77.00 & 95.79 \\
		\hline
	\end{tabular}
\end{table*}
\setcounter{table}{3}
\renewcommand{\thetable}{\arabic{table}}

\subsection{Comparison with Conformal Baselines, Sensitivity Analysis, and Ablation}
\label{subsec:conformal_baseline}
The proposed adaptive score differs from conventional adaptive conformal
prediction through the entropy-modulated, margin-aware weighting introduced
in~\eqref{eq:adaptive_score}. To isolate the contribution of this weighting,
the score function is varied while the backbone, data splits, calibration
set, and miscoverage level are held fixed. Three established alternatives are
considered: the least-ambiguous set classifier (LAC)~\cite{LAC}, which thresholds
individual class probabilities as in~\eqref{eq:cp_set}; adaptive prediction
sets (APS)~\cite{romano2020classification}, which accumulate ranked cumulative
probability mass; and regularized adaptive prediction sets (RAPS)~\cite{RAPS}, which
penalize set cardinality to discourage excessively large sets. For each score,
the redistribution rule of Section~\ref{subsec:adaconred} is applied
unchanged, so that any difference in refined accuracy is attributable to the
set construction alone. Results are reported in
Table~\ref{tab:conformal_comparison}.

\begin{table*}[!t]
	\caption{Comparison of conformal score functions under a fixed DF+MLP backbone at $\alpha = 0.2$. The redistribution rule is applied unchanged in all cases, so differences are attributable to set construction alone. Malignant-class accuracy refers to OCA for OSCC and MEL for ISIC. OSCC results use the 4-class setting, ISIC the 7-class setting.}
	\label{tab:conformal_comparison}
	\centering
	\begin{tabular}{|l|l|c|c|c|c|}
		\hline
		Dataset & Score & Malig. Acc. (\%) & All Acc. (\%) & All F1 (\%) & All Spec. (\%) \\
		\hline
		\multirow{4}{*}{OSCC}
		& LAC       & 67.86 & 76.06 & 73.86 & 90.42 \\
		& APS       & 72.62 & 77.21 & 75.46 & 91.62 \\
		& RAPS      & 75.00 & 77.72 & 75.80 & 92.10 \\
		& Proposed  & \textbf{79.76} & \textbf{79.64} & \textbf{77.18} & \textbf{93.79} \\
		\hline
		\multirow{4}{*}{ISIC}
		& LAC       & 66.31 & 86.03 & 70.47 & 94.27 \\
		& APS       & 66.85 & 86.26 & 70.88 & 95.36 \\
		& RAPS      & 66.98 & 87.05 & 71.64 & 96.79 \\
		& Proposed  & \textbf{68.34} & \textbf{87.19} & \textbf{75.33} & \textbf{97.72} \\
		\hline
	\end{tabular}

\vspace{1mm}
\raggedright
\scriptsize The corresponding no-redistribution baselines under the same backbone are 75.67\% overall accuracy on OSCC and 85.83\% on ISIC.
\end{table*}

Redistribution improves overall accuracy relative to the corresponding no-redistribution baseline (75.67\% on OSCC, 85.83\% on ISIC) under every score function considered, so the benefit of the rule is not contingent on a particular set construction. What varies is the magnitude of the gain. Under LAC the improvement is marginal, 0.39 points on OSCC and 0.20 points on ISIC, since thresholding individual class probabilities frequently yields singleton sets from which no alternative class can be recovered. Refined accuracy then increases progressively from LAC to APS to RAPS on both datasets, indicating that set constructions accounting for the ranked probability structure retain the correct alternative class more often. The proposed score attains the highest values across all reported metrics on both benchmarks, improving overall accuracy by 1.92 points over RAPS on OSCC and 0.14 points on ISIC. The margin is narrow on ISIC in overall accuracy but substantially wider in macro F1, where the proposed score reaches 75.33\% against 71.64\% for RAPS, indicating that the gain is concentrated in the smaller lesion categories rather than distributed uniformly. Since the redistribution rule is identical in every row, these differences are attributable solely to how the prediction sets are constructed, which indicates that the entropy-modulated margin weighting places the correct alternative class within the retained set more reliably than cumulative probability mass alone.

\begin{figure}[!t]
	\centering
	\includegraphics[width=0.7\linewidth]{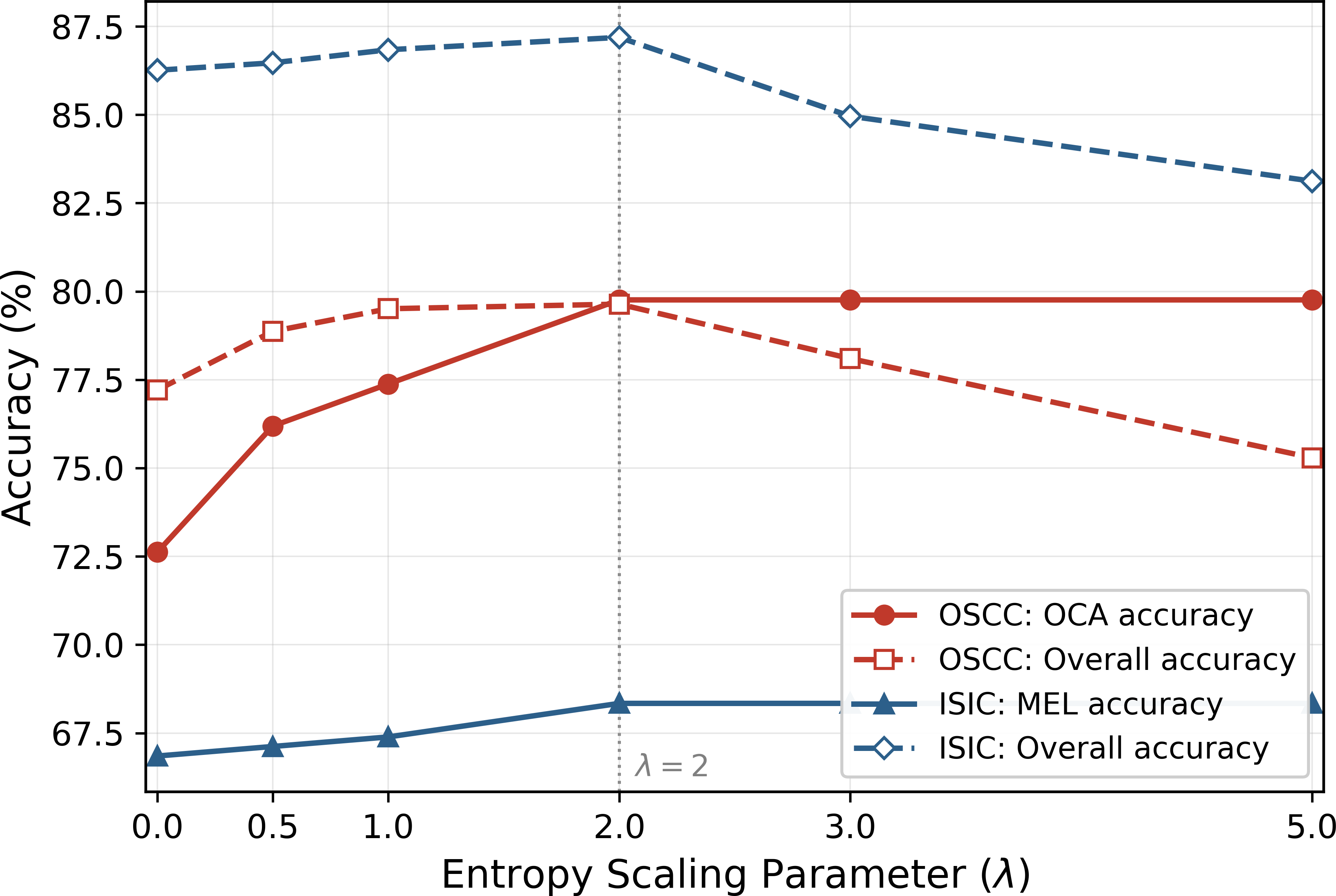}
	\caption{Effect of the entropy scaling parameter $\lambda$ on
		malignant-class and overall classification accuracy for the OSCC and ISIC
		datasets under the DF+MLP backbone at $\alpha = 0.2$.}
	\label{fig:lambda_sensitivity}
\end{figure}

The scaling hyperparameter $\lambda$ in~\eqref{eq:gamma} controls the influence
of the margin term relative to the cumulative probability mass. Its effect is
examined by varying $\lambda$ over the range $[0, 5]$ while holding the
backbone, data partitions, calibration set, and miscoverage level fixed, as
shown in Fig.~\ref{fig:lambda_sensitivity}. At $\lambda = 0$ the score reduces
to standard cumulative-mass accumulation, and the influence of the margin term
increases with $\lambda$. Larger values raise the calibrated threshold and
therefore yield larger prediction sets, which increases the number of samples
eligible for redistribution. Malignant-class accuracy rises with $\lambda$ on both datasets, from $72.62\%$
to $79.76\%$ for OCA and from $66.85\%$ to $68.34\%$ for MEL, and saturates
beyond $\lambda = 2$. This plateau arises because redistribution reassigns a
sample to the malignant class only when that class is ranked immediately below
the transitional label; once every such sample is associated with a multi-label
set, further enlargement admits only reassignments toward other classes.
Overall accuracy peaks at the same value, reaching $79.64\%$ on OSCC and
$87.19\%$ on ISIC, and declines thereafter as an increasing number of genuinely
transitional samples are redistributed away. A value of $\lambda = 2$ is
therefore adopted in all reported experiments, selected on the validation set
by maximising malignant-class accuracy with ties resolved by overall accuracy.

Table~\ref{tab:conformal_ablation} reports a component-wise ablation under a fixed backbone. On OSCC, generative augmentation raises accuracy from $72.50\%$ to $75.67\%$, APS-based redistribution adds a further 1.54 points, and the entropy-modulated margin weighting contributes an additional 2.43 points. On ISIC, where no augmentation is applied, APS-based redistribution improves accuracy from 85.83\% to 86.26\%, and the proposed weighting adds a further 0.93 points. On both datasets the margin-aware component therefore contributes a larger increment than redistribution over APS sets alone, which indicates that the weighting is not a neutral reparameterisation of APS but materially alters which samples become eligible for reassignment.


\begin{table}[!t]
	\caption{Component-wise ablation under a fixed DermFoundation backbone,
		reported as overall accuracy (\%). Augmentation applies to OSCC only.
		Redistribution is applied in the two rightmost configurations. OSCC uses the
		4-class setting, ISIC the 7-class setting.}
	\label{tab:conformal_ablation}
	\centering
	\begin{tabular}{|C{1.0cm}|C{1.2cm}|C{0.8cm}|C{0.8cm}|C{2.4cm}|}
		\hline
		Dataset & DF+MLP & +\,Aug. & +\,APS &
		\makecell{+\,Entropy-margin\\(AdaConRed)} \\
		\hline
		OSCC & 72.50 & 75.67 & 77.21 & \textbf{79.64} \\
		\hline
		ISIC & 85.83 & \textemdash & 86.26 & \textbf{87.19} \\
		\hline
	\end{tabular}
\end{table}

\subsection{State-of-the-Art Comparison}
\label{subsec:sota}
\begin{table*}[!t]
	\renewcommand{\thetable}{6a}
	\caption{Comparative study with state-of-the-art methods reported in the literature on the 4-class OSCC dataset}
	\vspace{2mm}
	\label{tab:4class_comparison}
	\centering
	\small
	\begin{tabular}{|l|c|c|c|c|c|}
		\hline
		{Framework} & {Embedding} & {Acc.} & {F1} & {Spec.} & {AUC} \\
		\hline
		BEiT3~\cite{wang2023image} & ViT-G & 71.24 & 69.37 & 72.56 & 76.31 \\
		\hline
		PubMedCLIP~\cite{eslami2023pubmedclip} & ViT-B/32 & 67.13 & 65.42 & 68.35 & 72.46 \\
		\hline
		BMCA-CLIP~\cite{lozano2025biomedica} & ViT-L/14 & 73.67 & 71.94 & 74.82 & 78.42 \\
		\hline
		ViP~\cite{fang2024aligning} & ViT-B/16 & 72.44 & 70.63 & 73.58 & 77.17 \\
		\hline
		MAVL~\cite{phan2024decomposing} & ResNet-50 & 74.42 & 72.84 & 75.53 & 79.13 \\
		\hline
		eCLIP~\cite{kumar2024improving} & Swin-T & 73.61 & 71.87 & 74.73 & 78.33 \\
		\hline
		RET-CLIP~\cite{du2024ret} & ViT-B & 72.46 & 70.68 & 73.62 & 77.25 \\
		\hline
		MSL+MLP & MedSigLIP & 63.20 & 60.77 & 86.71 & 84.53 \\
		\hline
		MSL+MLP (Aug.) & MedSigLIP & 72.73 & 71.03 & 90.12 & 88.10 \\
		\hline
		MSL+MLP+AdaConRed (Aug.) & MedSigLIP & 75.80 & 73.79 & 91.95 & 88.10 \\
		\hline
		DF+MLP & DermFoundation & 72.50 & 68.51 & 89.79 & 85.70 \\
		\hline
		DF+MLP (Aug.) & DermFoundation & 75.67 & 73.19 & 91.12 & \textbf{88.12} \\
		\hline
		DF+MLP+AdaConRed (Aug.) & DermFoundation & \textbf{79.64} & \textbf{77.18} & \textbf{93.79} & \textbf{88.12} \\
		\hline
	\end{tabular}
\end{table*}
\setcounter{table}{6}
\renewcommand{\thetable}{\arabic{table}}

\begin{table*}[!t]
	\renewcommand{\thetable}{6b}
	\caption{\small Comparative study with state-of-the-art methods reported in the literature on the multi-class ISIC dataset}
	\vspace{2mm}
	\label{tab:7class_comparison}
	\centering
	\small
	\begin{tabular}{|p{3.5cm}|c|p{3cm}|c|c|}
		\hline
		{Framework} & {Segmentation} & {Classes} & {Acc.} & {F1} \\
		\hline
		DenseNet201-Cubic SVM~\cite{benyahia2022multi} & $\times$ & MEL, NV, BCC, AK, BKL, DF, VASC, SCC & 80.87 & - \\
		\hline
		ResNet18+VGG19-SVM~\cite{shakya2025comprehensive} & \checkmark & BEN, MEL & 89.09 & 90.16 \\
		\hline
		DSCCNet~\cite{tahir2023dscc_net} & $\times$ & BCC, MEL, NV, SCC & 83.20 & 58.09 \\
		\hline
		ResNet50-Inception~\cite{gouda2022detection} & $\times$ & BEN, MEL & 84.10 & - \\
		\hline
		UNet~\cite{kim2021computer} & $\times$ & BEN, MEL & 79.90 & - \\
		\hline
		GLCM-SVM+RF~\cite{murugan2021diagnosis} & \checkmark & SK, BCC, MEL & 89.31 & 44.09 \\
		\hline
		MSL+MLP & $\times$ & AK, BCC, BKL, MEL, NV, SCC, UNK & 85.78 & 73.22\\
		\hline
		MSL+MLP+AdaConRed & $\times$ & AK, BCC, BKL, MEL, NV, SCC, UNK & 85.82 & 74.38\\
		\hline
		DF+MLP & $\times$ & AK, BCC, BKL, MEL, NV, SCC, UNK & 85.83 & 73.46 \\
		\hline
		DF+MLP+AdaConRed & $\times$ & AK, BCC, BKL, MEL, NV, SCC, UNK & 87.19 & 75.33\\
		\hline
	\end{tabular}
	
	\vspace{1mm}
	\raggedright
	\scriptsize \textit{Abbreviations:} AK: Actinic Keratosis, BCC: Basal Cell Carcinoma, BEN: Benign, BKL: Benign Keratosis, DF: Dermatofibroma, MEL: Melanoma, NV: Melanocytic Nevus, SCC: Squamous Cell Carcinoma, SK: Seborrheic Keratosis, VASC: Vascular Lesion, UNK: Unknown.
\end{table*}
\setcounter{table}{6}
\renewcommand{\thetable}{\arabic{table}}

Table~\ref{tab:4class_comparison} presents a comparative study against several state-of-the-art methods reported in the literature on the OSCC dataset~\cite{dai2026prompt}. The 4-class configuration is considered here to ensure fair comparison with previously reported studies. Notably, the proposed framework consistently improves performance across both embedding backbones. AdaConRed raises the overall accuracy from 72.73\% to 75.80\% for MedSigLIP, and from 75.67\% to 79.64\% for DermFoundation. The latter exceeds the strongest comparator in Table~\ref{tab:4class_comparison} in accuracy, MAVL at 74.42\%, by 5.22 points, with the augmented DermFoundation configuration already surpassing it by 1.25 points before redistribution is applied. Furthermore, in this 4-class setting, the OCA class accuracy improves from $61.90\%$ to $75.00\%$ for MedSigLIP and from $53.57\%$ to $79.76\%$ for DermFoundation. AUC values remain unchanged with AdaConRed, since redistribution reassigns discrete class labels without modifying the underlying softmax probabilities from which AUC is computed. The rank ordering of scores, and hence the ROC curve, is therefore invariant by construction. Similarly, Table~\ref{tab:7class_comparison} presents a comparative study with benchmark models on the ISIC dataset.



\subsection{Subgroup Reliability Analysis}

%
%
%

\begin{figure}[!t]
	\centering
	\begin{subfigure}[t]{0.48\linewidth}
		\centering
		\includegraphics[width=\linewidth]{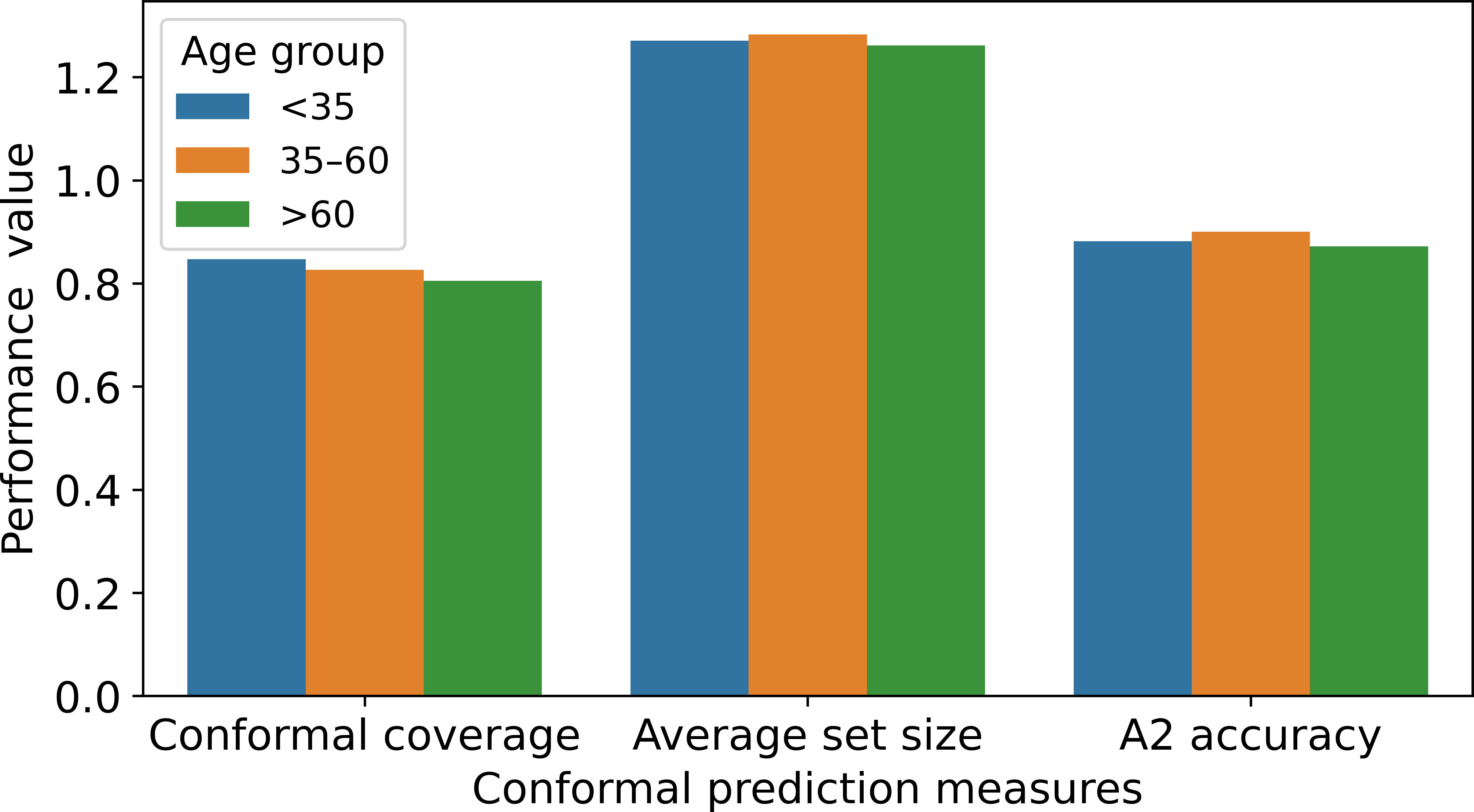}
		\caption{}
		\label{fig:oscc_fair_age}
	\end{subfigure}
	\hfill
	\begin{subfigure}[t]{0.48\linewidth}
		\centering
		\includegraphics[width=\linewidth]{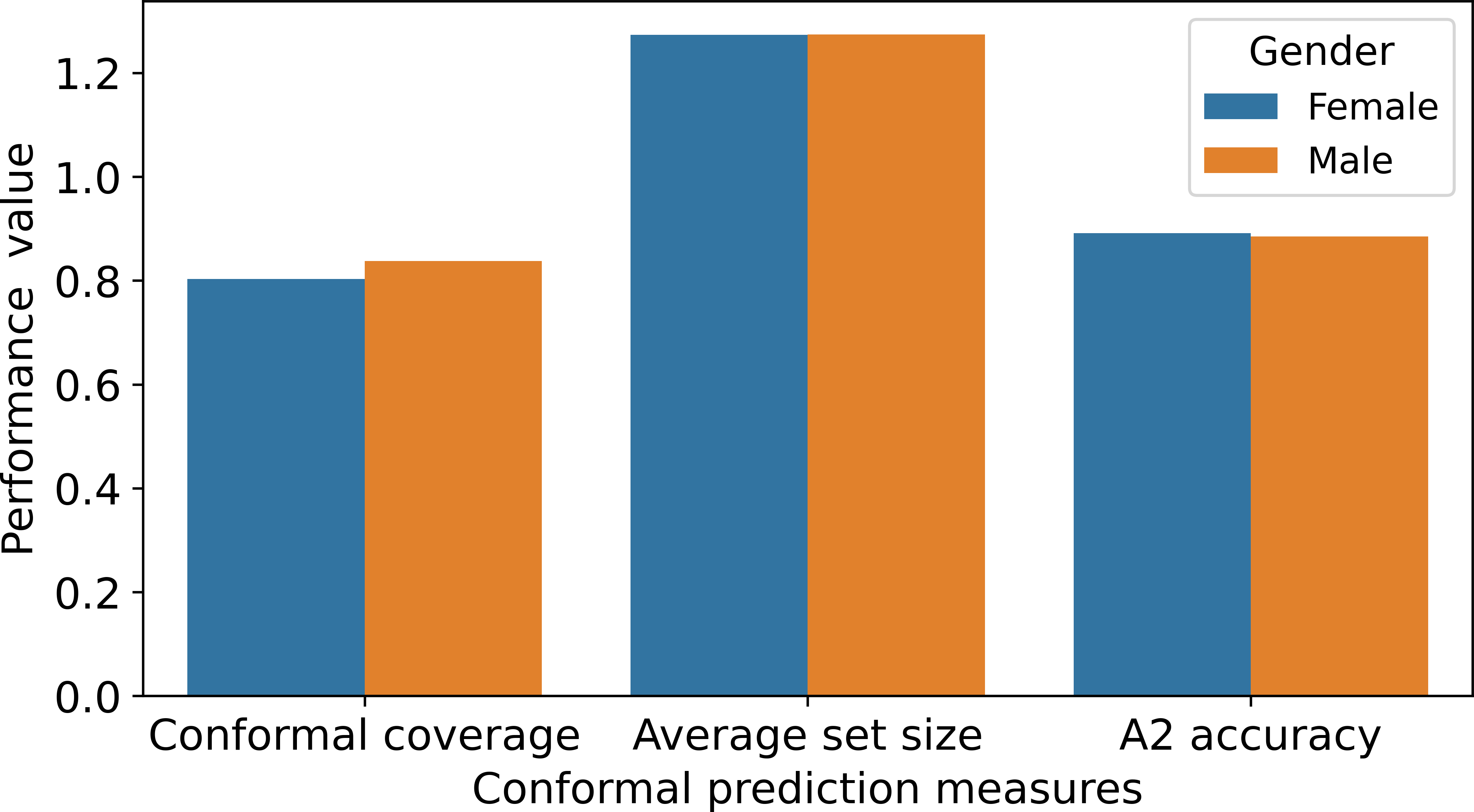}
		\caption{}
		\label{fig:oscc_fair_gender}
	\end{subfigure}
	\hfill
	\begin{subfigure}[t]{0.48\linewidth}
		\centering
		\includegraphics[width=\linewidth]{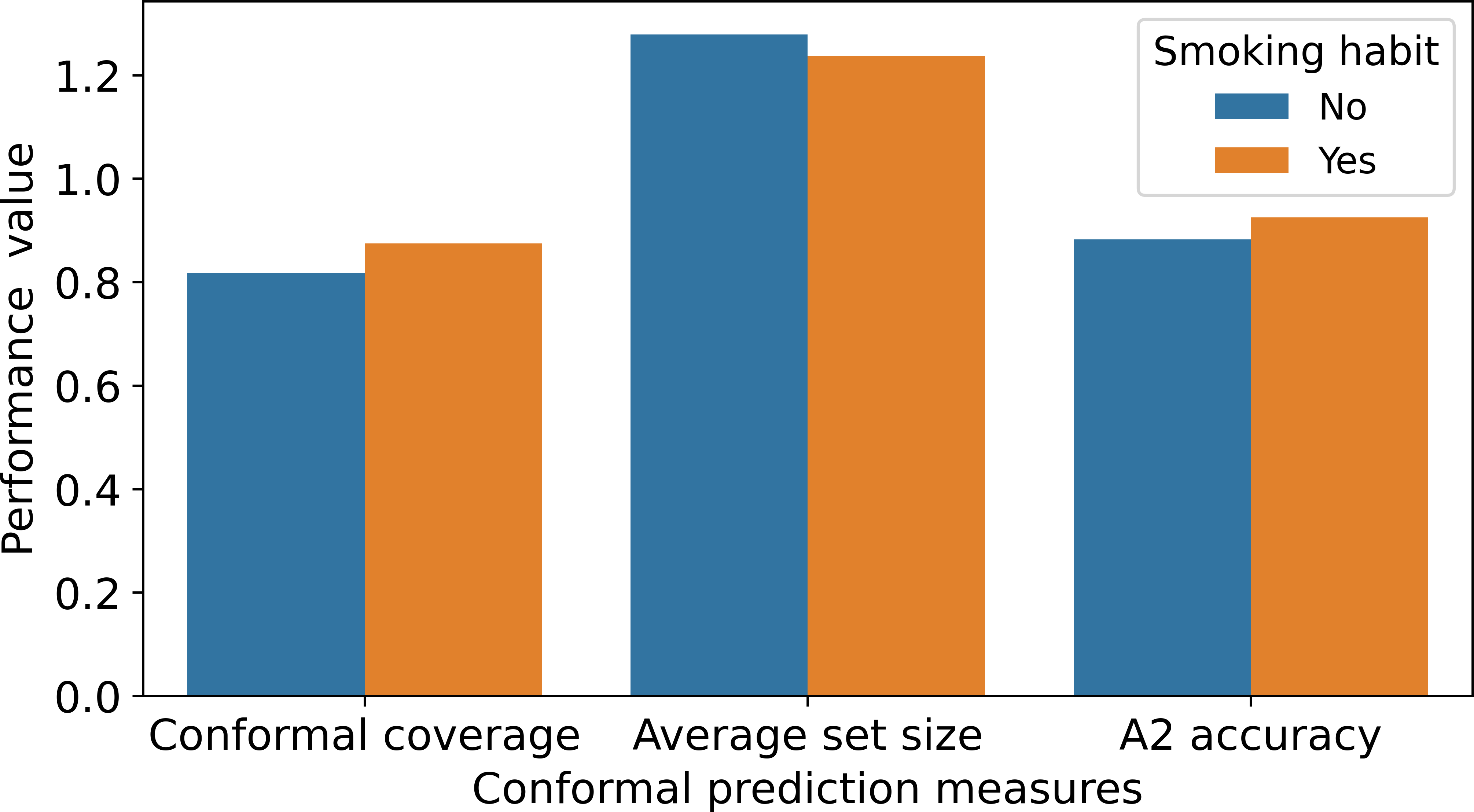}
		\caption{}
		\label{fig:oscc_fair_smoking}
	\end{subfigure}
	\hfill
	\begin{subfigure}[t]{0.48\linewidth}
		\centering
		\includegraphics[width=\linewidth]{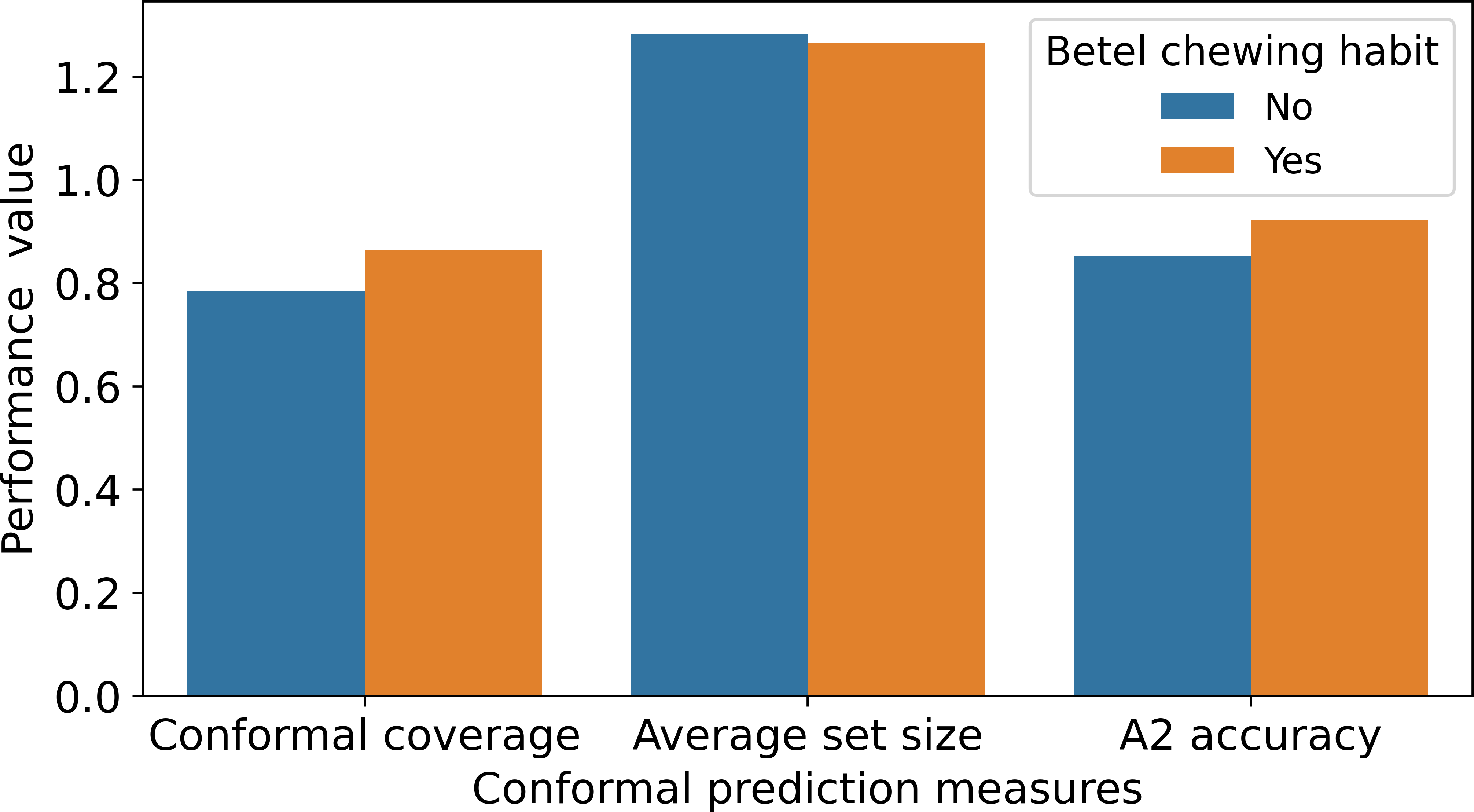}
		\caption{}
		\label{fig:oscc_fair_betel}
	\end{subfigure}
	\hfill
	\begin{subfigure}[t]{0.48\linewidth}
		\centering
		\includegraphics[width=\linewidth]{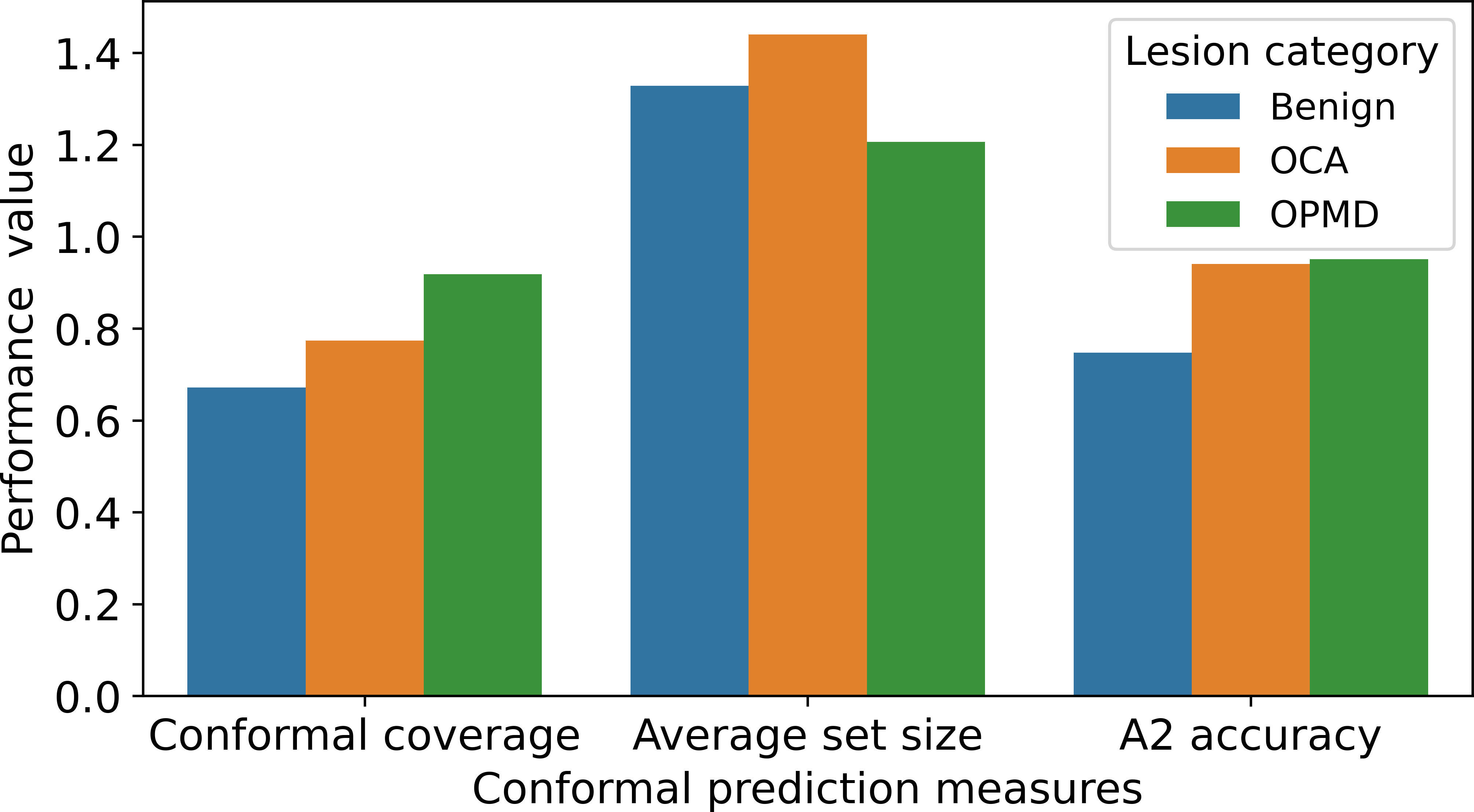}
		\caption{}
		\label{fig:oscc_fair_category}
	\end{subfigure}
	
	\caption{Subgroup-wise comparison of conformal coverage, average prediction set size, and A2 accuracy on the OSCC dataset across (a) age group, (b) gender, (c) smoking habit, (d) betel quid chewing habit, and (e) lesion category.}
	\label{fig:fairness_analysis}
\end{figure}

%
%
%
%

\begin{figure}[!t]
	\centering
	\begin{subfigure}[t]{0.48\linewidth}
		\centering
		\includegraphics[width=\linewidth]{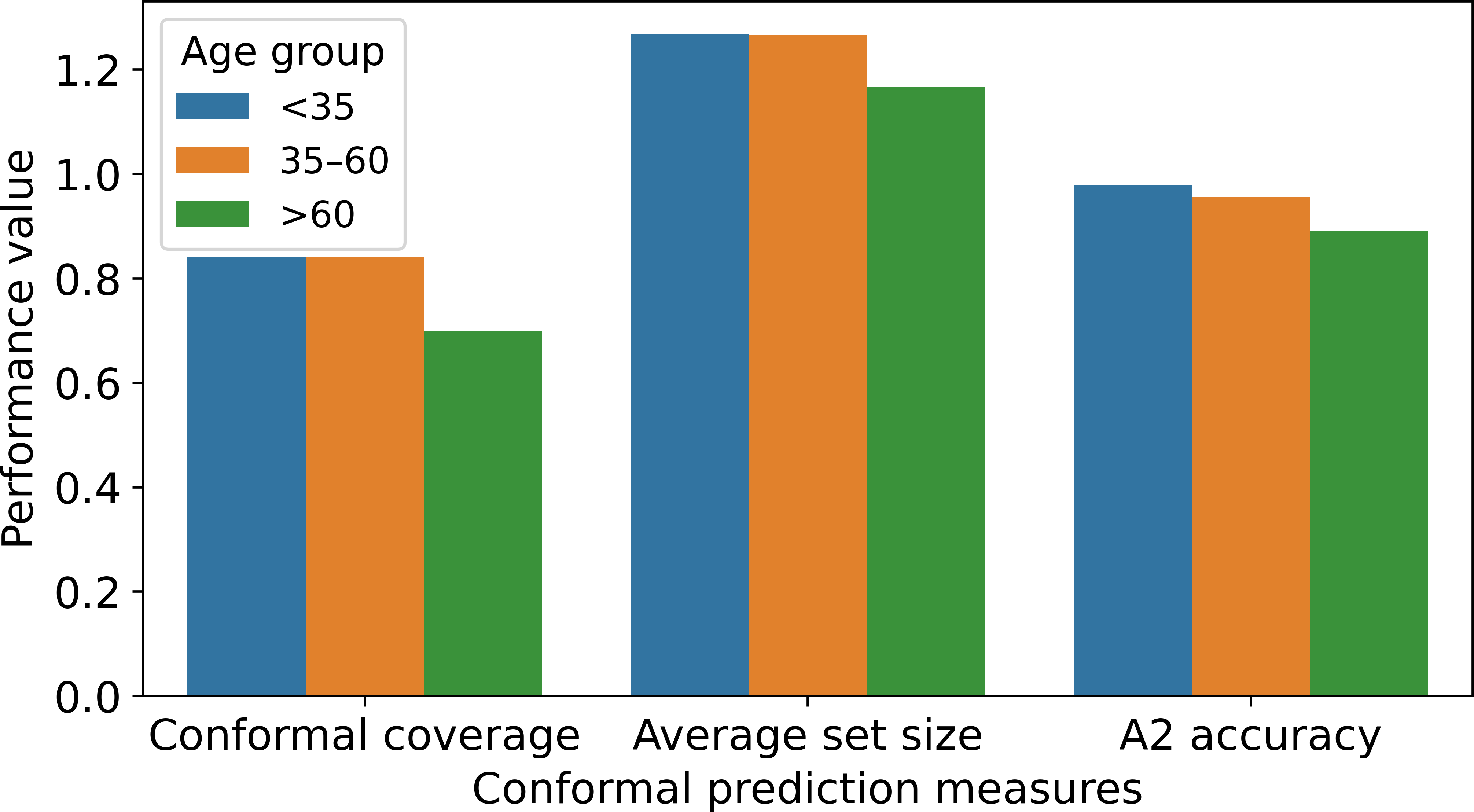}
		\caption{}
		\label{fig:isic_fair_age}
	\end{subfigure}
	\hfill
	\begin{subfigure}[t]{0.48\linewidth}
		\centering
		\includegraphics[width=\linewidth]{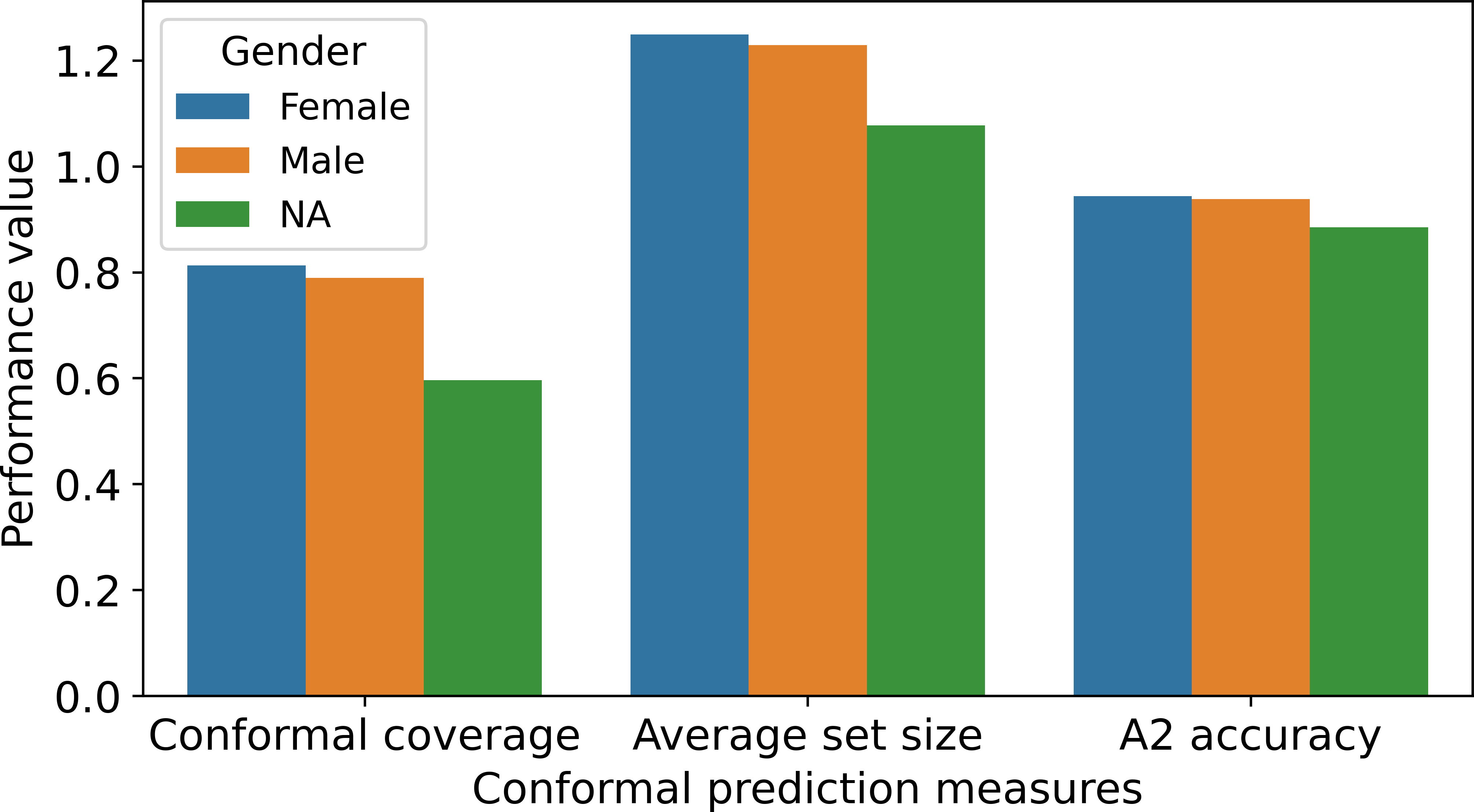}
		\caption{}
		\label{fig:isic_fair_gender}
	\end{subfigure}
	
	\vspace{2mm}
	
	\begin{subfigure}[t]{0.48\linewidth}
		\centering
		\includegraphics[width=\linewidth]{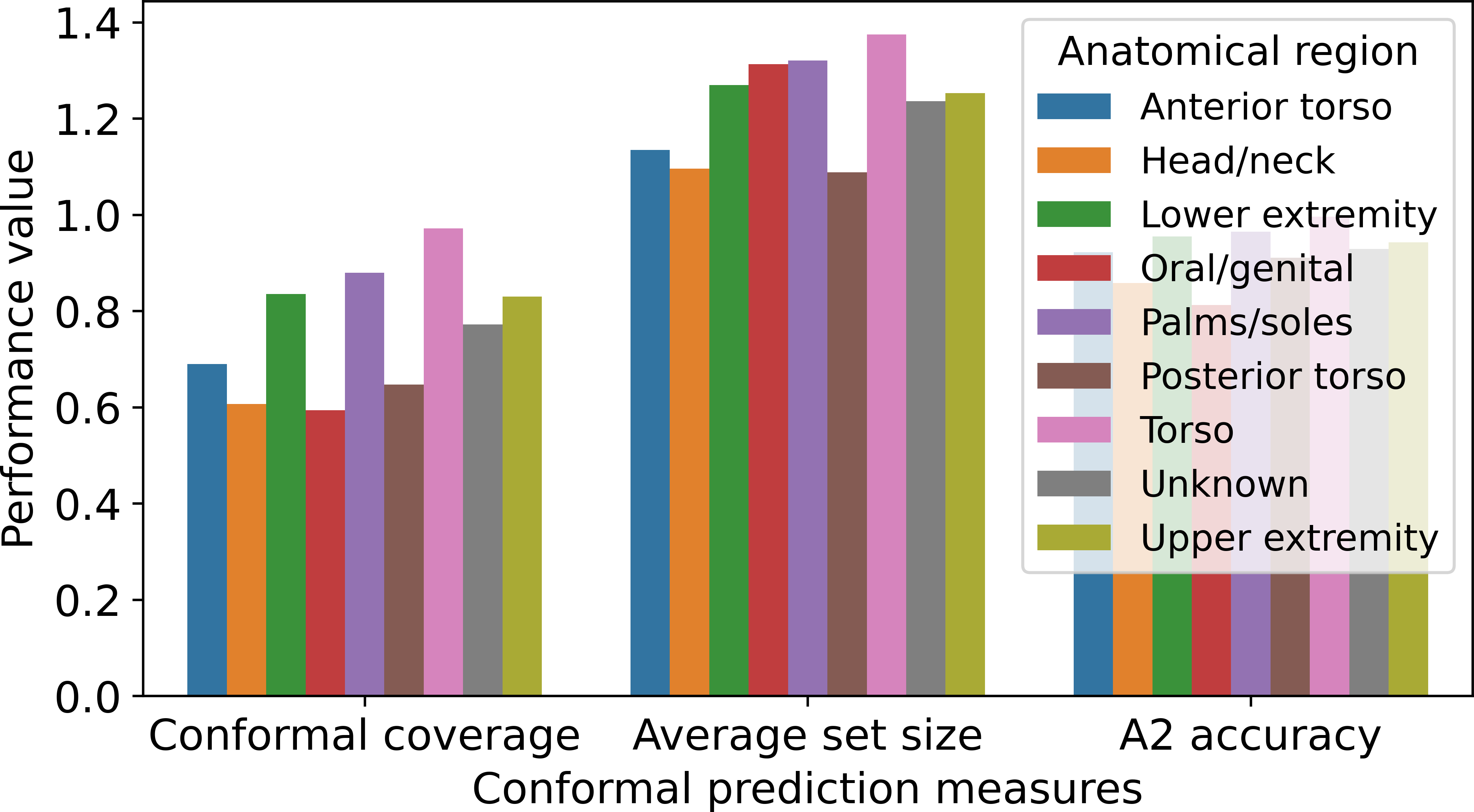}
		\caption{}
		\label{fig:isic_fair_region}
	\end{subfigure}
	\hfill
	\begin{subfigure}[t]{0.48\linewidth}
		\centering
		\includegraphics[width=\linewidth]{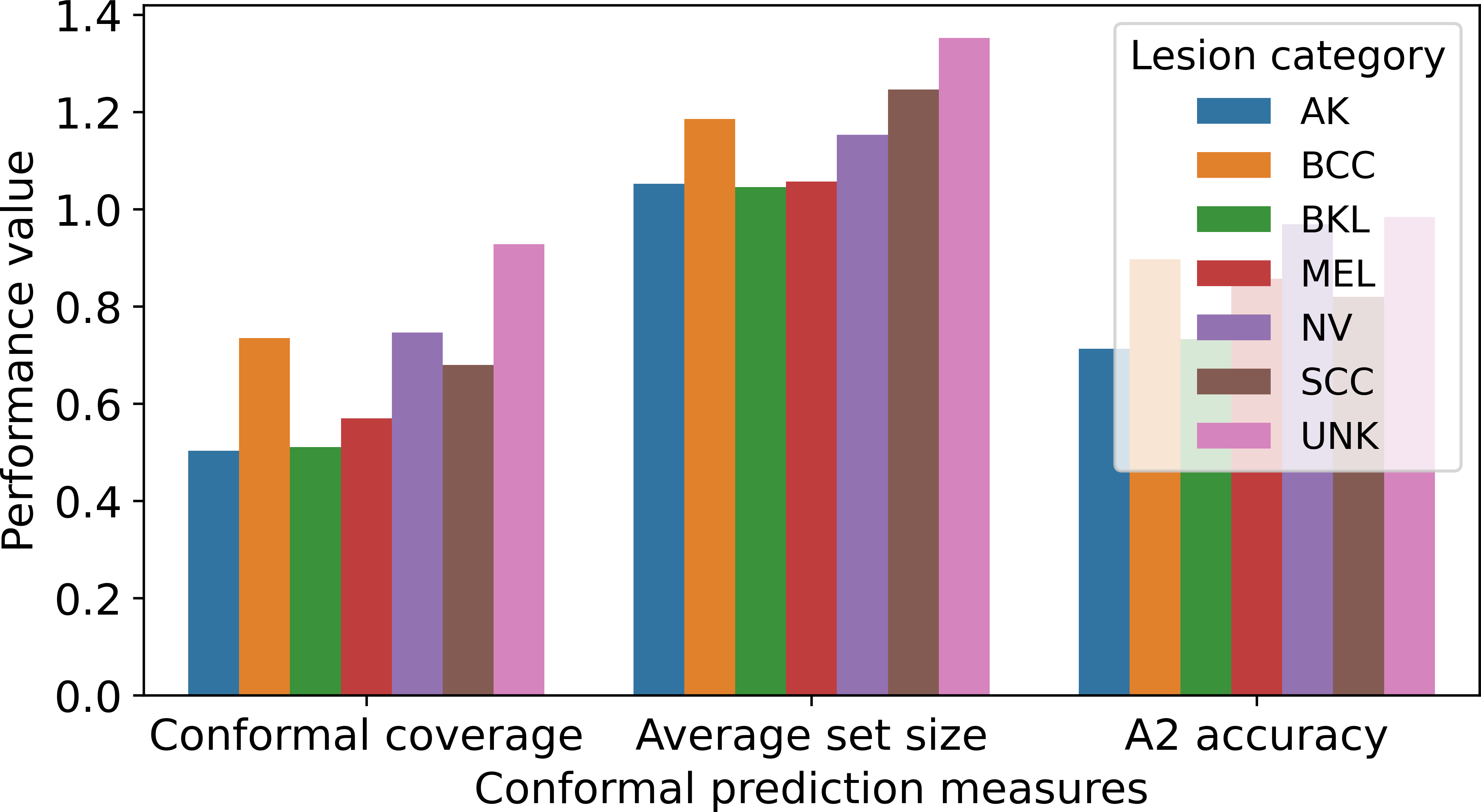}
		\caption{}
		\label{fig:isic_fair_category}
	\end{subfigure}
	
	\caption{Subgroup-wise comparison of conformal coverage, average prediction set size, and A2 accuracy on the ISIC dataset across (a) age group, (b) gender, (c) anatomical region, and (d) lesion category.}
	\label{fig:fairness_analysis_isic}
\end{figure}

A descriptive subgroup reliability analysis is performed on the OSCC dataset across strata defined by \textit{age}, \textit{gender}, \textit{smoking habit}, \textit{betel quid chewing habit}, and \textit{lesion category}, as shown in Fig.~\ref{fig:fairness_analysis}. The reported values are point estimates without confidence intervals or significance testing, and are therefore intended to characterize subgroup-dependent uncertainty behaviour rather than to establish formal fairness. A2 accuracy is defined as the proportion of test samples for which the GT label is contained within the two most probable predicted labels. The results indicate that conformal coverage remains relatively stable across all subgroups, showing its distribution-free behavior. However, variations in average prediction set size and A2 accuracy reveal subgroup-dependent differences in uncertainty and model confidence across clinically relevant populations. In particular, risk-associated groups (e.g., smoking and betel quid users) exhibit slightly improved A2 accuracy, suggesting more distinctive lesion characteristics. 
Lesion category-wise analysis highlights increased uncertainty for OCA, with the maximum average conformal set size of $1.44$, likely reflecting stronger overlap with neighboring lesion categories. 

A similar subgroup reliability analysis is performed on the ISIC dataset across \textit{age}, \textit{gender}, \textit{anatomical region}, and \textit{lesion category}, as illustrated in Fig.~\ref{fig:fairness_analysis_isic}. Similar to the OSCC observations, variations in average prediction set size and A2 accuracy reveal subgroup-specific differences in uncertainty and model confidence. 
Age-wise analysis indicates slightly lower performance in the $>60$ group, where both coverage and A2 accuracy decrease compared with younger populations. Anatomical region-wise analysis reveals notable variability, with the torso region showing higher coverage and A2 accuracy, and the head/neck regions showing lower values.

\begin{figure}[!t]
	\centering
	\begin{subfigure}[t]{0.49\linewidth}
		\centering
		\includegraphics[width=\linewidth]{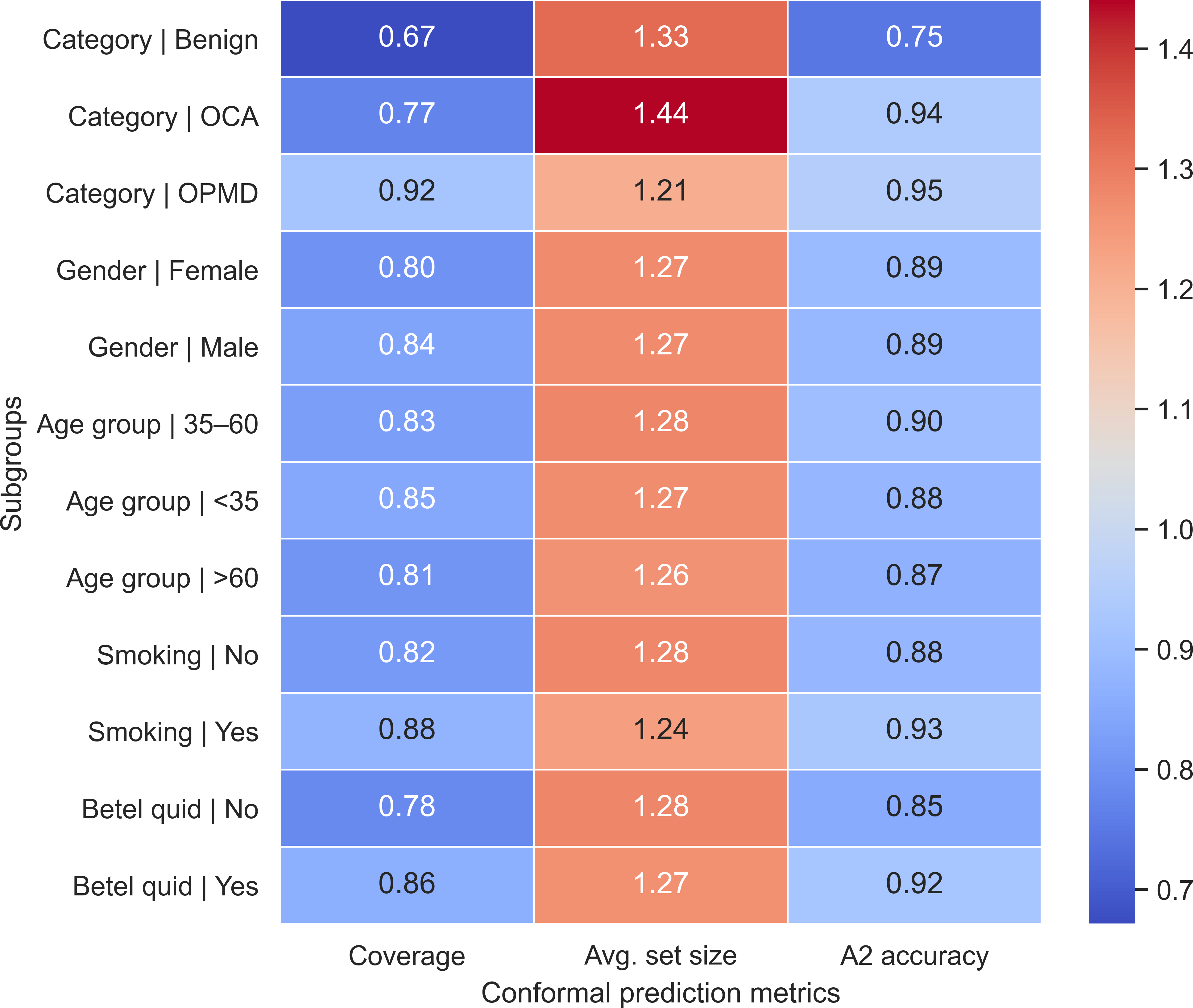}
		\caption{}
		\label{fig:fairness_heatmap_oscc}
	\end{subfigure}
	\begin{subfigure}[t]{0.49\linewidth}
		\centering
		\includegraphics[width=\linewidth]{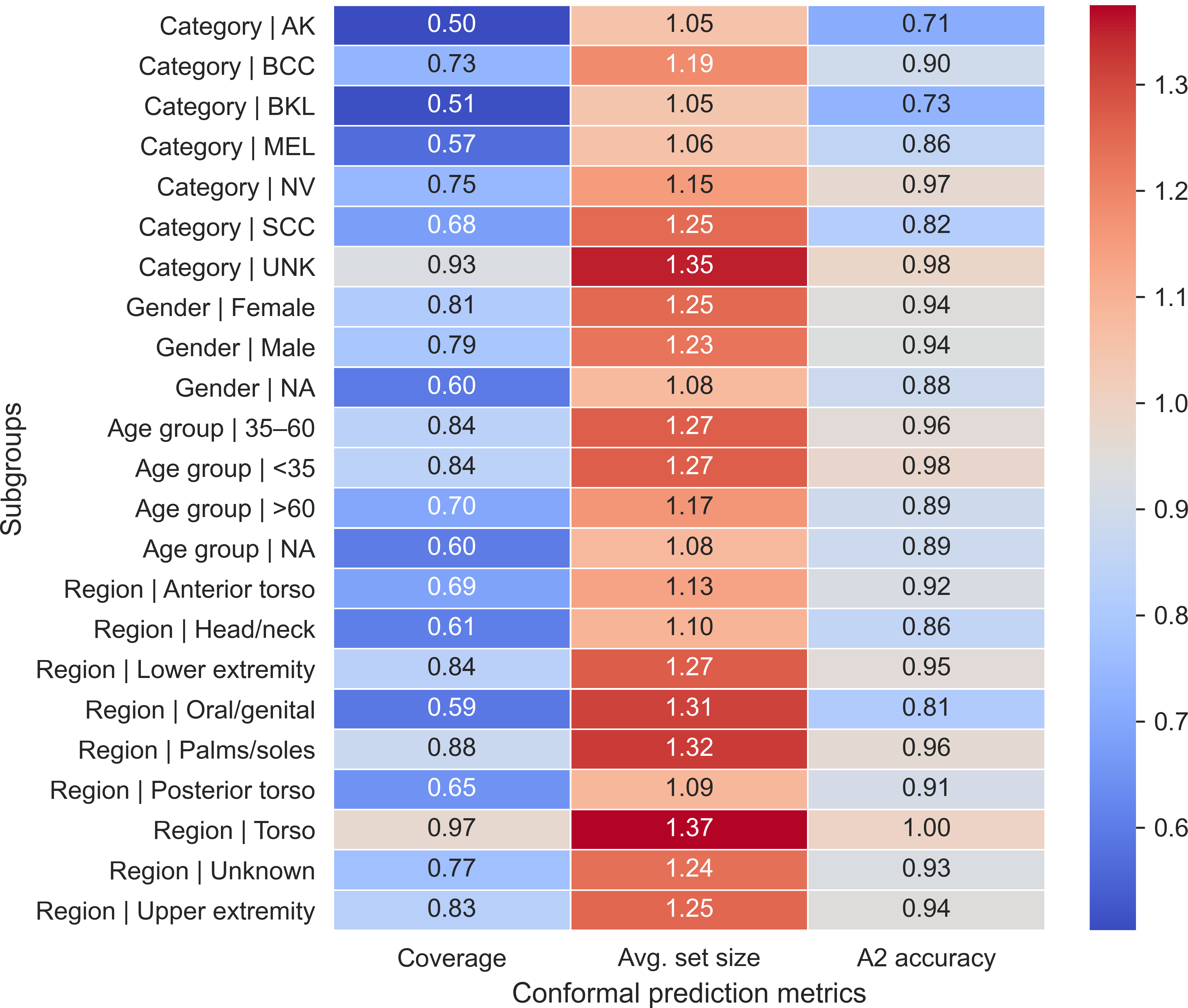}
		\caption{}
		\label{fig:fairness_heatmap_isic}
	\end{subfigure}
	
	\caption{Comprehensive heatmap visualization of subgroup-wise conformal prediction metrics across the (a) OSCC and (b) ISIC datasets.}
	\label{fig:fairness_heatmap}
\end{figure}

The subgroup-level behaviour summarised above is examined further through consolidated visualisations. Fig.~\hyperlink{fig:fairness_heatmap_oscc}{\ref{fig:fairness_heatmap}a} presents a heatmap of the conformal metrics across all OSCC strata. The visualization reiterates that coverage remains comparatively stable across most subgroups, whereas variations in prediction set size and A2 accuracy reveal subgroup-dependent differences in uncertainty and model confidence. In particular, lesion category and risk-associated strata exhibit relatively larger variations, indicating increased ambiguity and heterogeneous uncertainty behavior. Similarly, Fig.~\hyperlink{fig:fairness_heatmap_isic}{\ref{fig:fairness_heatmap}b} presents a consolidated heatmap visualization of subgroup-wise conformal metrics on the ISIC dataset. The heatmap reveals subgroup-dependent variations in uncertainty characteristics across lesion categories, demographic groups, and anatomical regions. In particular, larger variations are observed across lesion and anatomical region strata, suggesting heterogeneous uncertainty behavior and varying levels of ambiguity across different skin lesion patterns.

\begin{figure}[!t]
	\centering
	\begin{subfigure}[t]{0.48\linewidth}
		\centering
		\includegraphics[width=\linewidth]{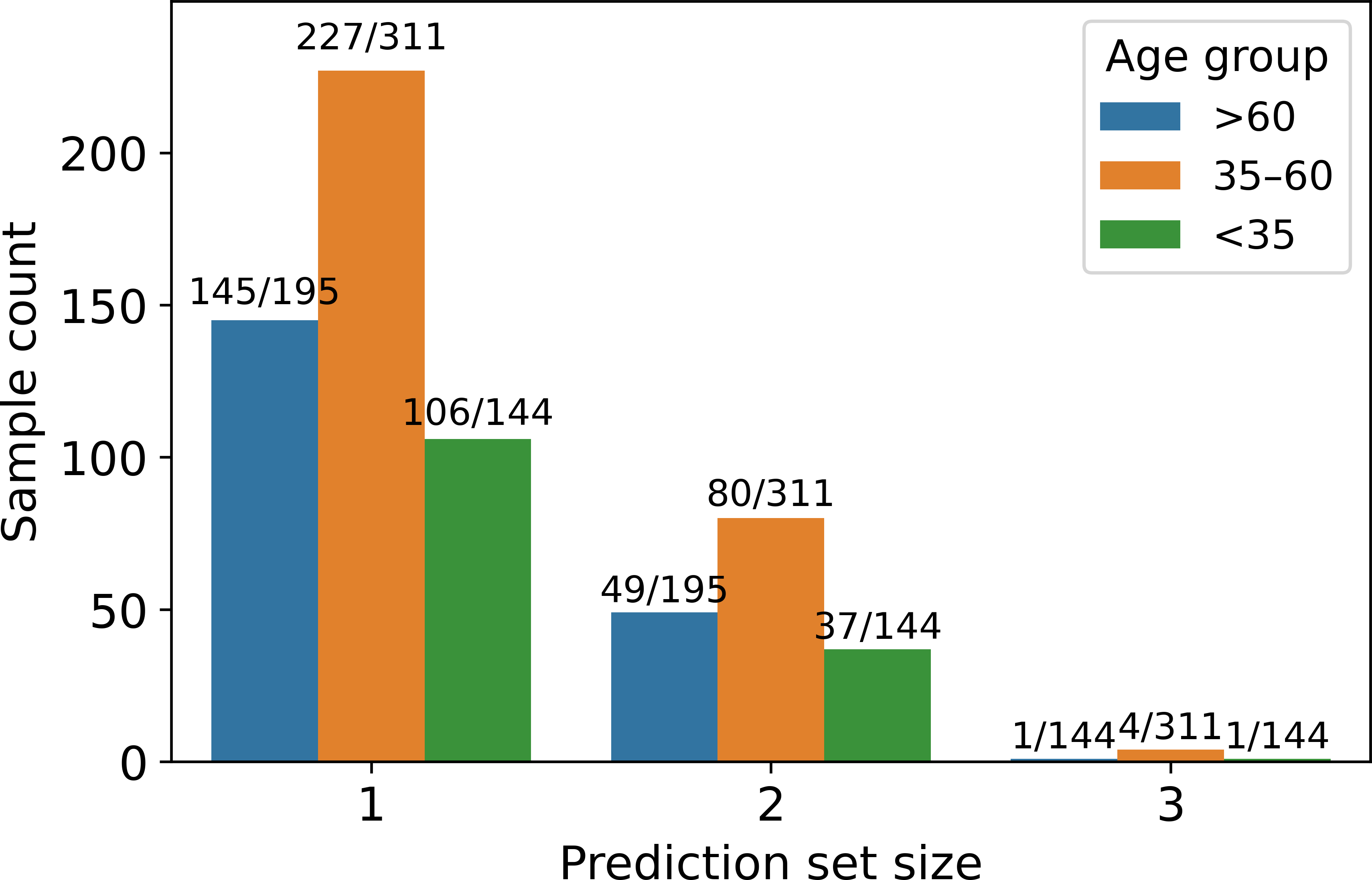}
		\caption{}
		\label{fig:setsize_oscc_age}
	\end{subfigure}
	\hfill
	\begin{subfigure}[t]{0.48\linewidth}
		\centering
		\includegraphics[width=\linewidth]{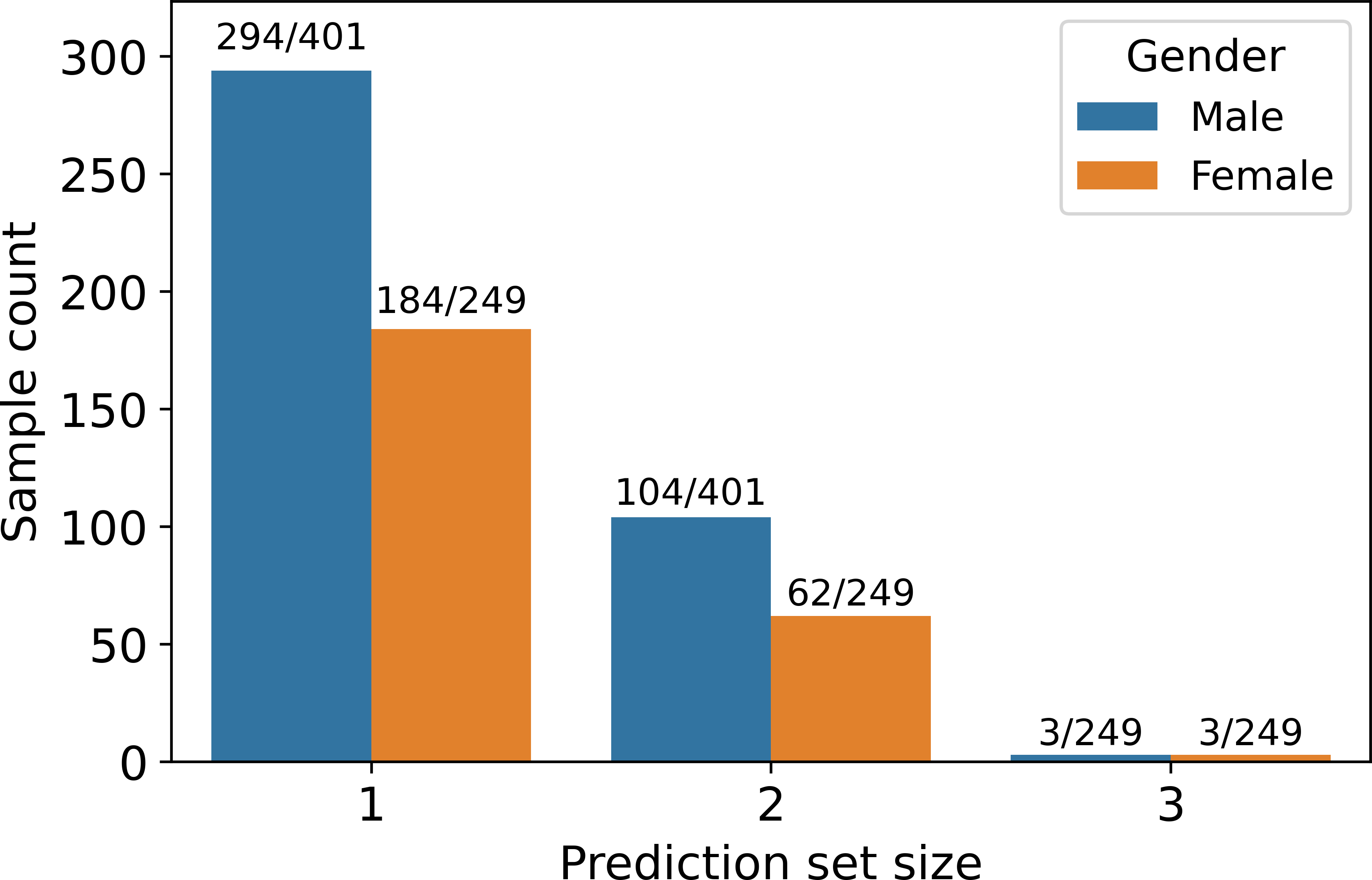}
		\caption{}
		\label{fig:setsize_oscc_gender}
	\end{subfigure}
	\hfill
	\begin{subfigure}[t]{0.48\linewidth}
		\centering
		\includegraphics[width=\linewidth]{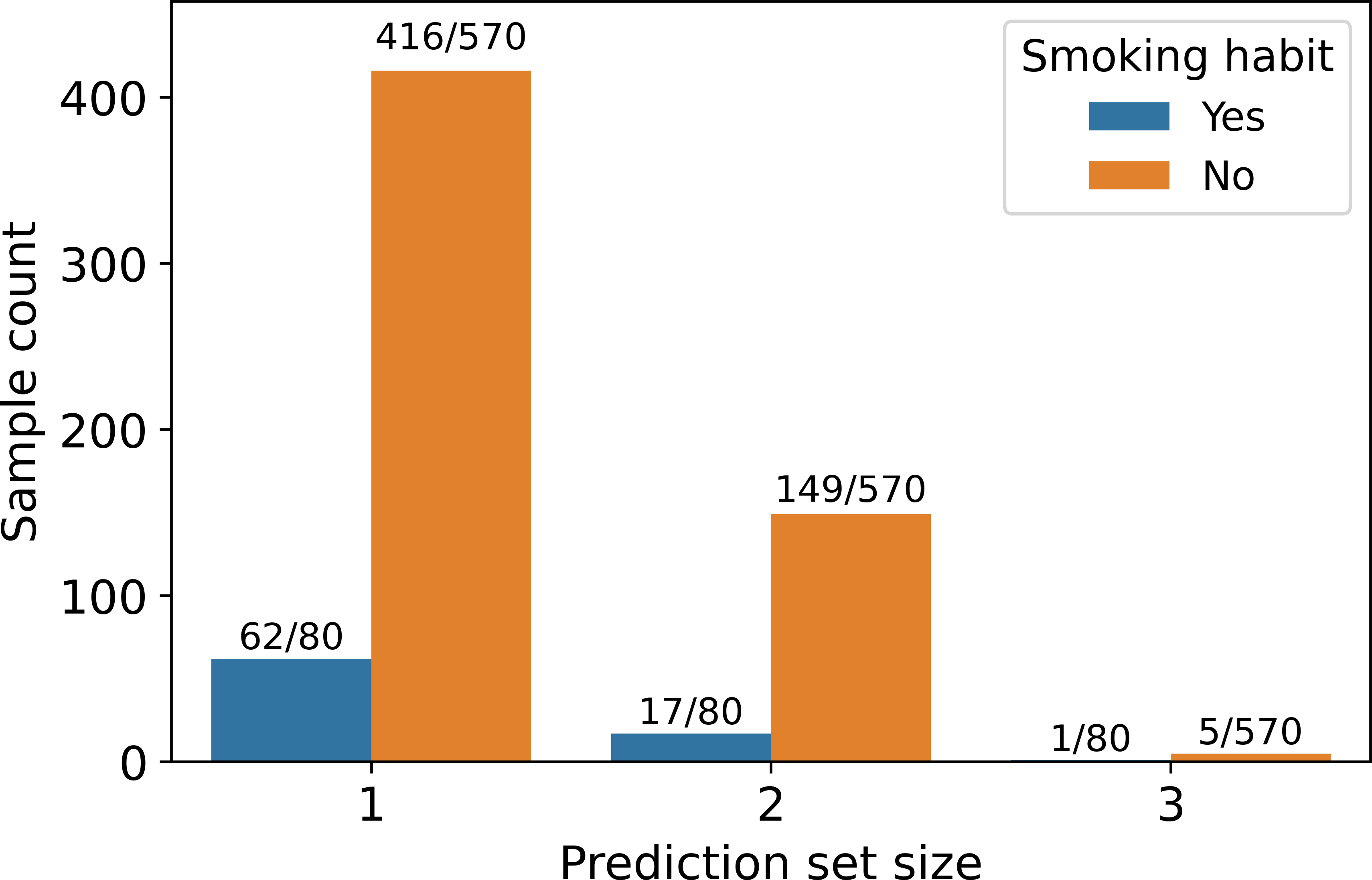}
		\caption{}
		\label{fig:setsize_oscc_smoking}
	\end{subfigure}
	\hfill
	\begin{subfigure}[t]{0.48\linewidth}
		\centering
		\includegraphics[width=\linewidth]{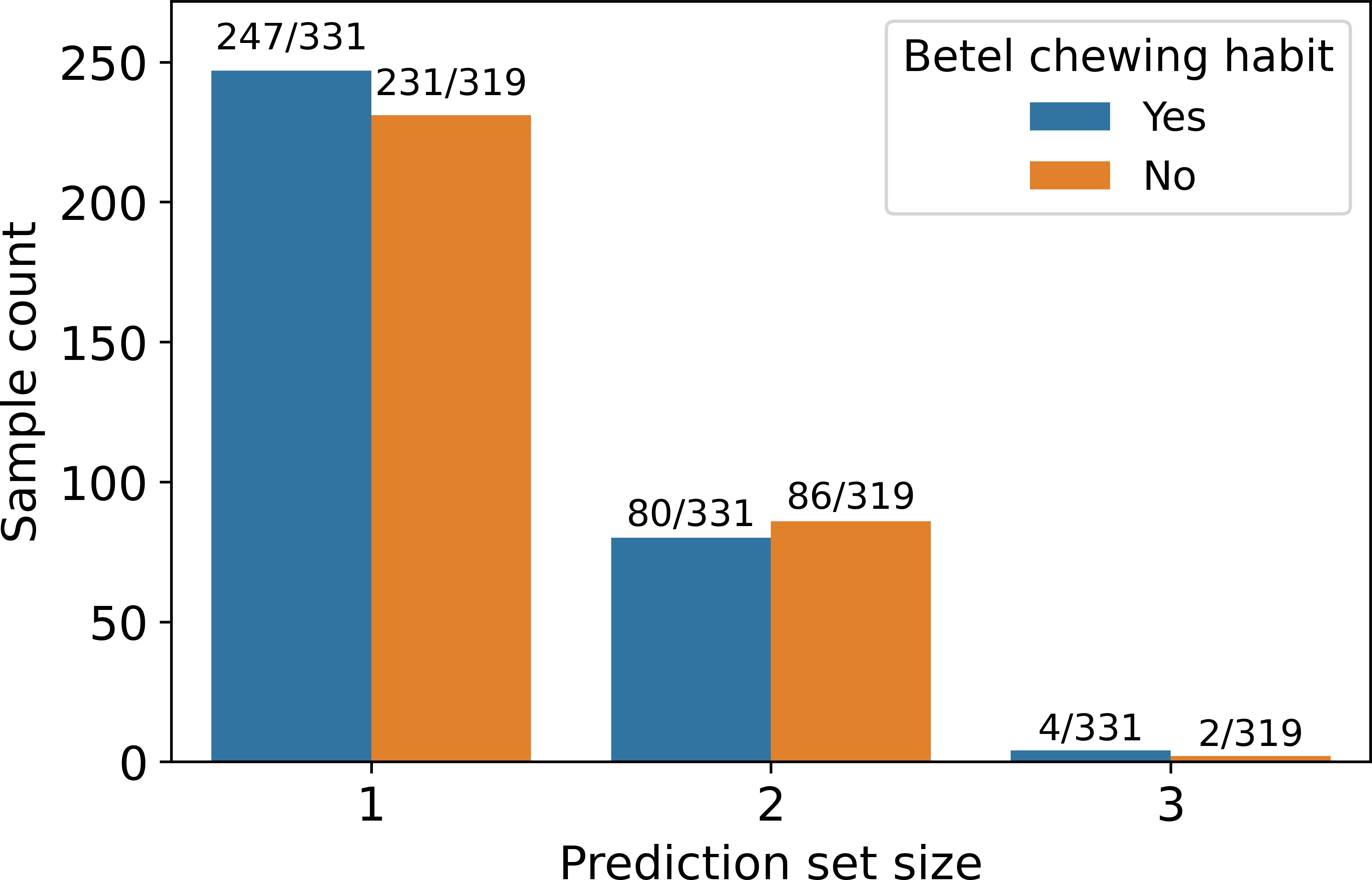}
		\caption{}
		\label{fig:setsize_oscc_betel}
	\end{subfigure}
	\hfill
	\begin{subfigure}[t]{0.48\linewidth}
		\centering
		\includegraphics[width=\linewidth]{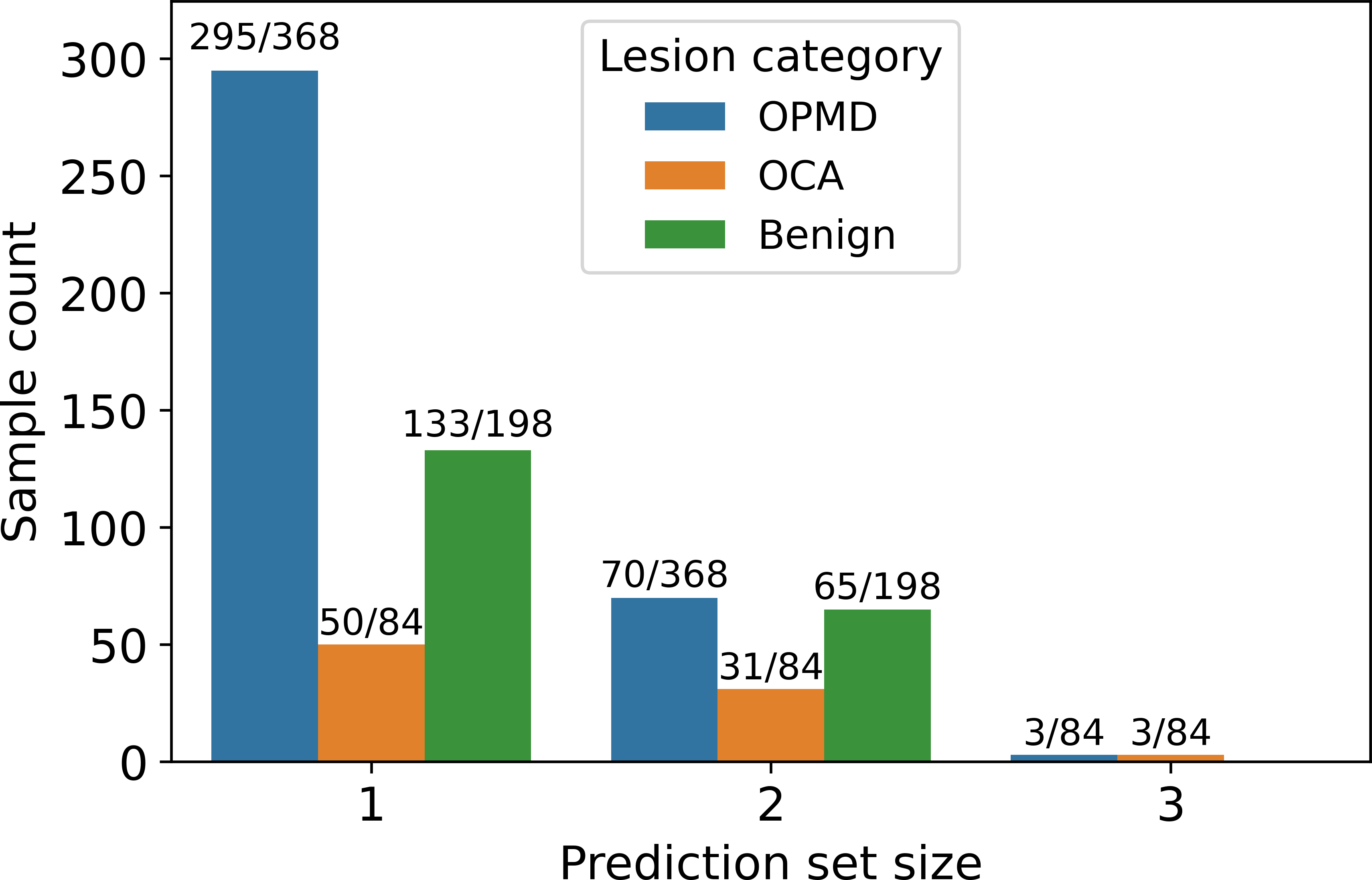}
		\caption{}
		\label{fig:setsize_oscc_category}
	\end{subfigure}
	\hfill
	\caption{Distribution of conformal prediction set cardinalities across different strata on the OSCC dataset across (a) age group, (b) gender, (c) smoking habit, (d) betel quid chewing habit, and (e) lesion category.}
	\label{fig:setsize_distribution}
\end{figure}

\begin{figure}[!t]
	\centering
	\begin{subfigure}[t]{0.48\linewidth}
		\centering
		\includegraphics[width=\linewidth]{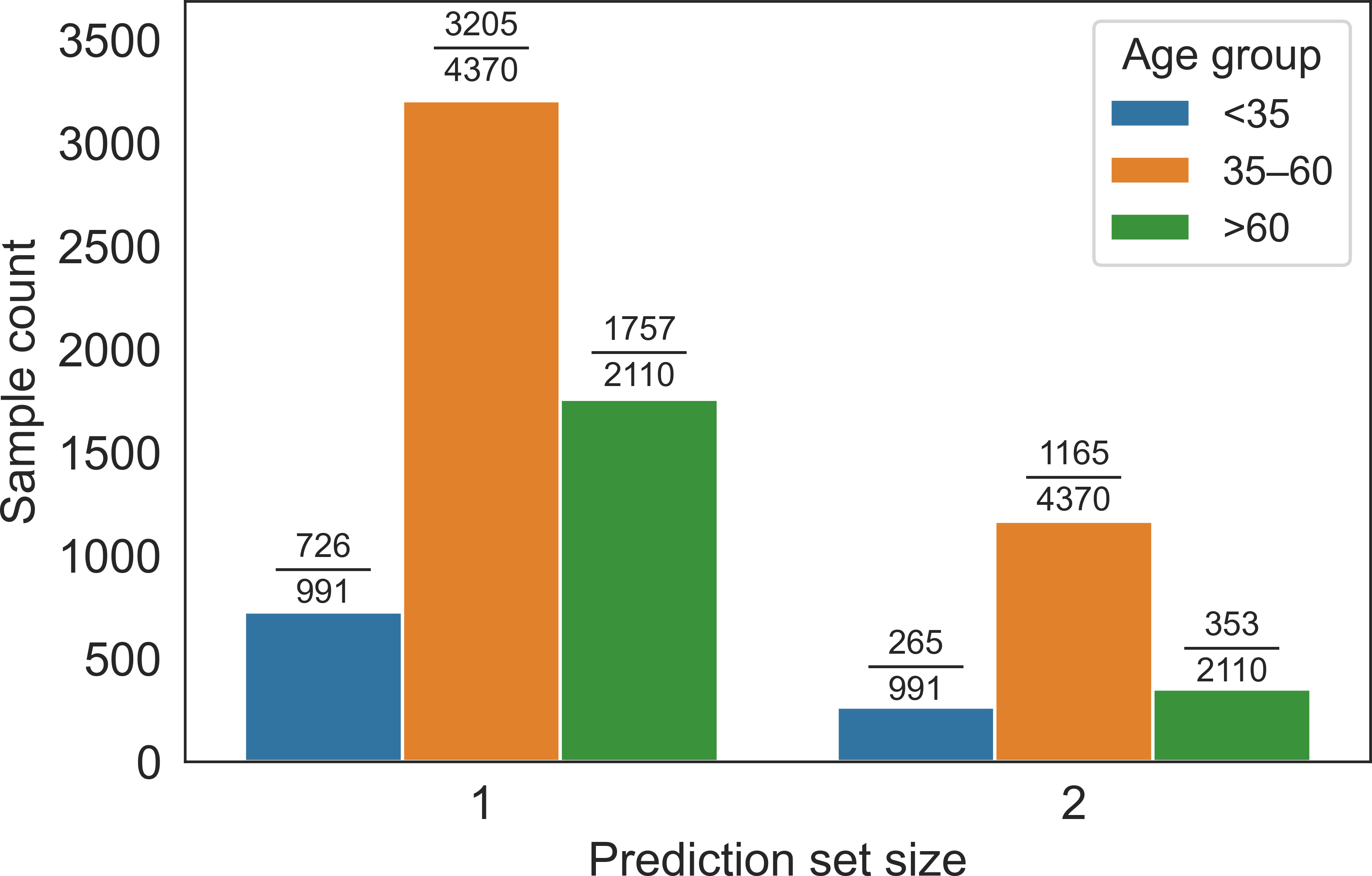}
		\caption{}
		\label{fig:setsize_isic_age}
	\end{subfigure}
	\hfill
	\begin{subfigure}[t]{0.48\linewidth}
		\centering
		\includegraphics[width=\linewidth]{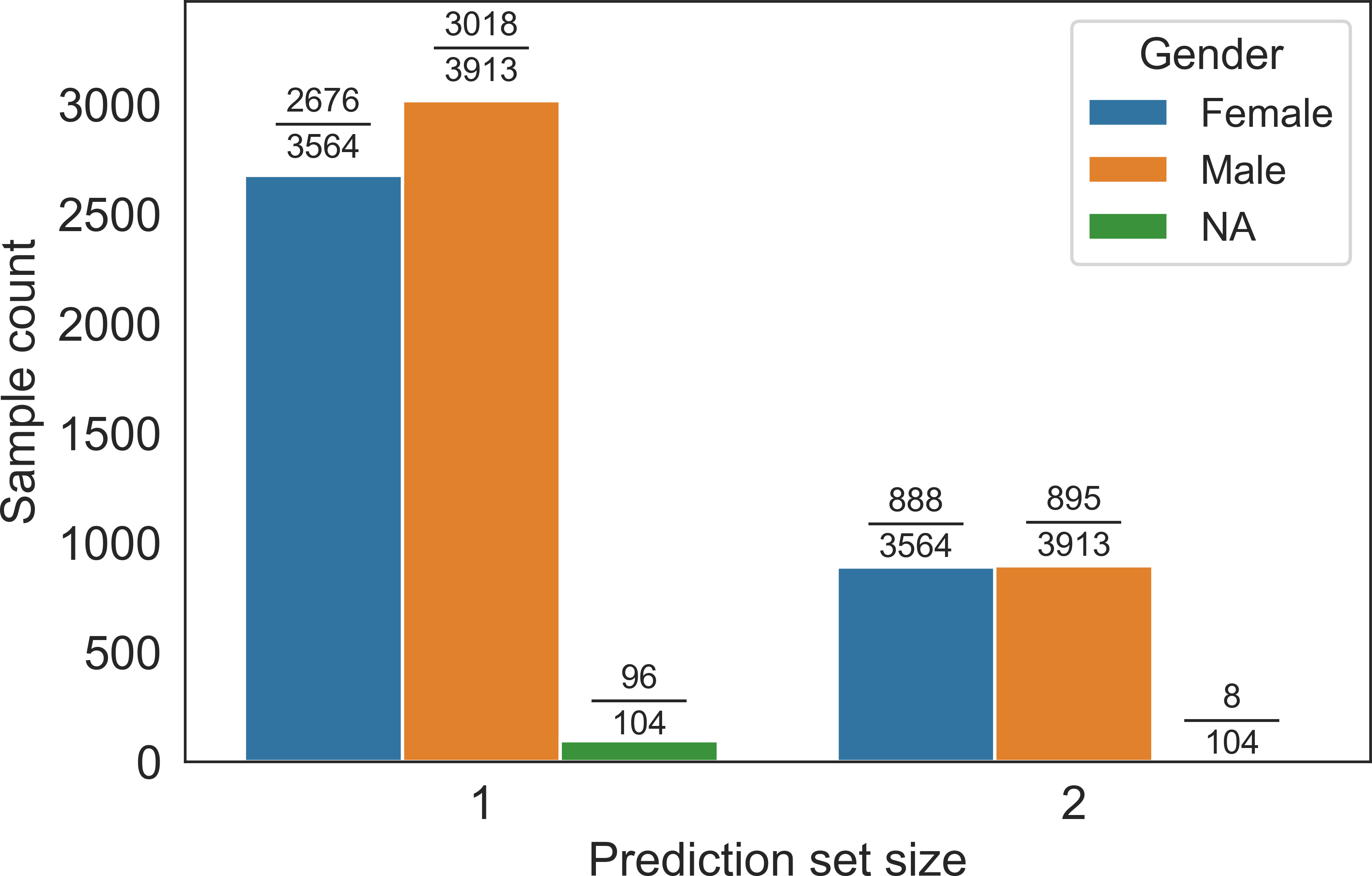}
		\caption{}
		\label{fig:setsize_isic_gender}
	\end{subfigure}
	\hfill
	\begin{subfigure}[t]{0.48\linewidth}
		\centering
		\includegraphics[width=\linewidth]{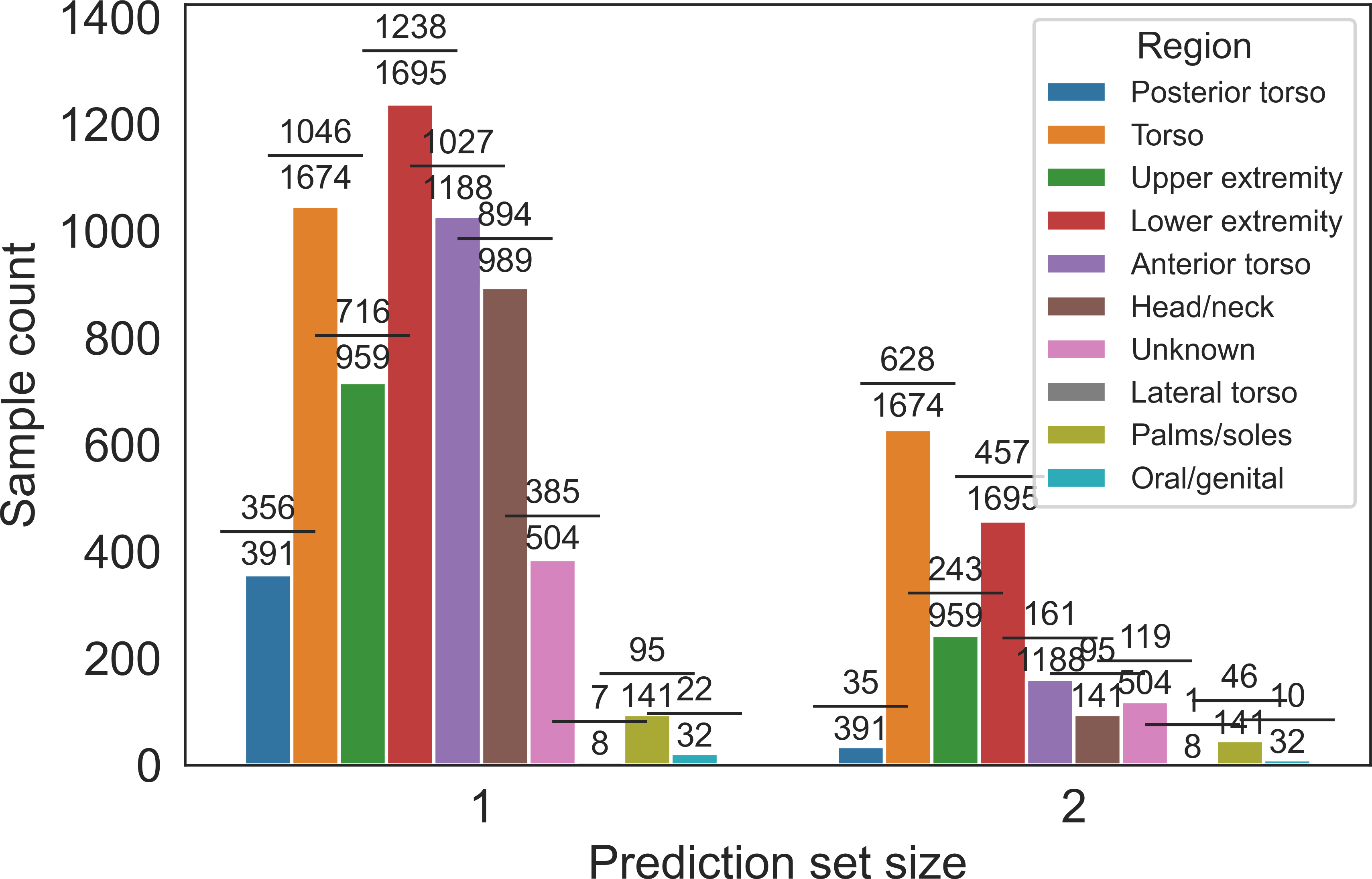}
		\caption{}
		\label{fig:setsize_isic_region}
	\end{subfigure}
	\hfill
	\begin{subfigure}[t]{0.48\linewidth}
		\centering
		\includegraphics[width=\linewidth]{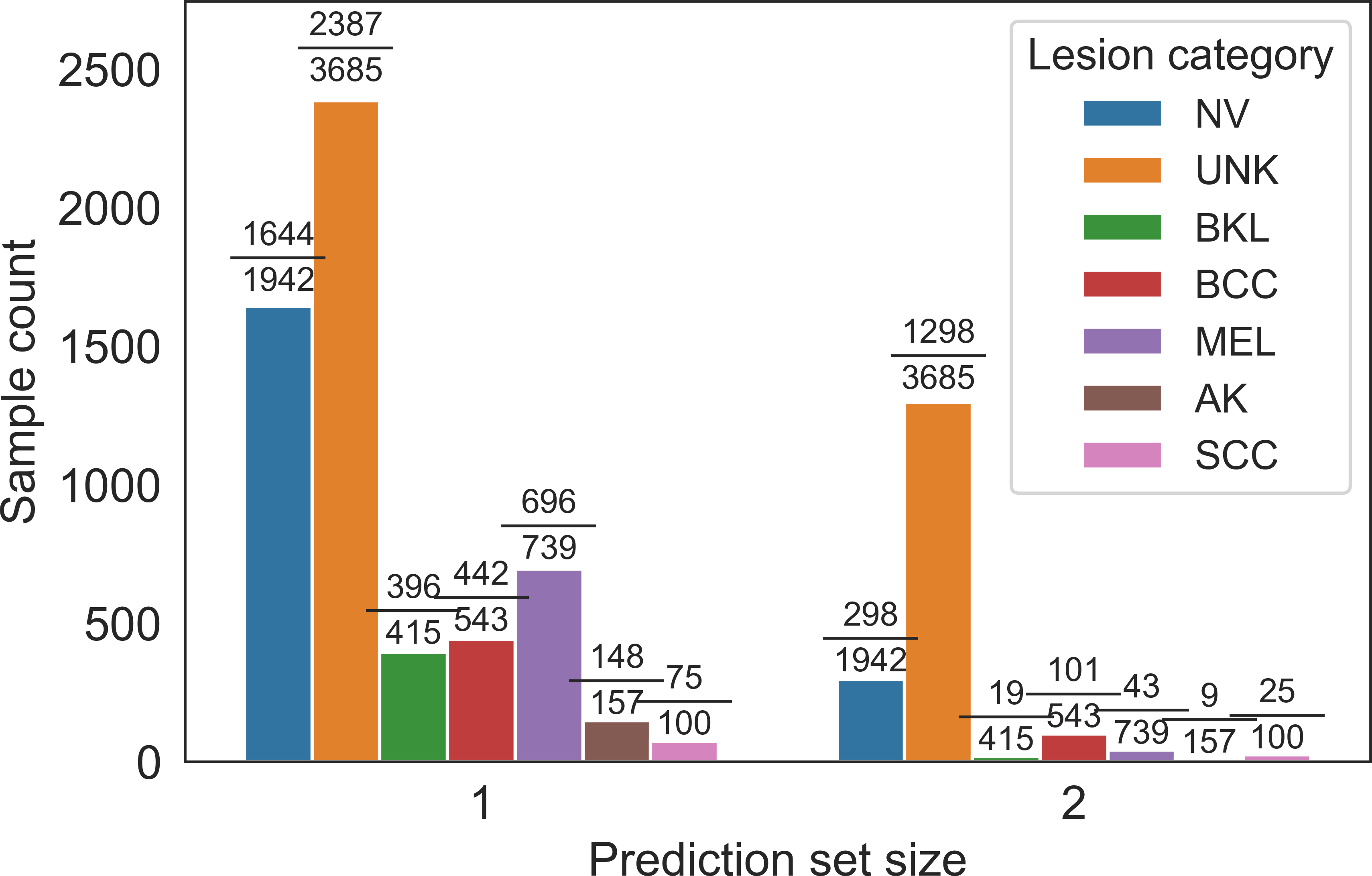}
		\caption{}
		\label{fig:setsize_isic_category}
	\end{subfigure}
	\hfill
	\caption{Distribution of conformal prediction set cardinalities across different strata on the ISIC dataset (a) age group, (b) gender, (c) anatomical region, and (d) lesion category.}
	\label{fig:setsize_distribution_isic}
\end{figure}

\begin{figure}[!t]
	\centering
	\begin{subfigure}[t]{0.48\linewidth}
		\centering
		\includegraphics[width=\linewidth]{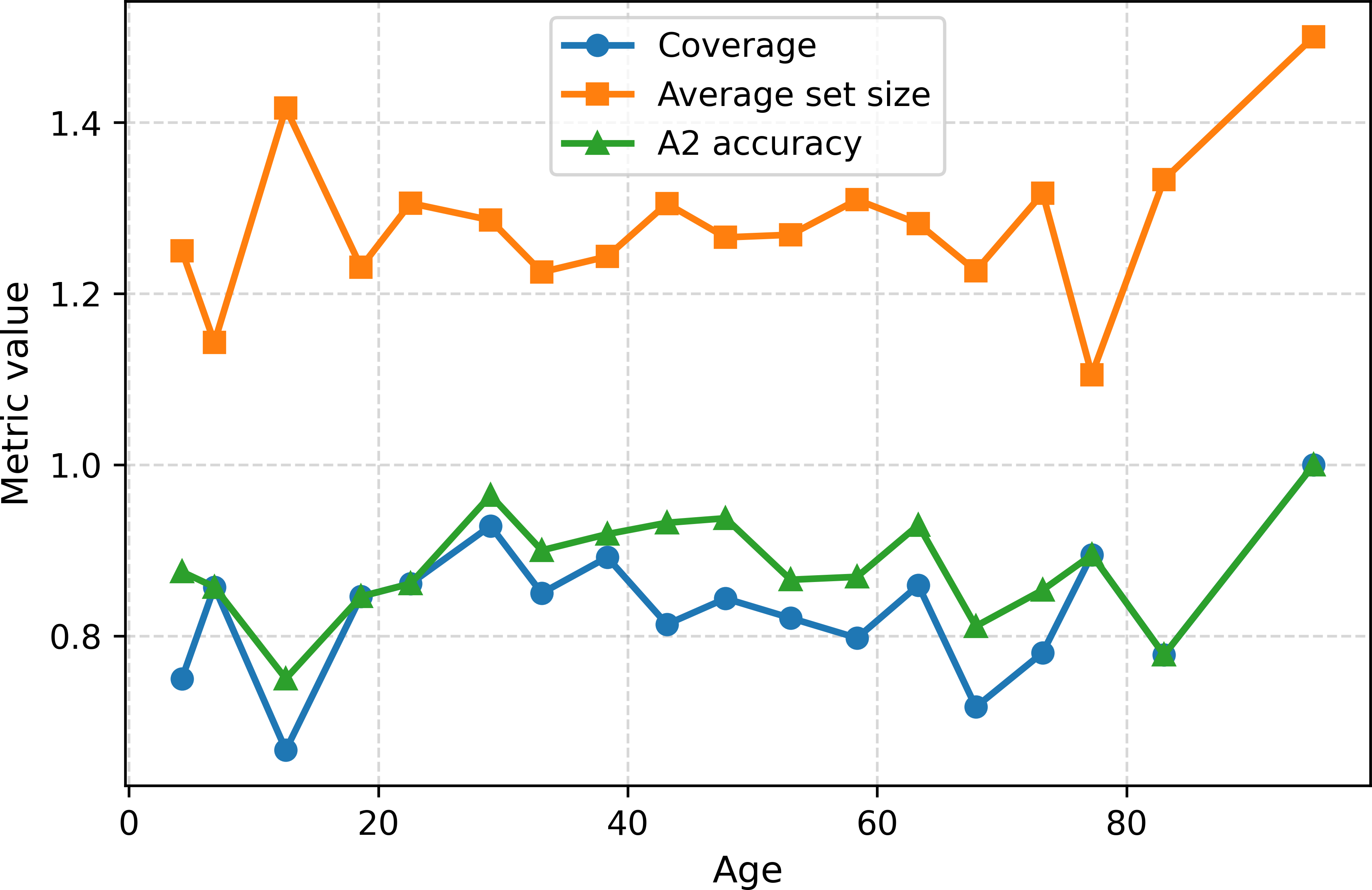}
		\caption{}
		\label{fig:age_trend_oscc}
	\end{subfigure}
	\hfill
	\begin{subfigure}[t]{0.48\linewidth}
		\centering
		\includegraphics[width=\linewidth]{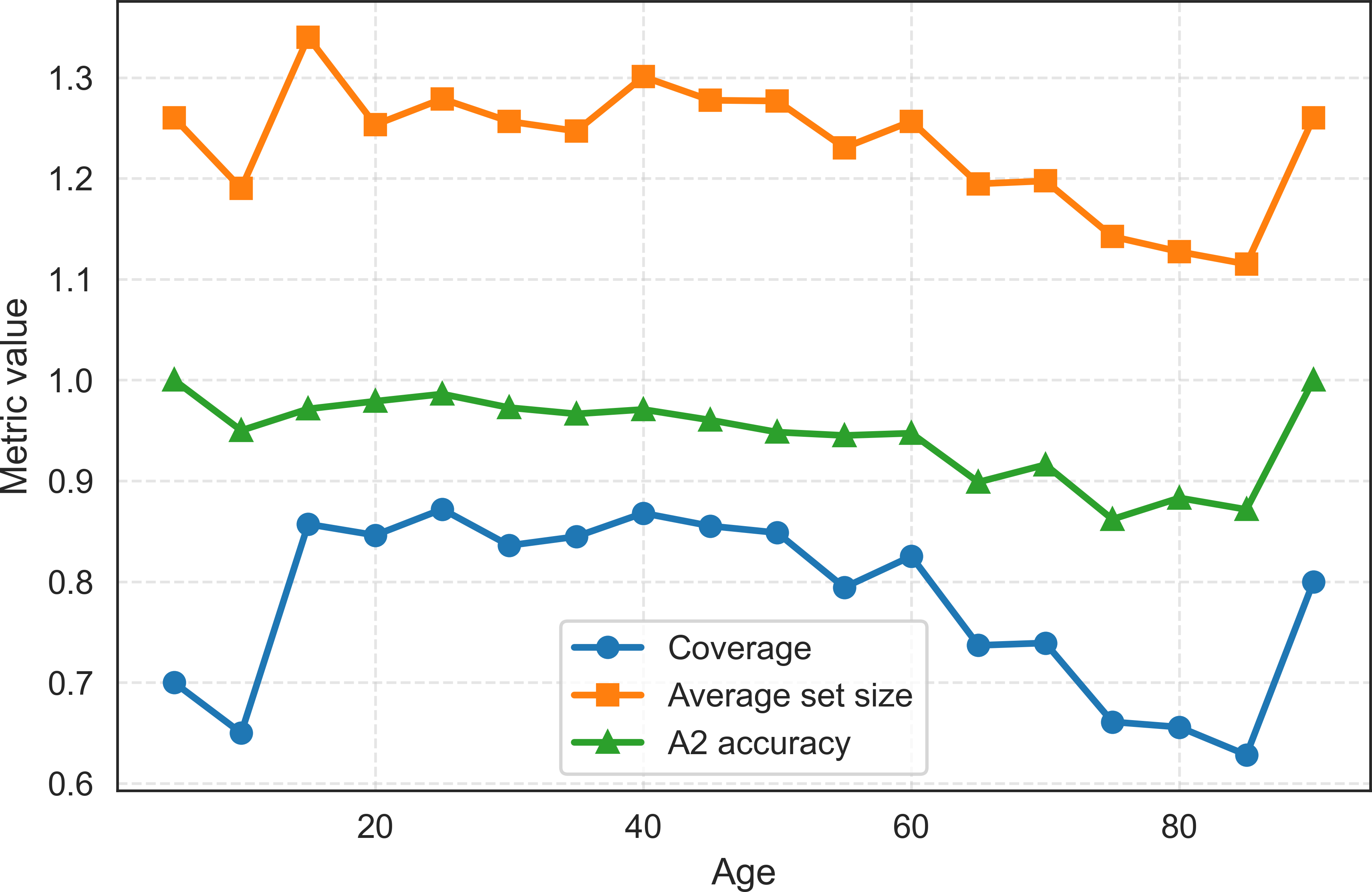}
		\caption{}
		\label{fig:age_trend_isic}
	\end{subfigure}
	\hfill
	\caption{Age-wise variation of conformal coverage, average prediction set size, and A2 accuracy across the (a) OSCC and (b) ISIC datasets.}
	\label{fig:age_trend}
\end{figure}

To further investigate uncertainty characteristics on the OSCC dataset, Fig.~\ref{fig:setsize_distribution} illustrates the distribution of conformal prediction set sizes across different strata. The results indicate that the majority of samples across all subgroups are associated with singleton prediction sets, reflecting confident predictions. However, a noticeable proportion of samples belong to prediction set size two, particularly across clinically relevant risk-related subgroups, suggesting increased uncertainty near ambiguous decision regions. Prediction sets of size three appear only in a limited number of cases, indicating that highly uncertain cases constitute a relatively small fraction of the dataset. Similarly, Fig.~\ref{fig:setsize_distribution_isic} illustrates the distribution of conformal prediction set sizes across different strata on the ISIC dataset. The results show that singleton prediction sets remain dominant across most demographic, anatomical region, and lesion-category subgroups, indicating generally confident predictions. Nevertheless, a considerable number of samples belong to prediction set size two, particularly across certain lesion categories and anatomical regions, reflecting increased uncertainty and overlap among visually similar skin lesion patterns. For clarity of visualization, only the most frequent prediction set sizes are displayed, while larger set sizes occurring in relatively fewer cases are omitted from the figure. Overall, the observed distribution suggests that uncertainty remains concentrated around low-cardinality prediction sets across ISIC subgroups.

Additionally, age-wise variations of conformal metrics for both OSCC and ISIC are illustrated in Fig.~\ref{fig:age_trend} using finer-grained age bins. Across both datasets, conformal coverage remains comparatively stable within the middle age ranges, while localized fluctuations become more apparent toward younger and older populations. Prediction set size and A2 accuracy also exhibit age-dependent variations, revealing subgroup-specific differences in uncertainty and model confidence across the age spectrum.

\section{Conclusion} 
\label{sec:conclusion}
This work addresses a recurring difficulty in medical image classification, where disease categories lying between well-defined states produce ambiguous predictions that standard classifiers resolve arbitrarily and conformal predictors leave unresolved as multi-label sets.
An adaptive conformal redistribution (AdaConRed) strategy is proposed, in which an entropy-modulated, margin-aware score constructs adaptive prediction sets whose internal structure is then used to resolve ambiguity. Formal coverage remains the role of the standard split conformal set; AdaConRed operates on top of it as a label-free decision rule, using only model outputs at inference.
Conformal prediction is thereby extended beyond uncertainty estimation toward uncertainty-aware decision support.
Experimental results on two real-world medical imaging tasks involving oral and skin lesion datasets demonstrate improved classification performance, with gains concentrated in the clinically critical malignant categories. A descriptive subgroup analysis further characterizes how uncertainty varies across demographic and lesion strata.
The non-malignant, ambiguous, and malignant grouping used here is a conceptual illustration for the two medical case studies; the redistribution mechanism itself requires only a designated transitional category and is not tied to these particular labels. Future work will investigate extensions including risk-stratified conformal prediction, hierarchical uncertainty modeling, and task-adaptive nonconformity measures with tighter integration into foundation model representations.

\begin{credits}
\subsubsection{\ackname} S. Ghosh acknowledges support from the National Postdoctoral Fellowship, Anusandhan National Research Foundation (ANRF), Government of India, under File No. PDF/2025/003781. 

\subsubsection{\discintname}
The authors declare that they have no known competing financial interests or personal relationships that could have appeared to influence the work reported in this paper.
\end{credits}
%
%
%

\begin{thebibliography}{50}
	
	\bibitem{welikala2020automated}
	Welikala, R.A., Remagnino, P., Lim, J.H., Chan, C.S., Rajendran, S.,
	Kallarakkal, T.G., Zain, R.B., Jayasinghe, R.D., Rimal, J., Kerr, A.R., et al.:
	Automated detection and classification of oral lesions using deep learning for
	early detection of oral cancer. IEEE Access \textbf{8}, 132677--132693 (2020).
	\doi{10.1109/ACCESS.2020.3010180}
	
	\bibitem{ISIC1}
	Rotemberg, V., Kurtansky, N., Betz-Stablein, B., Caffery, L., Chousakos, E.,
	Codella, N., Combalia, M., Dusza, S., Guitera, P., Gutman, D., et al.:
	A patient-centric dataset of images and metadata for identifying melanomas
	using clinical context. Scientific Data \textbf{8}, 34 (2021).
	\doi{10.1038/s41597-021-00815-z}
	
	\bibitem{fu2020deep}
	Fu, Q., Chen, Y., Li, Z., Jing, Q., Hu, C., Liu, H., Bao, J., Hong, Y.,
	Shi, T., Li, K., et al.: A deep learning algorithm for detection of oral cavity
	squamous cell carcinoma from photographic images: a retrospective study.
	eClinicalMedicine \textbf{27}, 100558 (2020). \doi{10.1016/j.eclinm.2020.100558}
	
	\bibitem{ma2025cmpb}
	Ma, S., Chan, Y.H., Ngai, E.C.H., Ho, J.W.K.: Asynchronous and focal federated
	learning for skin lesion classification under local data scarcity and class
	imbalance. Computer Methods and Programs in Biomedicine \textbf{272}, 109073
	(2025). \doi{10.1016/j.cmpb.2025.109073}
	
	\bibitem{cui2019class}
	Cui, Y., Jia, M., Lin, T.Y., Song, Y., Belongie, S.: Class-balanced loss based
	on effective number of samples. In: Proceedings of the IEEE/CVF Conference on
	Computer Vision and Pattern Recognition, pp. 9260--9269 (2019).
	\doi{10.1109/CVPR.2019.00949}
	
	\bibitem{moris2024adapted}
	Mor\'{i}s, D.I., de Moura, J., Novo, J., Ortega, M.: Adapted generative latent
	diffusion models for accurate pathological analysis in chest X-ray images.
	Medical \& Biological Engineering \& Computing \textbf{62}, 2189--2212 (2024).
	\doi{10.1007/s11517-024-03056-5}
	
	\bibitem{frid2018gan}
	Frid-Adar, M., Diamant, I., Klang, E., Amitai, M., Goldberger, J.,
	Greenspan, H.: GAN-based synthetic medical image augmentation for increased CNN
	performance in liver lesion classification. Neurocomputing \textbf{321},
	321--331 (2018). \doi{10.1016/j.neucom.2018.09.013}
	
	\bibitem{kazerouni2023diffusion}
	Kazerouni, A., Aghdam, E.K., Heidari, M., Azad, R., Fayyaz, M.,
	Hacihaliloglu, I., Merhof, D.: Diffusion models in medical imaging: a
	comprehensive survey. Medical Image Analysis \textbf{88}, 102846 (2023).
	\doi{10.1016/j.media.2023.102846}
	
	\bibitem{rani2024self}
	Rani, V., Kumar, M., Gupta, A., Sachdeva, M., Mittal, A., Kumar, K.:
	Self-supervised learning for medical image analysis: a comprehensive review.
	Evolving Systems \textbf{15}, 1607--1633 (2024).
	\doi{10.1007/s12530-024-09581-w}
	
	\bibitem{zhang2023biomedclip}
	Zhang, S., Xu, Y., Usuyama, N., Xu, H., Bagga, J., Tinn, R., Preston, S.,
	Rao, R., Wei, M., Valluri, N., et al.: BiomedCLIP: a multimodal biomedical
	foundation model pretrained from fifteen million scientific image-text pairs.
	arXiv preprint arXiv:2303.00915 (2023). \doi{10.48550/arXiv.2303.00915}
	
	\bibitem{kiraly2024dermfound}
	Kiraly, A.P., Baur, S., Philbrick, K., Mahvar, F., Yatziv, L., Chen, T.,
	Sterling, B., George, N., Jamil, F., Tang, J., et al.: Health AI developer
	foundations. arXiv preprint arXiv:2411.15128 (2024).
	\doi{10.48550/arXiv.2411.15128}
	
	\bibitem{sellergreen2025medsiglip}
	Sellergreen, A., Kazemzadeh, S., Jaroensri, T., Kiraly, A., Traverse, M.,
	Kohlberger, T., Xu, S., Jamil, F., Hughes, C., Lau, C., et al.: MedGemma
	technical report. arXiv preprint arXiv:2507.05201 (2025).
	\doi{10.48550/arXiv.2507.05201}
	
	\bibitem{tzeng2025uncertaintydiffusion}
	Tzeng, J.H., Huang, C.Y., Chen, K.W.: Uncertainty diffusion: parameter-efficient
	depth refinement via uncertainty-guided diffusion models. In: British Machine
	Vision Conference (2025). \url{https://bmvc2025.bmva.org/proceedings/128/}
	
	\bibitem{fu2025counting}
	Fu, S., Zhou, J., Chen, Q., Jing, H., Nguyen, H.A., Liu, X., Zeng, Z., Ma, L.,
	Zhang, Q., Wu, Q.: Counting hallucinations in diffusion models. arXiv preprint
	arXiv:2510.13080 (2025). \doi{10.48550/arXiv.2510.13080}
	
	\bibitem{aithal2024understanding}
	Aithal, S.K., Maini, P., Lipton, Z.C., Kolter, J.Z.: Understanding
	hallucinations in diffusion models through mode interpolation. In: Advances in
	Neural Information Processing Systems, pp. 134614--134644 (2024).
	\doi{10.52202/079017-4278}
	
	\bibitem{zou2023uncertainty}
	Zou, K., Chen, Z., Yuan, X., Shen, X., Wang, M., Fu, H.: A review of uncertainty
	estimation and its application in medical imaging. Meta-Radiology
	\textbf{1}(1), 100003 (2023). \doi{10.1016/j.metrad.2023.100003}
	
	\bibitem{nair2020exploring}
	Nair, T., Precup, D., Arnold, D.L., Arbel, T.: Exploring uncertainty measures in
	deep networks for multiple sclerosis lesion detection and segmentation. Medical
	Image Analysis \textbf{59}, 101557 (2020). \doi{10.1016/j.media.2019.101557}
	
	\bibitem{vovk2005algorithmic}
	Vovk, V., Gammerman, A., Shafer, G.: Algorithmic Learning in a Random World.
	2nd edn. Springer, Cham (2022). \doi{10.1007/978-3-031-06649-8}
	
	\bibitem{shi2024conformal}
	Shi, Y., Ghosh, S., Belkhouja, T., Doppa, J.R., Yan, Y.: Conformal prediction
	for class-wise coverage via augmented label rank calibration. In: Advances in
	Neural Information Processing Systems, pp. 132133--132178 (2024).
	\doi{10.52202/079017-4200}
	
	\bibitem{fayyad2024cmpb}
	Fayyad, J., Alijani, S., Najjaran, H.: Empirical validation of conformal
	prediction for trustworthy skin lesions classification. Computer Methods and
	Programs in Biomedicine \textbf{253}, 108231 (2024).
	\doi{10.1016/j.cmpb.2024.108231}
	
	\bibitem{OSCC}
	Piyarathne, N.S., Liyanage, S.N., Rasnayaka, R.M.S.G.K., Hettiarachchi, P.V.K.S.,
	Devindi, G.A.I., Francis, F.B.A.H., Dissanayake, D.M.D.R., Ranasinghe, R.A.N.S.,
	Pavithya, M.B.D., Nawinne, I.B., et al.: A comprehensive dataset of annotated
	oral cavity images for diagnosis of oral cancer and oral potentially malignant
	disorders. Oral Oncology \textbf{156}, 106946 (2024).
	\doi{10.1016/j.oraloncology.2024.106946}
	
	\bibitem{dosovitskiy2020image}
	Dosovitskiy, A., Beyer, L., Kolesnikov, A., Weissenborn, D., Zhai, X.,
	Unterthiner, T., Dehghani, M., Minderer, M., Heigold, G., Gelly, S., et al.:
	An image is worth 16x16 words: transformers for image recognition at scale.
	In: International Conference on Learning Representations (2021).
	\doi{10.48550/arXiv.2010.11929}
	
	\bibitem{romano2020classification}
	Romano, Y., Sesia, M., Cand\`{e}s, E.: Classification with valid and adaptive
	coverage. In: Advances in Neural Information Processing Systems, pp. 3581--3591
	(2020). \doi{10.48550/arXiv.2006.02544}
	
	\bibitem{bhattacharyya2026}
	Bhattacharyya, S., Pal, U., Chakraborti, T.: Conformal uncertainty
	quantification to evaluate predictive fairness of foundation AI model for skin
	lesion classes across patient demographics. Health Information Science and
	Systems \textbf{14}(1), 18 (2026). \doi{10.1007/s13755-025-00412-z}
	
	\bibitem{banerji2023}
	Banerji, C.R.S., Chakraborti, T., Harbron, C., MacArthur, B.D.: Clinical AI
	tools must convey predictive uncertainty for each individual patient. Nature
	Medicine \textbf{29}(12), 2996--2998 (2023). \doi{10.1038/s41591-023-02562-7}
	
	\bibitem{chakraborti2025}
	Chakraborti, T., Banerji, C.R.S., Marandon, A., Hellon, V., Mitra, R.,
	Lehmann, B., Br\"{a}uninger, L., McGough, S., Turkay, C., Frangi, A.F., et al.:
	Personalized uncertainty quantification in artificial intelligence. Nature
	Machine Intelligence \textbf{7}(4), 522--530 (2025).
	\doi{10.1038/s42256-025-01024-8}
	
	\bibitem{dey2026}
	Dey, S., Banerji, C.R.S., Basuchowdhuri, P., Saha, S.K., Parashar, D.,
	Chakraborti, T.: Generating crossmodal gene expression from cancer
	histopathology improves multimodal AI predictions. Nature Communications
	\textbf{17}, 259 (2026). \doi{10.1038/s41467-025-66961-9}
	
	\bibitem{ISIC2}
	Nugroho, E.S., Ardiyanto, I., Nugroho, H.A.: Addressing imbalance ISIC-2019
	dataset in dermoscopic pigmented skin lesion classification. ICIC Express
	Letters \textbf{18}(6), 563--573 (2024). \doi{10.24507/icicel.18.06.563}
	
	\bibitem{sun2025cmpb}
	Sun, Y., Wen, X., Zhang, Y., Jin, L., Yang, C., Zhang, Q., Jiang, M., Xu, Z.,
	Guo, W., Su, J., et al.: Visual-language foundation models in medical imaging:
	a systematic review and meta-analysis of diagnostic and analytical applications.
	Computer Methods and Programs in Biomedicine \textbf{268}, 108870 (2025).
	\doi{10.1016/j.cmpb.2025.108870}
	
	\bibitem{google2025flow}
	Google Labs: Google Flow: AI creative studio for video, images and custom tools.
	\url{https://labs.google/fx/tools/flow}, last accessed 2026/08/15
	
	\bibitem{comanici2025gemini25}
	Comanici, G., Bieber, E., Schaekermann, M., Pasupat, I., Sachdeva, N.,
	Dhillon, I., Blistein, M., Ram, O., Zhang, D., Rosen, E., et al.: Gemini 2.5:
	pushing the frontier with advanced reasoning, multimodality, long context, and
	next generation agentic capabilities. arXiv preprint arXiv:2507.06261 (2025).
	\doi{10.48550/arXiv.2507.06261}
	
	\bibitem{google2024dermfoundation}
	Google Health AI: Derm Foundation model.
	\url{https://developers.google.com/health-ai-developer-foundations/derm-foundation},
	last accessed 2026/08/15
	
	\bibitem{pal2026isbi}
	Pal, V., Bandyopadhyay, R., Bhattacharya, S., Saha, S., Chakraborti, T.:
	Vision foundation models pretrained on skin lesion images can classify oral
	lesions too! In: IEEE International Symposium on Biomedical Imaging, pp. 1--5
	(2026). \doi{10.1109/ISBI61048.2026.11515748}
	
	\bibitem{Angelopoulos2022}
	Angelopoulos, A.N., Bates, S.: A gentle introduction to conformal prediction and
	distribution-free uncertainty quantification. arXiv preprint arXiv:2107.07511
	(2021). \doi{10.48550/arXiv.2107.07511}
	
	\bibitem{LAC}
	Sadinle, M., Lei, J., Wasserman, L.: Least ambiguous set-valued classifiers with
	bounded error levels. Journal of the American Statistical Association
	\textbf{114}(525), 223--234 (2019). \doi{10.1080/01621459.2017.1395341}
	
	\bibitem{RAPS}
	Angelopoulos, A.N., Bates, S., Jordan, M.I., Malik, J.: Uncertainty sets for
	image classifiers using conformal prediction. In: International Conference on
	Learning Representations (2021). \doi{10.48550/arXiv.2009.14193}
	
	\bibitem{wang2023image}
	Wang, W., Bao, H., Dong, L., Bjorck, J., Peng, Z., Liu, Q., Aggarwal, K.,
	Mohammed, O.K., Singhal, S., Som, S., et al.: Image as a foreign language: BEiT
	pretraining for vision and vision-language tasks. In: Proceedings of the
	IEEE/CVF Conference on Computer Vision and Pattern Recognition, pp. 19175--19186
	(2023). \doi{10.1109/CVPR52729.2023.01838}
	
	\bibitem{eslami2023pubmedclip}
	Eslami, S., Meinel, C., de Melo, G.: PubMedCLIP: how much does CLIP benefit
	visual question answering in the medical domain? In: Findings of the Association
	for Computational Linguistics: EACL 2023, pp. 1181--1193 (2023).
	\doi{10.18653/v1/2023.findings-eacl.88}
	
	\bibitem{lozano2025biomedica}
	Lozano, A., Sun, M.W., Burgess, J., Chen, L., Nirschl, J.J., Gu, J., Lopez, I.,
	Aklilu, J., Rau, A., Katzer, A.W., et al.: BIOMEDICA: an open biomedical
	image-caption archive, dataset, and vision-language models derived from
	scientific literature. In: Proceedings of the IEEE/CVF Conference on Computer
	Vision and Pattern Recognition, pp. 19724--19735 (2025).
	\doi{10.1109/CVPR52734.2025.01837}
	
	\bibitem{fang2024aligning}
	Fang, X., Lin, Y., Zhang, D., Cheng, K.T., Chen, H.: Aligning medical images
	with general knowledge from large language models. In: International Conference
	on Medical Image Computing and Computer-Assisted Intervention, pp. 57--67.
	Springer (2024). \doi{10.1007/978-3-031-72117-5_6}
	
	\bibitem{phan2024decomposing}
	Phan, V.M.H., Xie, Y., Qi, Y., Liu, L., Liu, L., Zhang, B., Liao, Z., Wu, Q.,
	To, M.S., Verjans, J.W.: Decomposing disease descriptions for enhanced pathology
	detection: a multi-aspect vision-language pre-training framework. In:
	Proceedings of the IEEE/CVF Conference on Computer Vision and Pattern
	Recognition, pp. 11492--11501 (2024). \doi{10.1109/CVPR52733.2024.01092}
	
	\bibitem{kumar2024improving}
	Kumar, Y., Marttinen, P.: Improving medical multi-modal contrastive learning
	with expert annotations. In: European Conference on Computer Vision,
	pp. 468--486. Springer (2024). \doi{10.1007/978-3-031-72661-3_27}
	
	\bibitem{du2024ret}
	Du, J., Guo, J., Zhang, W., Yang, S., Liu, H., Li, H., Wang, N.: RET-CLIP: a
	retinal image foundation model pre-trained with clinical diagnostic reports.
	In: International Conference on Medical Image Computing and Computer-Assisted
	Intervention, pp. 709--719. Springer (2024). \doi{10.1007/978-3-031-72390-2_66}
	
	\bibitem{benyahia2022multi}
	Benyahia, S., Meftah, B., L\'{e}zoray, O.: Multi-features extraction based on
	deep learning for skin lesion classification. Tissue and Cell \textbf{74},
	101701 (2022). \doi{10.1016/j.tice.2021.101701}
	
	\bibitem{shakya2025comprehensive}
	Shakya, M., Patel, R., Joshi, S.: A comprehensive analysis of deep learning and
	transfer learning techniques for skin cancer classification. Scientific Reports
	\textbf{15}(1), 4633 (2025). \doi{10.1038/s41598-024-82241-w}
	
	\bibitem{tahir2023dscc_net}
	Tahir, M., Naeem, A., Malik, H., Tanveer, J., Naqvi, R.A., Lee, S.W.:
	DSCC\_Net: multi-classification deep learning models for diagnosing of skin
	cancer using dermoscopic images. Cancers \textbf{15}(7), 2179 (2023).
	\doi{10.3390/cancers15072179}
	
	\bibitem{gouda2022detection}
	Gouda, W., Sama, N.U., Al-Waakid, G., Humayun, M., Jhanjhi, N.Z.: Detection of
	skin cancer based on skin lesion images using deep learning. Healthcare
	\textbf{10}(7), 1183 (2022). \doi{10.3390/healthcare10071183}
	
	\bibitem{kim2021computer}
	Kim, C.I., Hwang, S.M., Park, E.B., Won, C.H., Lee, J.H.: Computer-aided
	diagnosis algorithm for classification of malignant melanoma using deep neural
	networks. Sensors \textbf{21}(16), 5551 (2021). \doi{10.3390/s21165551}
	
	\bibitem{murugan2021diagnosis}
	Murugan, A., Nair, S.A.H., Preethi, A.A.P., Kumar, K.P.S.: Diagnosis of skin
	cancer using machine learning techniques. Microprocessors and Microsystems
	\textbf{81}, 103727 (2021). \doi{10.1016/j.micpro.2020.103727}
	
	\bibitem{dai2026prompt}
	Dai, G., Tian, Z., Xin, C., Dai, D., Wang, C., Zhang, Y., Chen, H.,
	Hamilton, M.: Prompt-level contrastive learning for context-aware multi-modal
	image representation in medical diagnosis. Pattern Recognition \textbf{174},
	113027 (2026). \doi{10.1016/j.patcog.2025.113027}
	
\end{thebibliography}
%

\end{document}